\documentclass{article}

\PassOptionsToPackage{numbers, compress}{natbib}

\usepackage[preprint]{neurips_2026}

\usepackage{hyperref}
\usepackage{url}
\usepackage{booktabs}
\usepackage{amsfonts}
\usepackage{nicefrac}
\usepackage{microtype}
\usepackage{xcolor}
\usepackage{multirow}
\usepackage{graphicx}
\usepackage{subcaption}

\usepackage[most]{tcolorbox}
\usepackage{xcolor}
\usepackage{caption}
\usepackage{cancel}
\definecolor{findingblue}{RGB}{72,120,180}
\definecolor{findingbg}{RGB}{248,251,255}

\newtcolorbox[auto counter]{finding}[2][]{%
    enhanced,
    breakable,
    sharp corners,
    colback=findingbg,
    colframe=findingblue,
    boxrule=1pt,
    left=2mm,
    right=2mm,
    top=1.2mm,
    bottom=1.2mm,
    colbacktitle=findingblue,
    coltitle=white,
    fonttitle=\bfseries,
    title={Finding~\thetcbcounter: #2},
    #1
}

\title{The Many Faces of On-Policy Distillation: Pitfalls, Mechanisms, and Fixes}

\makeatletter
\let\NeurIPS@orig@maketitle\@maketitle
\renewcommand{\@maketitle}{%
  \renewcommand{\@makefnmark}{\mbox{\textsuperscript{\normalfont\@thefnmark}}}%
  \NeurIPS@orig@maketitle%
}
\makeatother

\author{%
  Siqi Zhu\textsuperscript{1} \quad
  Xuyan Ye\textsuperscript{2}%
  \thanks{Co-second authorship}%
  \thanks{Work done during an internship at UIUC.} \quad
  Hongyu Lu\textsuperscript{2}\footnotemark[1]\footnotemark[2] \quad
  Weiye Shi\textsuperscript{3}\footnotemark[1]\footnotemark[2] \quad
  Ge Liu\textsuperscript{1} \\[0.4em]
  {\normalfont\textsuperscript{1}UIUC \quad \textsuperscript{2}Renmin University of China \quad \textsuperscript{3}Peking University}
}

\usepackage[most]{tcolorbox}
\usepackage{xcolor}
\usepackage{fvextra}

\newtcolorbox{promptbox}{
  enhanced,
  breakable,
  colback=blue!3,
  colframe=blue!50!black,
  boxrule=0.6pt,
  arc=2mm,
  title={User Prompt}
}

\newtcolorbox{responsebox}{
  enhanced,
  breakable,
  colback=green!3,
  colframe=green!45!black,
  boxrule=0.6pt,
  arc=2mm,
  title={Assistant Response}
}

\usepackage{amsthm}
\usepackage{amsmath}

\begin{document}

\maketitle

\begin{abstract}

On-policy distillation (OPD) and on-policy self-distillation (OPSD) have emerged
as promising post-training methods for large language models, offering dense
token-level supervision on trajectories sampled from the model's own policy.
However, existing results on their effectiveness remain mixed: while OP(S)D has
shown promise in system prompt and knowledge internalization, recent studies also report
instability and degradation. In this work, we present a
comprehensive empirical study of when OPD and OPSD work, when they fail, and why. We find that
OPD on mathematical reasoning is highly sensitive to teacher choice and loss formulation, whereas OPSD fails in our tested settings due to test-time absence of instance-specific privileged information (PI). In contrast,
OPSD is effective when PI represents a shared latent rule,
such as a system prompt or alignment preference. We identify three failure
mechanisms: (1) distribution mismatch between teacher and student caused by conditioning on student-generated prefixes, (2) optimization
instability from biased TopK reverse-KL gradients, and (3) an
OPSD-specific limitation where the student learns a PI-free policy that aggregates PI-conditioned teachers, which
is insufficient when PI is instance-specific. We further show that stop-gradient TopK objectives, RLVR-adapted teachers, and SFT-stabilized students mitigate these failures.

\end{abstract}

\section{Introduction}

\begin{figure}[!t]
  \centering
  \includegraphics[width=\textwidth]{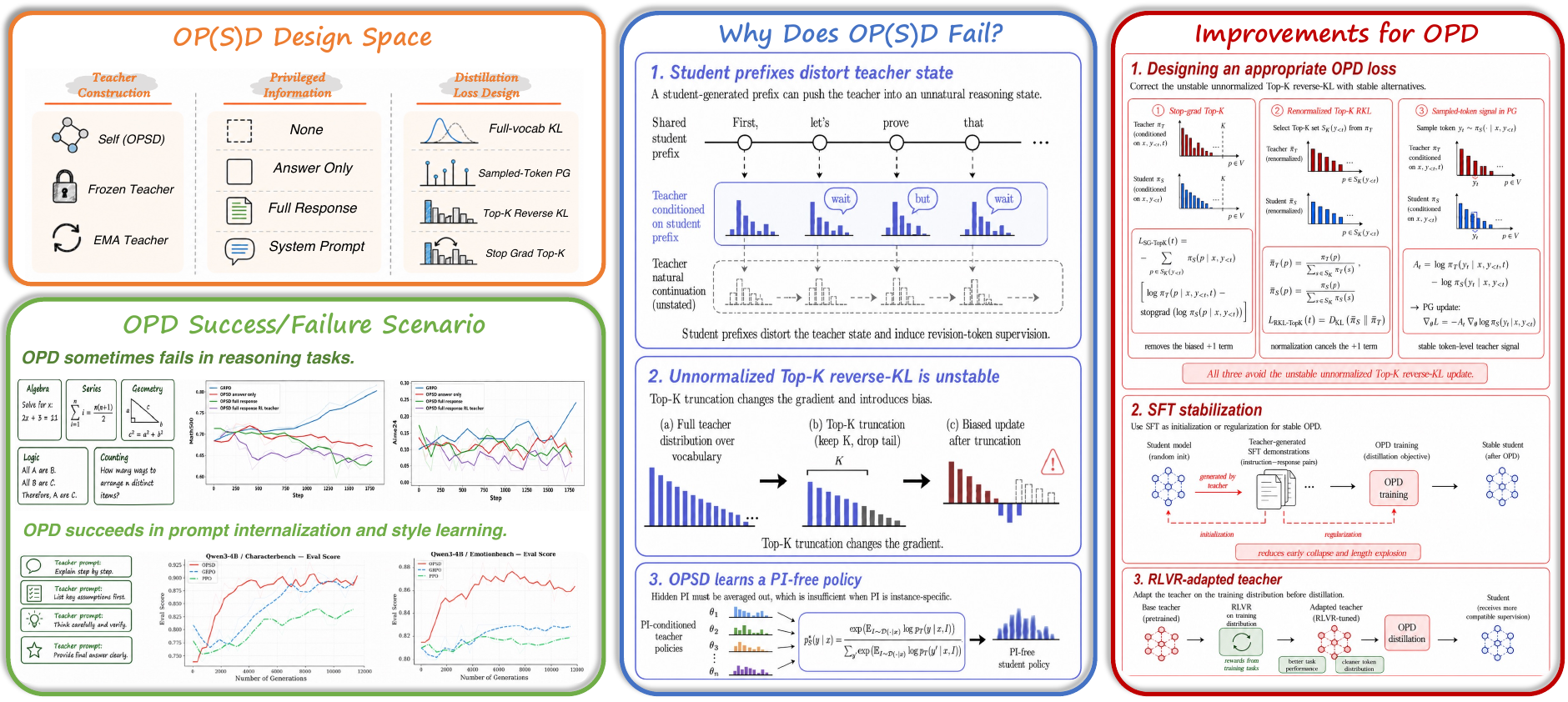}
  \caption{
    \textbf{Overview.} We map the OP(S)D \emph{design space} (left, top) and its task-dependent success/failure behavior (left, bottom), identify three failure mechanisms—prefix-distorted teacher state, biased Top-$K$ reverse-KL, and PI-marginalized OPSD policy (middle), and propose practical fixes: stable Top-$K$ losses, SFT stabilization, and RLVR-adapted teachers (right).
  }
  \label{fig:overview}
\end{figure}

On-policy distillation (OPD) \cite{agarwal2024onpolicydistillationlanguagemodels, gu2026minillmonpolicydistillationlarge} is a promising post-training
algorithm for large language models (LLMs). Instead of training on off-policy responses, OPD trains the student on trajectories sampled from
its own policy and uses one or more teacher models to provide dense token-level supervision
along these on-policy rollouts. This makes OPD appealing as a practical mechanism
for integrating the capabilities of multiple stronger teachers into the student \cite{deepseekai2026deepseekv4, coreteam2026mimov2flashtechnicalreport}, mitigating
catastrophic forgetting, and improving sample efficiency \cite{lu2025onpolicydistillation, shenfeld2026selfdistillationenablescontinuallearning}. A closely related topic is on-policy self-distillation (OPSD), where the teacher is not a stronger model but the student model itself conditioned on
additional privileged information (PI) \cite{zhao2026selfdistilledreasoneronpolicyselfdistillation, sang2026onpolicyselfdistillationreasoningcompression,ye2026onpolicycontextdistillationlanguage}. The PI may include ground-truth answers, system prompts or user preferences. These methods are attractive for post-training because they can convert teacher behavior or training-time context into supervision on the student’s own state distribution.

Despite the conceptual simplicity, the empirical behavior of OPD and OPSD remains
poorly understood. Prior work has reported successful applications of
self-distillation for context internalization and reasoning \cite{zhao2026selfdistilledreasoneronpolicyselfdistillation, sang2026onpolicyselfdistillationreasoningcompression, ye2026onpolicycontextdistillationlanguage}. However, recent studies argue that
on-policy (self) distillation can be unstable and may degrade
performance \cite{fu2026revisitingonpolicydistillationempirical,  kim2026doesselfdistillationsometimesdegrade}. These mixed results raise a basic question: \textbf{when do OPD and OPSD
actually work, when do they fail, and what mechanisms explain the difference}?

In this paper, we present a comprehensive empirical study of OPD and OPSD across
both successful and failing regimes. On reasoning tasks, OPD provides only
conditional benefits and can suffer from length explosion, repetition, and
teacher-student mismatch, while OPSD fails in our tested math reasoning
settings. In contrast, OPSD is effective for system-prompt internalization and
alignment, where the privileged information induces a shared latent behavior that can be compressed into a PI-free policy at test time.

We identify three mechanisms behind these outcomes. First, student-generated
prefixes can move the teacher into states where its token-level supervision is
locally incompatible with the student's trajectory. Second, Top-$K$ reverse-KL
approximations can introduce biased gradients and destabilize optimization. Third, OPSD only learns a single PI-free consensus policy across PI-conditioned teachers, which is insufficient when PI is instance-specific.

Motivated by these findings, we study practical stabilizers for OPD: a
stop-gradient Top-$K$ KL surrogate that removes the biased gradient term, Reinforcement Learning with Verifiable Rewards (RLVR)
adaptation of the teacher to improve task performance, and Supervised Fine-Tuning (SFT) the student to improve output format and stabilize response-length dynamics.

Our contributions are:
\begin{itemize}
    \item We empirically study when OPD and OPSD succeed or fail in reasoning, system-prompt internalization, and alignment. \item We identify three failure mechanisms: prefix-induced teacher-student mismatch, biased TopK reverse-KL gradients, and OPSD's PI-free aggregation across instance-specific PI. 
    \item We propose practical stabilizers based on stop-gradient Top-$K$ KL,
    RLVR-adapted teachers, and student SFT for output well-formedness
    and response length stabilization.
\end{itemize}

\section{Related Work}

\paragraph{On-Policy Distillation and Context Distillation.}
\textbf{On-policy distillation} trains the student on its own sampled trajectories and uses a teacher to provide dense supervision. Agarwal et al.~\cite{agarwal2024onpolicydistillationlanguagemodels} show that this formulation can outperform standard off-policy distillation, and later work demonstrates its effectiveness for math reasoning~\cite{lu2025onpolicydistillation}. A closely related topic is \textbf{context distillation}, which turns in-context behaviors into model parameters. Prior work shows that models can distill instructions and knowledge from context into persistent capabilities~\cite{snell2022learningdistillingcontext}. Recent work extends this idea to on-policy distillation for continual learning~\cite{shenfeld2026selfdistillationenablescontinuallearning, ye2026onpolicycontextdistillationlanguage}.

\paragraph{Reinforcement Learning from Textual Feedback.}
Another related direction augments reinforcement learning with language-based feedback instead of relying only on scalar rewards. Song et al.~\cite{song2026expandingcapabilitiesreinforcementlearning} show that textual feedback can provide richer supervision over intermediate behaviors and improve agent capabilities. In reasoning tasks, POPE~\cite{qu2026popelearningreasonhard} uses privileged guidance to improve exploration on hard problems. Compared with these methods, our work also studies the role of additional textual information, but focuses on on-policy distillation rather than reinforcement learning objectives.
\section{Preliminary: On-Policy Distillation}

\begin{figure}[t]
    \centering
    \begin{subfigure}[c]{0.5\textwidth}
        \centering
        \includegraphics[width=\textwidth]{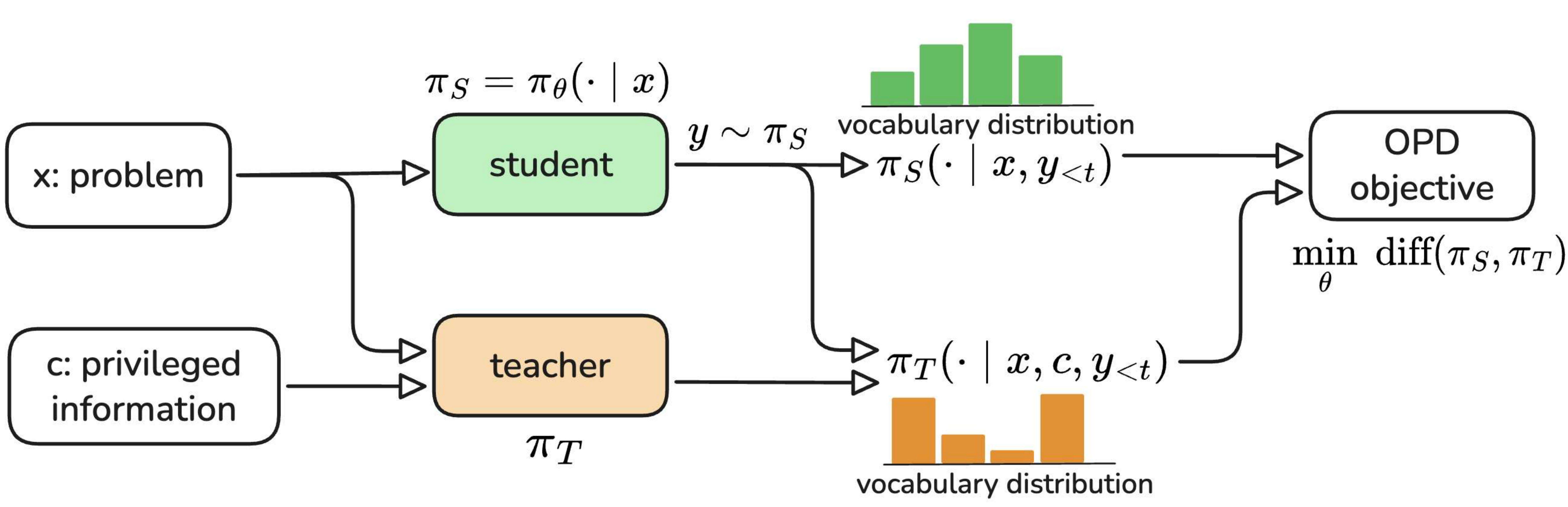}
        \label{fig:first}
    \end{subfigure}
    \begin{subfigure}[c]{0.23\textwidth}
        \centering
        \includegraphics[width=\textwidth]{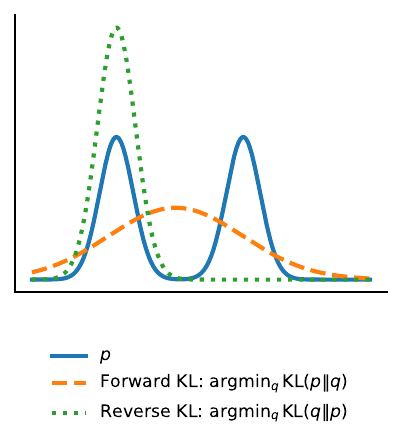}
        \label{fig:second}
    \end{subfigure}
    \vspace{-7mm}
    \caption{(\textbf{Left}) On-Policy (Self-)Distillation. In OPSD, the teacher is constructed from the student itself and privileged information (PI) is necessary. In OPD, the teacher is a stronger model and PI is optional. (\textbf{Right}) \textbf{p}: teacher distribution, \textbf{q}: student distribution. Reverse KL is mode-seeking, whereas forward KL is mode-covering.}
    \label{fig:teaser}
\end{figure}


Let $x\sim\mathcal D$ be an input, $\pi_\theta$ be the student policy, and
$\pi_T(\cdot\mid x,I)$ be the teacher policy with optional privileged information
$I$. In OPD, $\pi_T$ is a stronger external model. In OPSD, $\pi_T$ is a
student-derived policy augmented with PI.  As illustrated in Figure~\ref{fig:teaser}, OP(S)D trains on student-generated trajectories
$y=(y_1,\ldots,y_T)\sim\pi_\theta(\cdot\mid x)$. At each prefix $y_{<t}$, the
student receives token-level supervision from the teacher:
\begin{equation}
\mathcal L_{\mathrm{OP(S)D}}
=
\mathbb E_{x,y}
\left[
\frac{1}{T}\sum_{t=1}^T
\ell_t\!\left(
\pi_\theta(\cdot\mid x,y_{<t}),
\operatorname{stograd}\!\left(\pi_T(\cdot\mid x,y_{<t},I)\right)
\right)
\right].
\end{equation}

\paragraph{Full-vocabulary KL.}
A standard choice for $\ell_t$ is to match the teacher and student vocabulary with reverse KL:
\begin{equation}
D_{\mathrm{KL}}(\pi_\theta\|\pi_T)
=
\sum_v \pi_\theta(v)\log\frac{\pi_\theta(v)}{\pi_T(v)} .
\label{eq:fullreverseKL}
\end{equation}
Forward KL is mode-covering and encourages the student to cover the teacher
distribution. Reverse KL is mode-seeking and tends to preserve the student's
high-probability modes, making it less prone to catastrophic forgetting \cite{chan2022greedificationoperatorspolicyoptimization}. In OPD, forward KL is undesirable because it pushes the student toward teacher-preferred but student-unlikely tokens. 

For reverse KL, writing $p_\theta(v)=\pi_\theta(v\mid x,y_{<t})$
and $p_T(v)=\pi_T(v\mid x,y_{<t},I)$, its gradient is
\begin{equation}
\nabla_\theta D_{\mathrm{KL}}(p_\theta\|p_T)
=
\sum_v p_\theta(v)
\left[
\log\frac{p_\theta(v)}{p_T(v)} \cancel{\textbf{+1}}
\right]
\nabla_\theta \log p_\theta(v).
\end{equation}
In the full vocabulary, the constant $\textbf{+1}$ has zero contribution since
$
\sum_v p_\theta(v)\nabla_\theta\log p_\theta(v) = 0 .
$

\paragraph{Sampled-token KL.}
Another choice of $\ell_t$ uses the sampled token $y_t\sim\pi_\theta$.
The sampled-token objective treats the teacher-student log-probability gap as an advantage:
\begin{equation}
\mathcal L_{\mathrm{PG}}
=
-\mathbb E_{y\sim\pi_\theta}
\sum_t
\textbf{stopgrad}\big(\log\pi_T(y_t\mid x,y_{<t},I)
-\log\pi_\theta(y_t\mid x,y_{<t})\big)
\log\pi_\theta(y_t\mid x,y_{<t}),
\label{eq:policygrad_sampledtoken}
\end{equation}
whose policy-gradient form is
\begin{equation}
\nabla_\theta \mathcal L_{\mathrm{PG}}
=
-\mathbb E_{y\sim\pi_\theta}
\sum_t
\textbf{stopgrad}\big(\log\pi_T(y_t\mid x,y_{<t},I)
-\log\pi_\theta(y_t\mid x,y_{<t})\big)
\nabla_\theta\log\pi_\theta(y_t\mid x,y_{<t}).
\end{equation}
\section{Experiments}

We evaluate OPD and OPSD on reasoning, system-prompt internalization, and alignment, covering both failure and success regimes.

\subsection{Math Reasoning}

\paragraph{OPSD.}

\begin{figure}[t]
    \centering
    \begin{subfigure}[t]{0.32\textwidth}
        \centering
        \includegraphics[width=\textwidth]{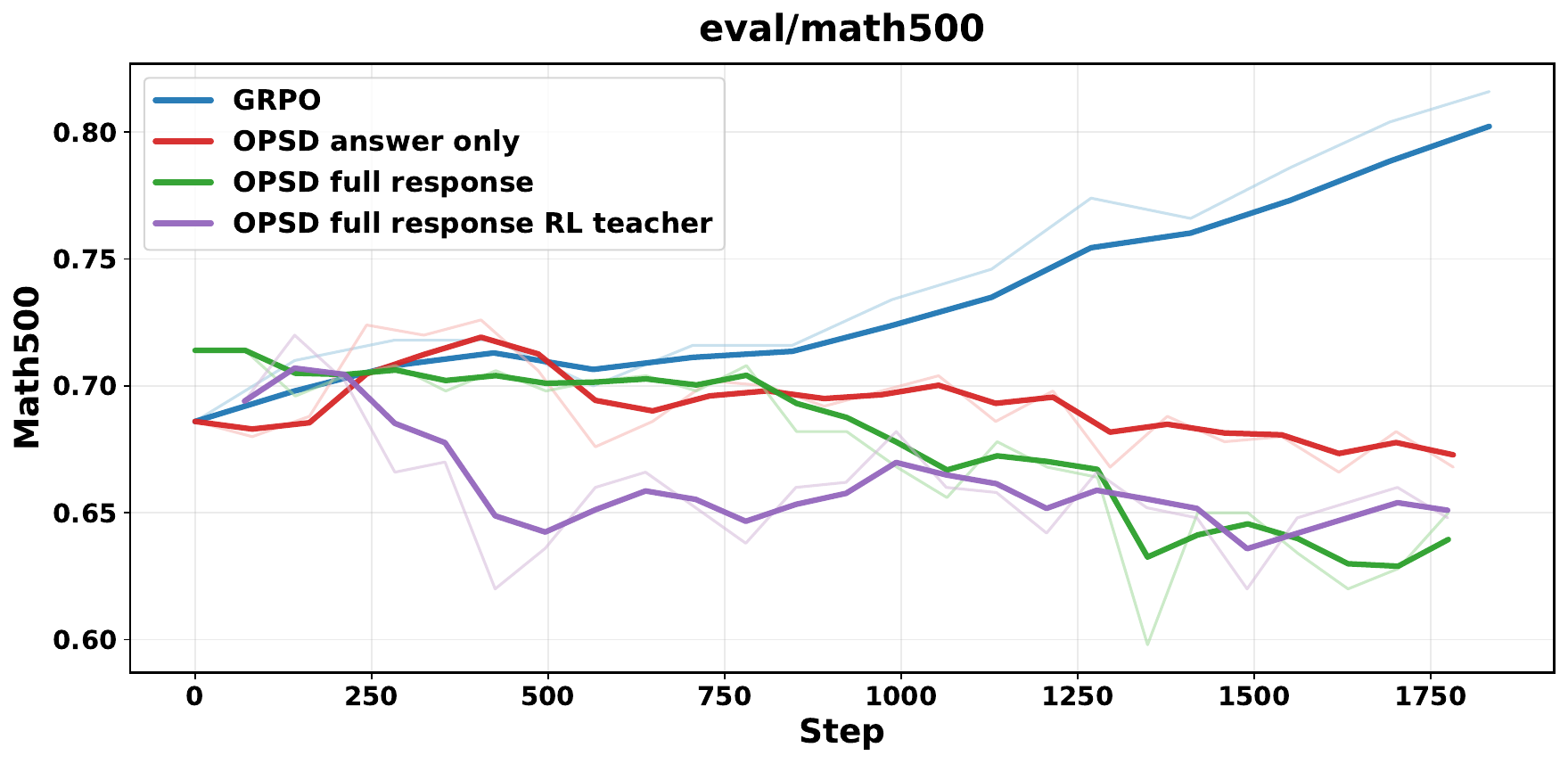}
        \label{fig:first}
    \end{subfigure}
    \begin{subfigure}[t]{0.33\textwidth}
        \centering
        \includegraphics[width=\textwidth]{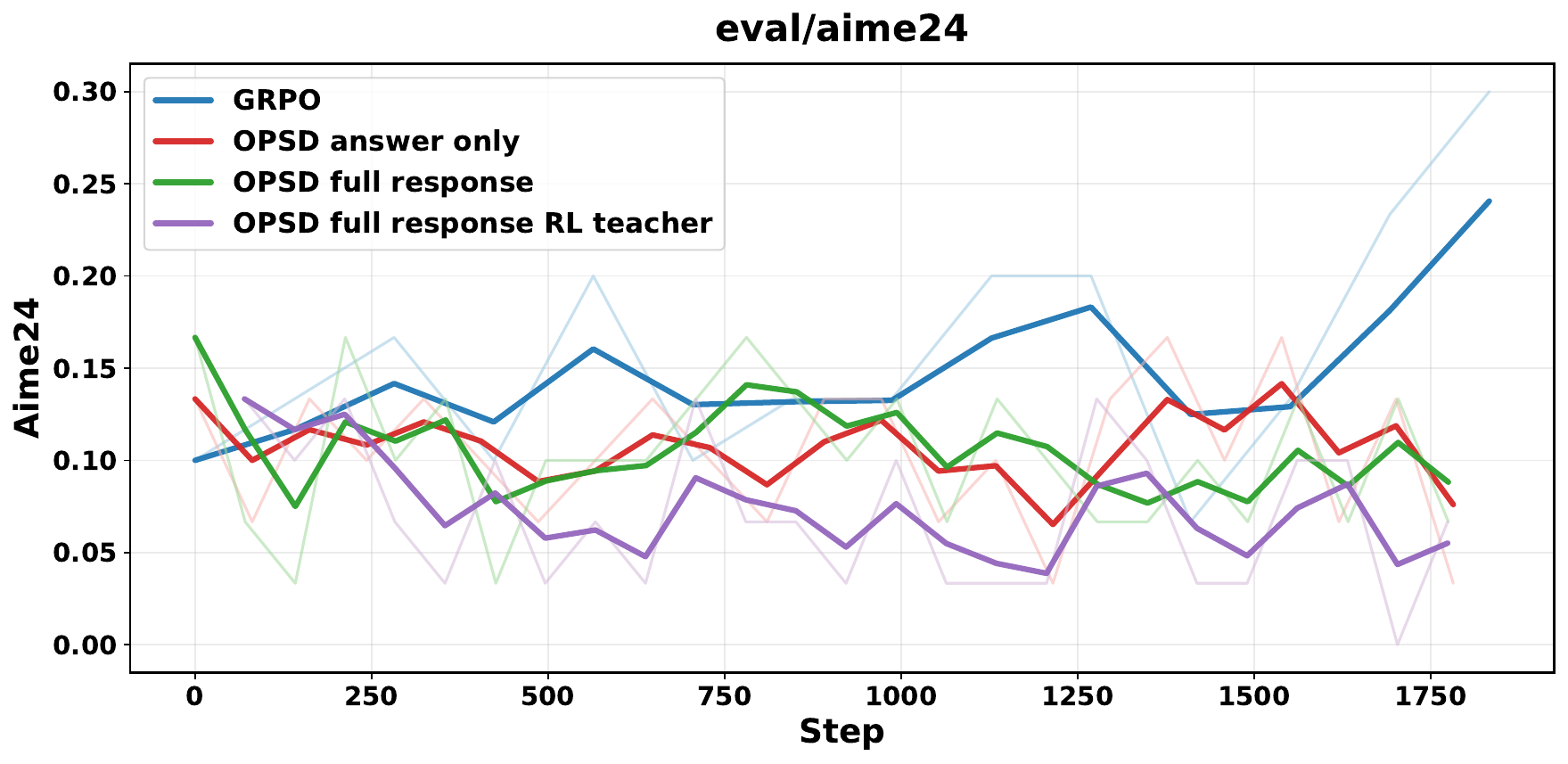}
        \label{fig:second}
    \end{subfigure}
    \begin{subfigure}[t]{0.33\textwidth}
        \centering
        \includegraphics[width=\textwidth]{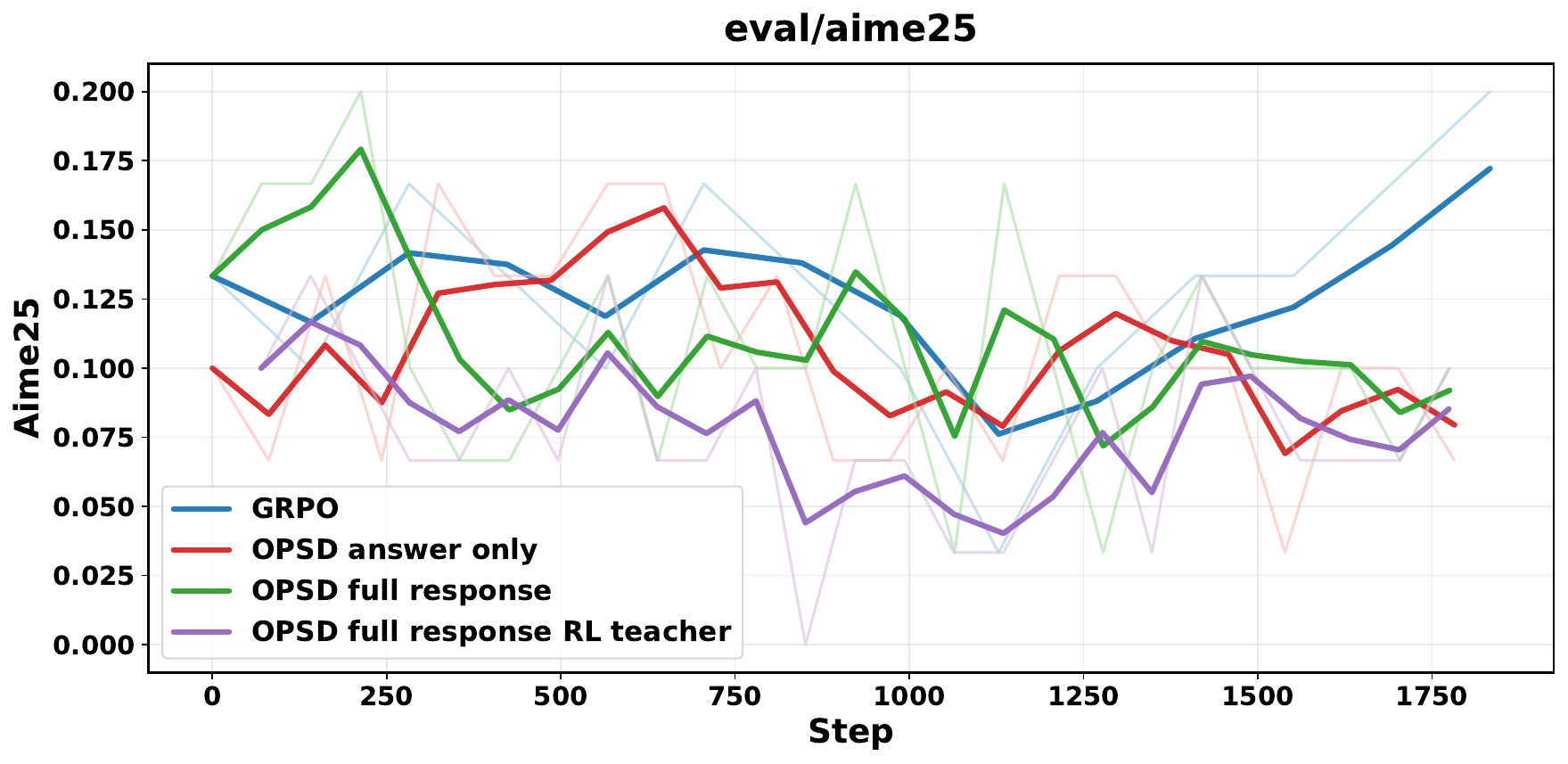}
        \label{fig:second}
    \end{subfigure}
    \vspace{-3mm}
    \caption{Qwen3-1.7B, trained on OpenThoughts. OPSD fails to improve student.}
    \label{fig:2opsdfails}
\end{figure}

We train Qwen3-1.7B on OpenThoughts using stopgrad Top20 reverse KL (Equation \ref{eq:topksgKL}), with the step-0 checkpoint as the PI-conditioned teacher. We keep English problems with verifiable boxed answers. As shown in Figure~\ref{fig:2opsdfails}, neither answer-only nor full-response PI makes stable improvements on Math500, AIME24, or AIME25. Full-response PI performs worse than answer-only
PI, suggesting that richer PI can increase mismatch rather than improve
supervision. We further test whether this failure is due to a weak
PI-conditioned teacher by training a Qwen3-1.7B teacher with GRPO under
full-response PI (purple line in Figure~\ref{fig:2opsdfails}). This RL-trained teacher performs even worse, indicating that
PI alone does not make math OPSD effective.

\begin{figure}[t]
    \centering
    \begin{subfigure}[t]{0.3\textwidth}
        \centering
        \includegraphics[width=\textwidth]{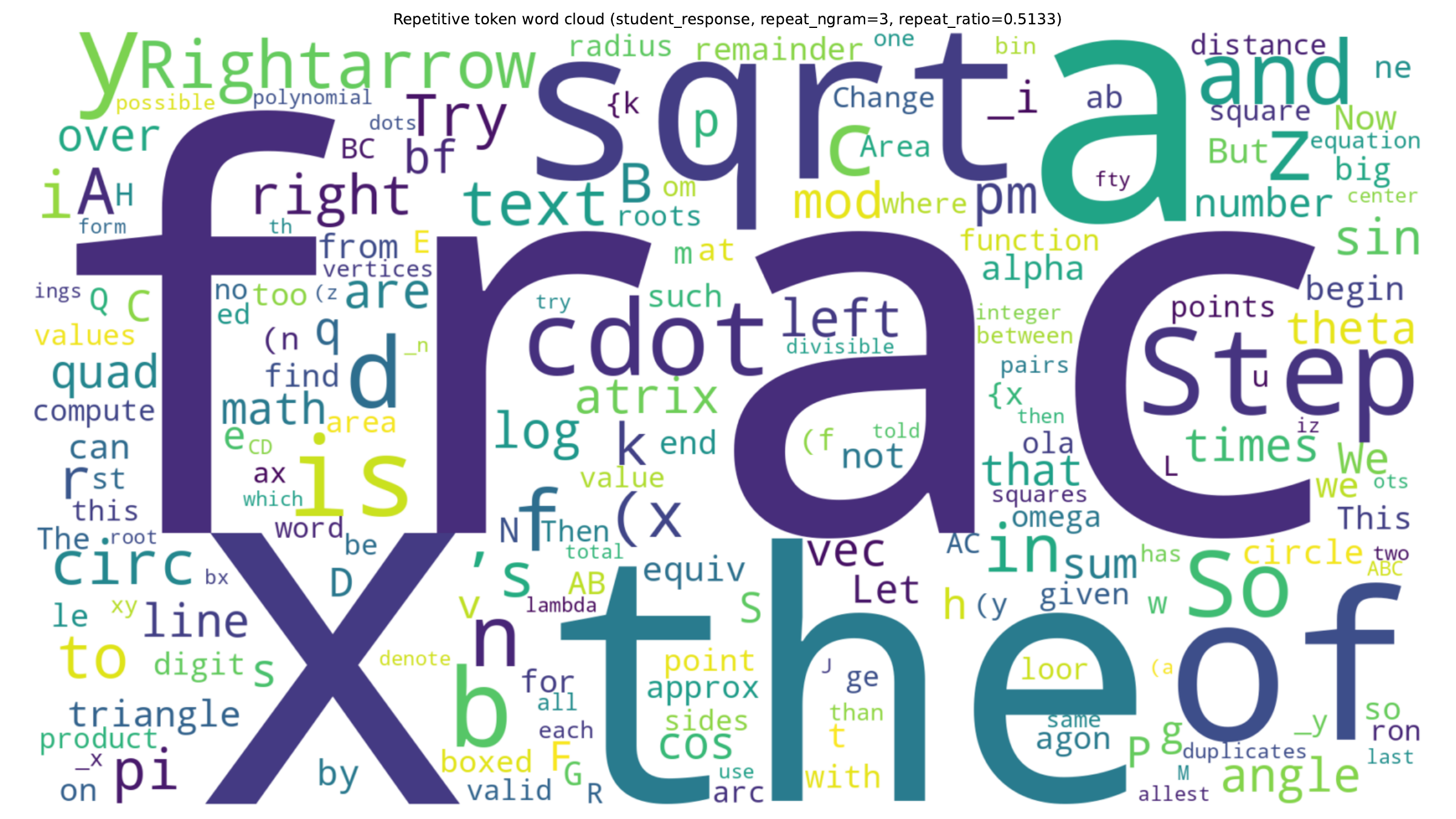}
        \caption{step0}
        \label{fig:step0}
    \end{subfigure}
    \hfill
    \begin{subfigure}[t]{0.3\textwidth}
        \centering
        \includegraphics[width=\textwidth]{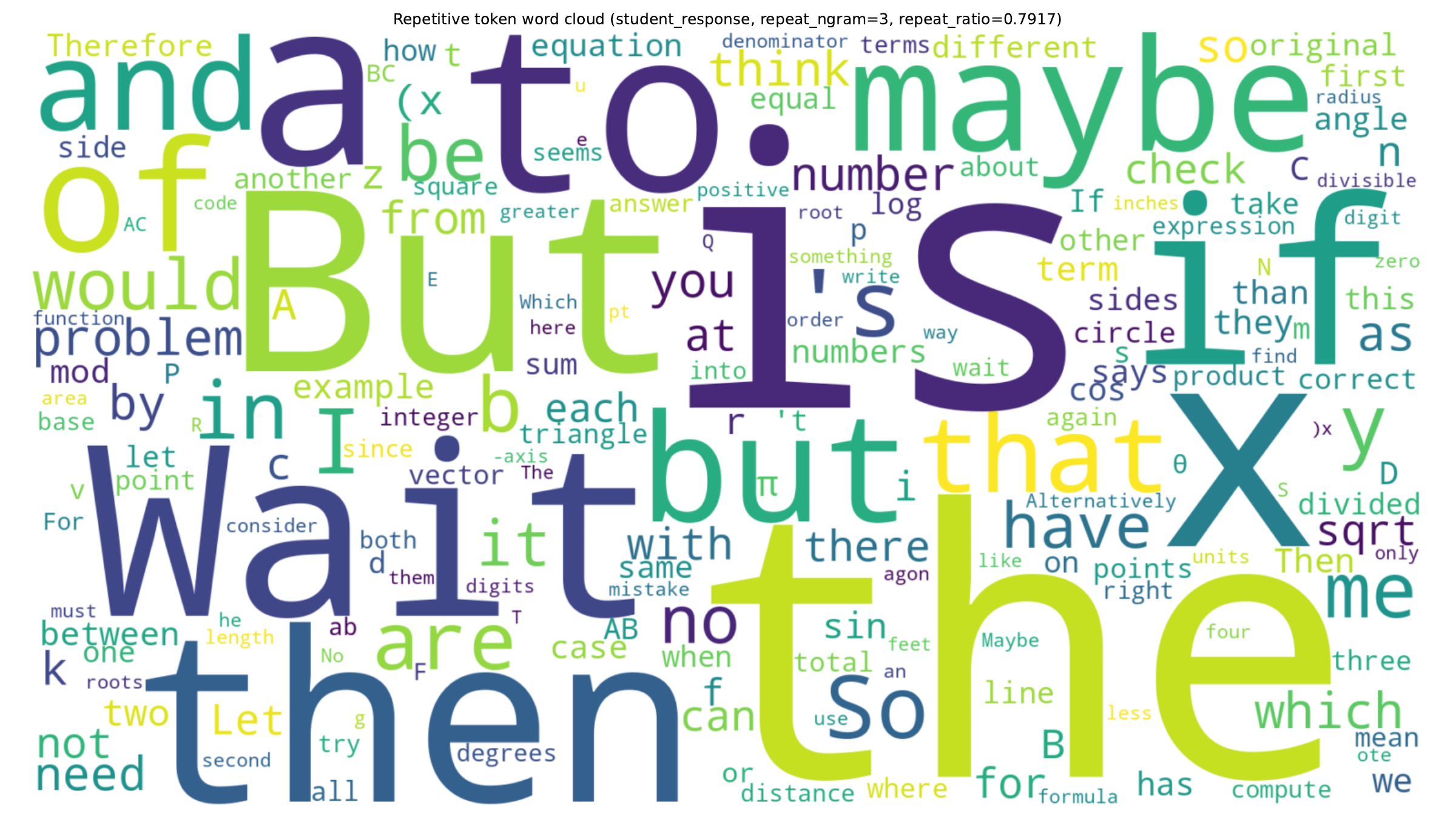}
        \caption{step700}
        \label{fig:step700}
    \end{subfigure}
    \hfill
    \begin{subfigure}[t]{0.3\textwidth}
        \centering
        \includegraphics[width=\textwidth]{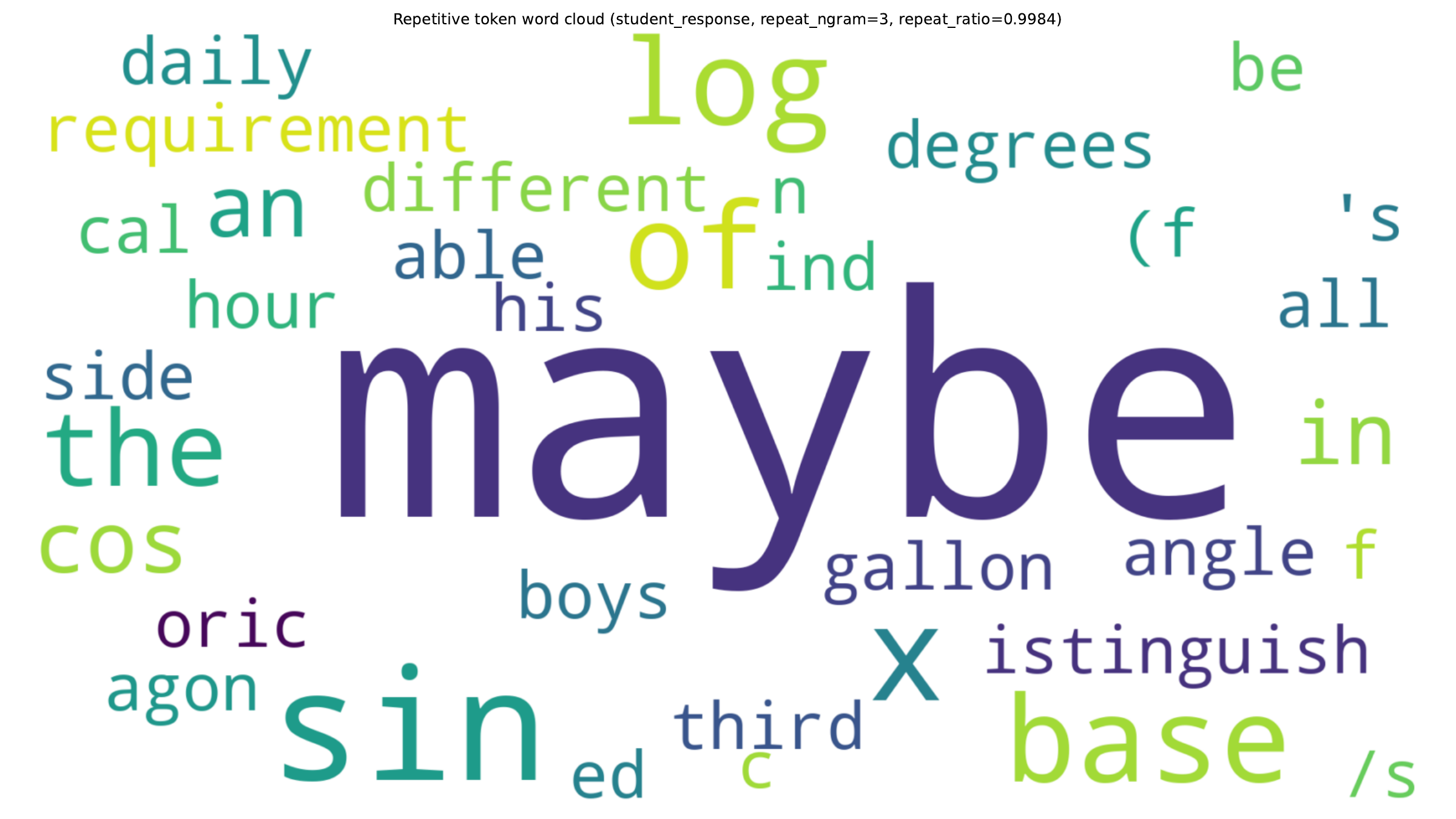}
        \caption{step1000}
        \label{fig:step1000}
    \end{subfigure}

    \centering
    \begin{subfigure}[t]{0.3\textwidth}
        \centering
        \includegraphics[width=\textwidth]{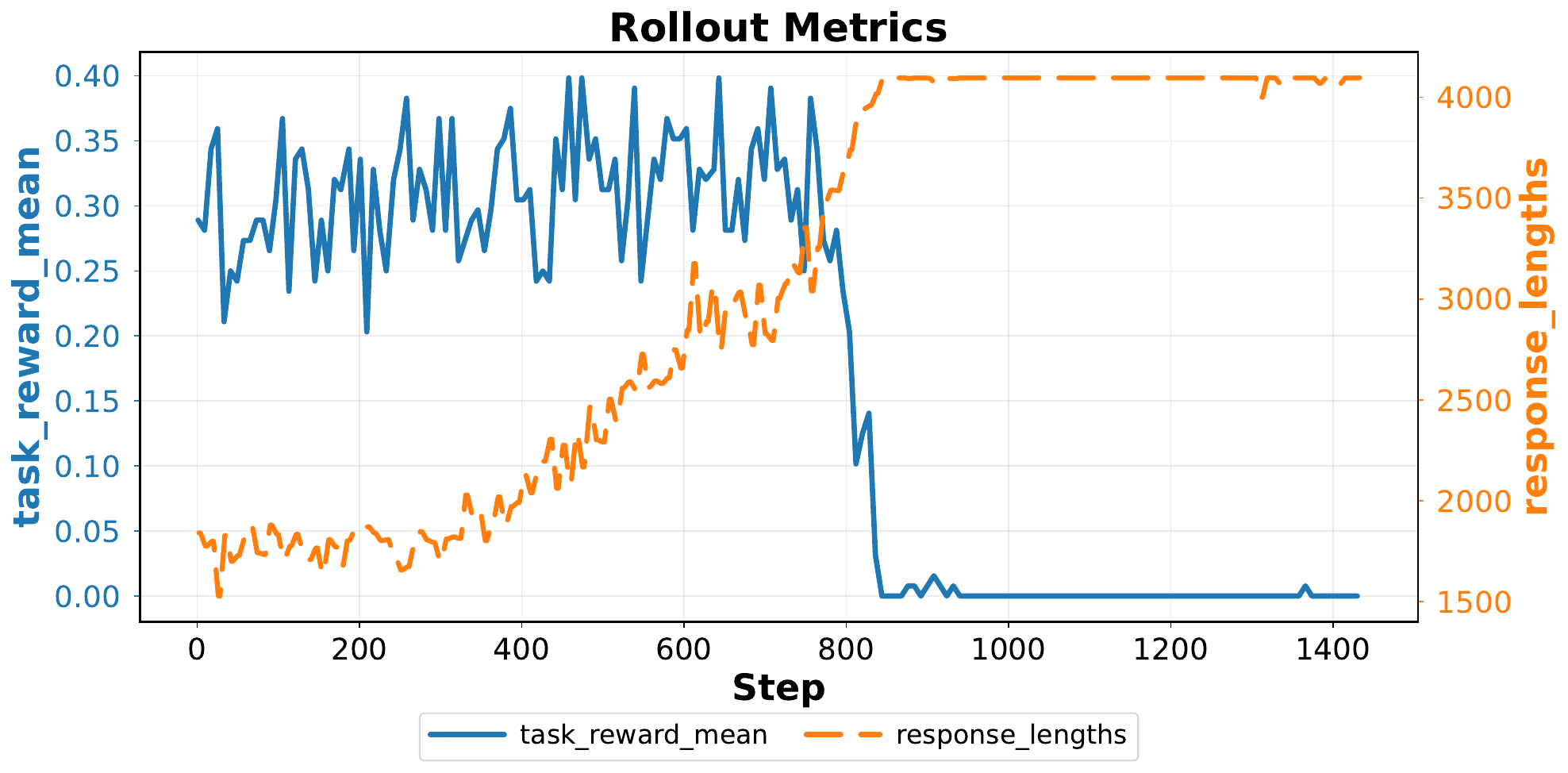}
        \caption{rollout metrics}
        \label{fig:rollout_metrics}
    \end{subfigure}
    \hfill
    \begin{subfigure}[t]{0.35\textwidth}
        \centering
        \includegraphics[width=\textwidth]{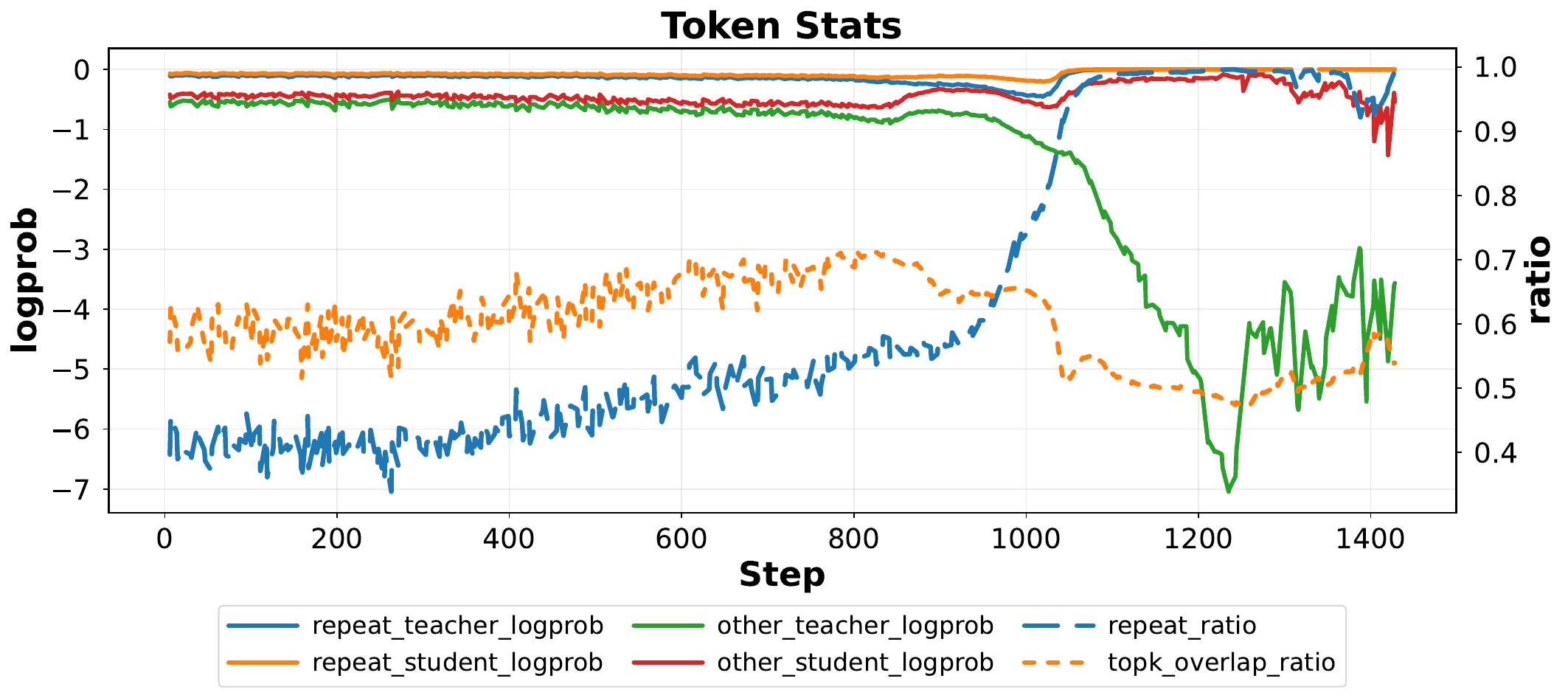}
        \caption{token stats}
        \label{fig:token_stats}
    \end{subfigure}
    \hfill
    \begin{subfigure}[t]{0.3\textwidth}
        \centering
        \includegraphics[width=\textwidth]{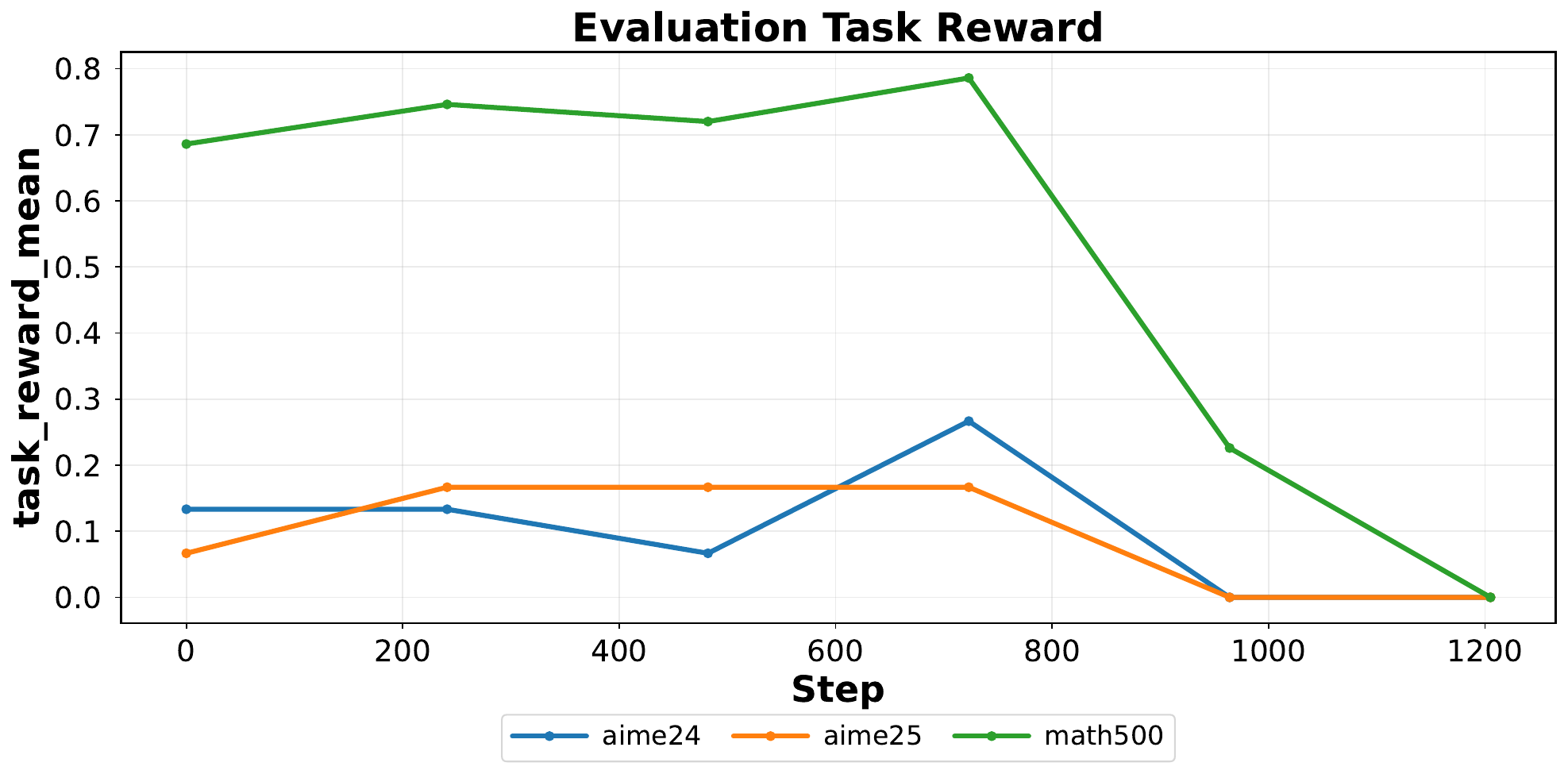}
        \caption{eval acc}
        \label{fig:eval_acc}
    \end{subfigure}
    
    \caption{
    Collapse under unnormalized Top-$20$ reverse KL. The model first
    becomes verbose, then degenerates into repetitive ``maybe'' outputs as response
    length reaches the limit and evaluation accuracy drops. Token statistics show that repetitive tokens dominate as the repeat ratio approaches one.
    }
    \label{fig:3opdcollapse}
\end{figure}

\paragraph{OPD.}We use a Qwen3-1.7B student and a Qwen3-8B teacher, both with thinking mode disabled, trained on OpenThoughts with an unnormalized Top20 reverse-KL objective (Figure~\ref{fig:3opdcollapse}). OPD initially improves but later collapses. Around step 700, rollout length and revision tokens such as ``wait'' and ``maybe'' increase sharply; by step 1000, the model degenerates into repetitive ``maybe'' outputs and accuracy on Math500, AIME24, and AIME25 drops nearly to zero. Token statistics show a corresponding abrupt rise in repetition and a collapse of teacher-student
overlap.

\subsection{Alignment \& System Prompt Internalization.}

\begin{figure}[t]
    \centering
    \begin{subfigure}[t]{0.48\textwidth}
        \centering
        \includegraphics[width=\textwidth]{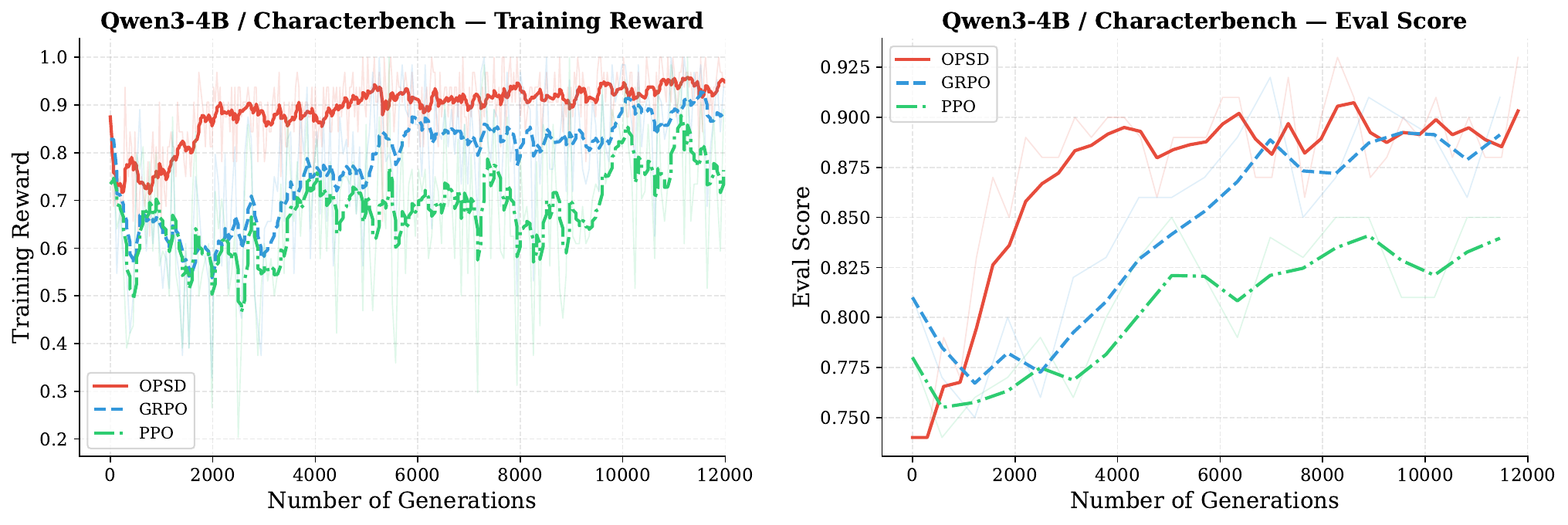}
        \label{fig:4b_char}
    \end{subfigure}
    \hfill
    \begin{subfigure}[t]{0.48\textwidth}
        \centering
        \includegraphics[width=\textwidth]{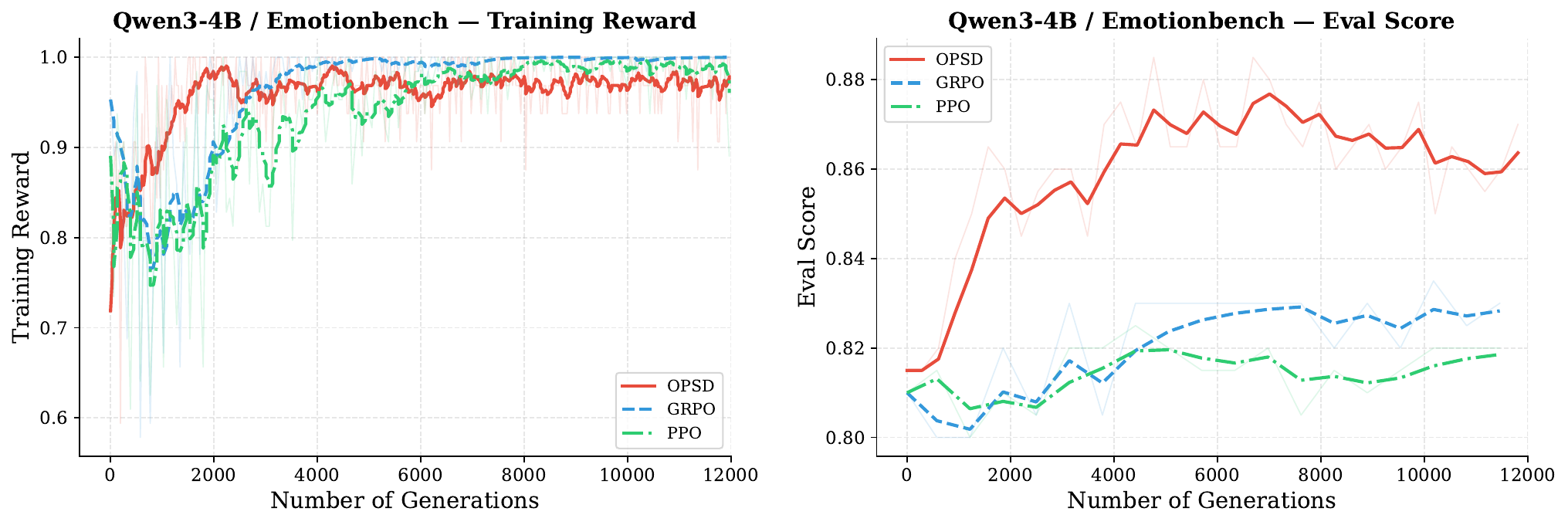}
        \label{fig:4b_emo}
    \end{subfigure}
    \vspace{-3mm}
    \caption{Training reward (left) and evaluation score (right) curves for OPSD, GRPO, and PPO on CharacterBench and EmotionBench, using Qwen3-4B-Instruct as student models.
    }
    \label{fig:style_alignment}
\end{figure}

\paragraph{Alignment.}We evaluate OPSD on two style-alignment benchmarks: CharacterBench~\cite{zhou2024characterbenchbenchmarkingcharactercustomization} and EmotionBench~\cite{huang2024emotionallynumbempatheticevaluating}.
For both tasks, we provide the teacher with a question-specific alignment prompt as privileged information; details are provided in Appendix~\ref{app:style_pi}.
We compare OPSD with GRPO and PPO using Qwen3-4B-Instruct and Qwen3-8B students.
As shown in Figure~\ref{fig:style_alignment}, under the same sampling budget, OPSD improves and converges faster than both RL baselines.

\begin{figure*}[t]
    \centering

    \begin{minipage}[t]{0.3\textwidth}
        \centering
        \includegraphics[width=\textwidth]{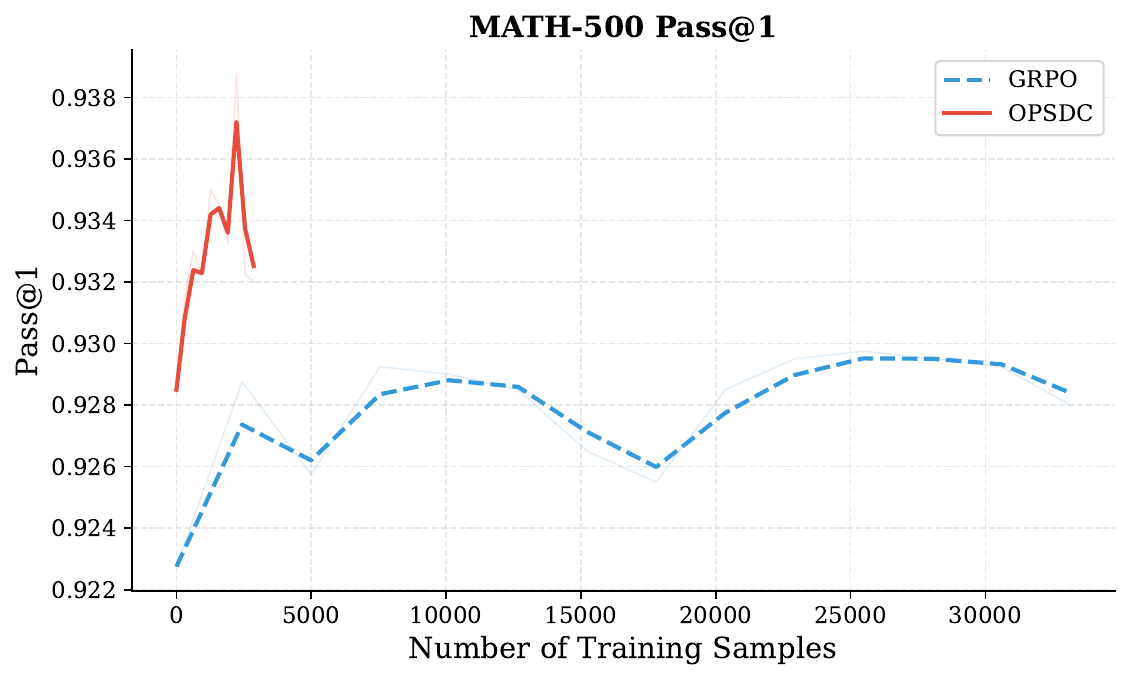}
    \end{minipage}
    \hfill
    \begin{minipage}[t]{0.3\textwidth}
        \centering
        \includegraphics[width=\textwidth]{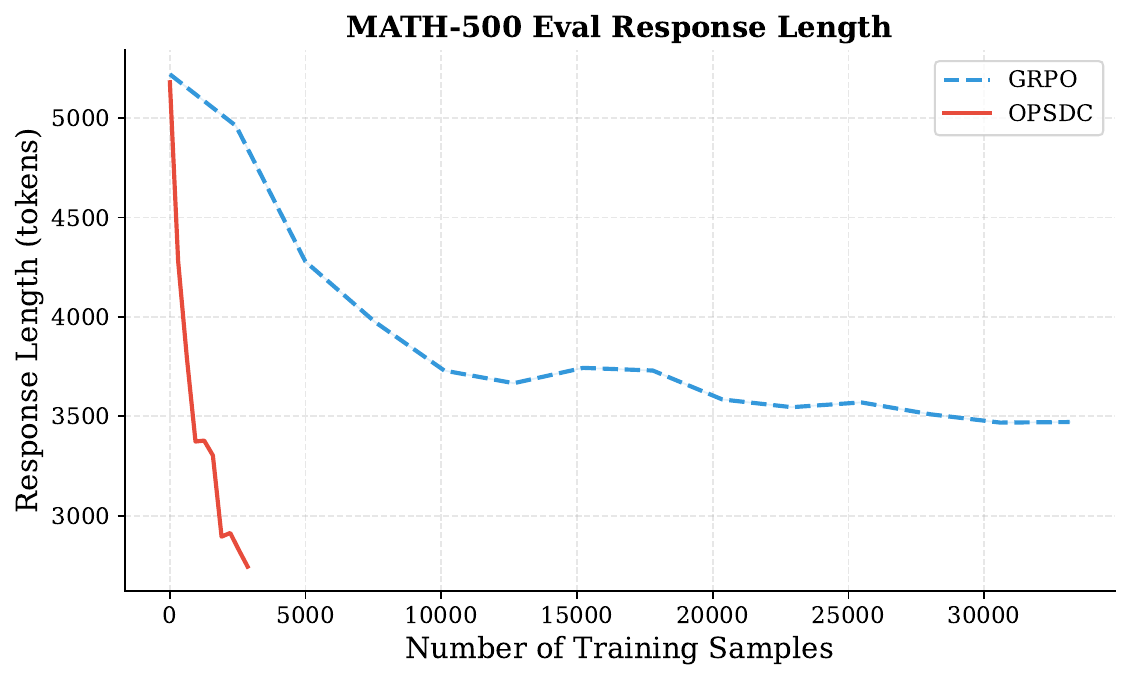}
    \end{minipage}
    \hfill
    \begin{minipage}[t]{0.3\textwidth}
        \centering
        \includegraphics[width=\textwidth]{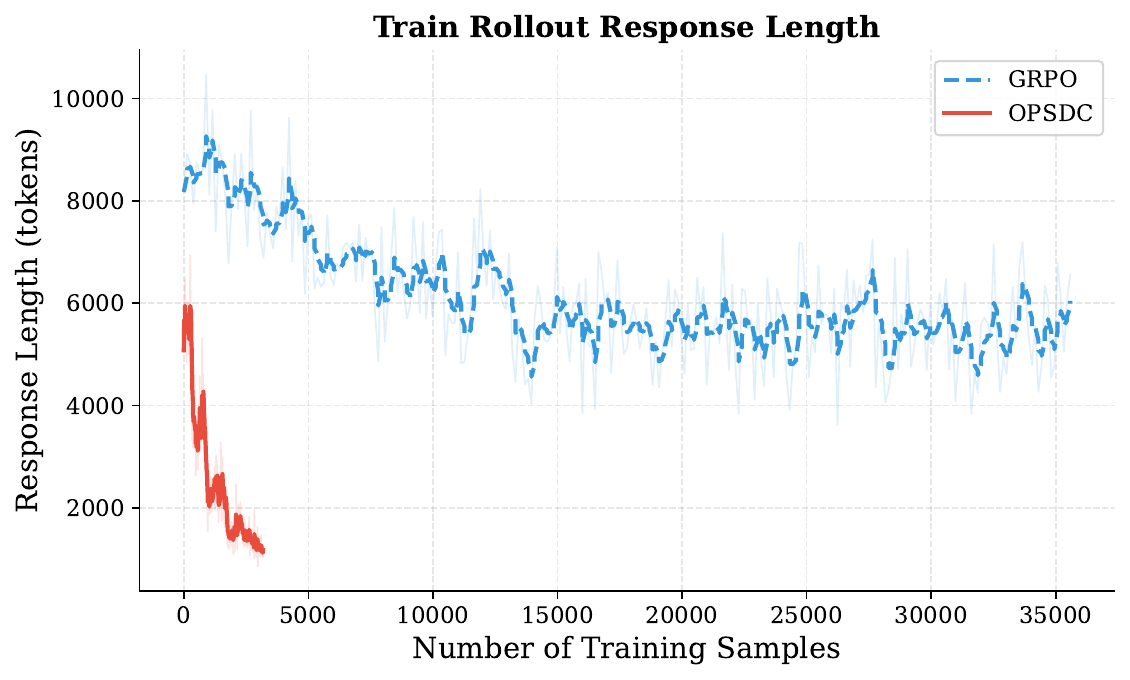}
    \end{minipage}
    \caption{Comparison of GRPO and OPSD on Qwen3-8B (thinking mode) trained with DAPO-Math-17k~\cite{yu2025dapoopensourcellmreinforcement}.
    From left to right: MATH-500 Pass@1, evaluation response length, and training rollout response length.}
    \label{fig:opsdc_vs_grpo}
\end{figure*}

\begin{figure*}[t]
    \centering
    \includegraphics[width=0.3\textwidth]{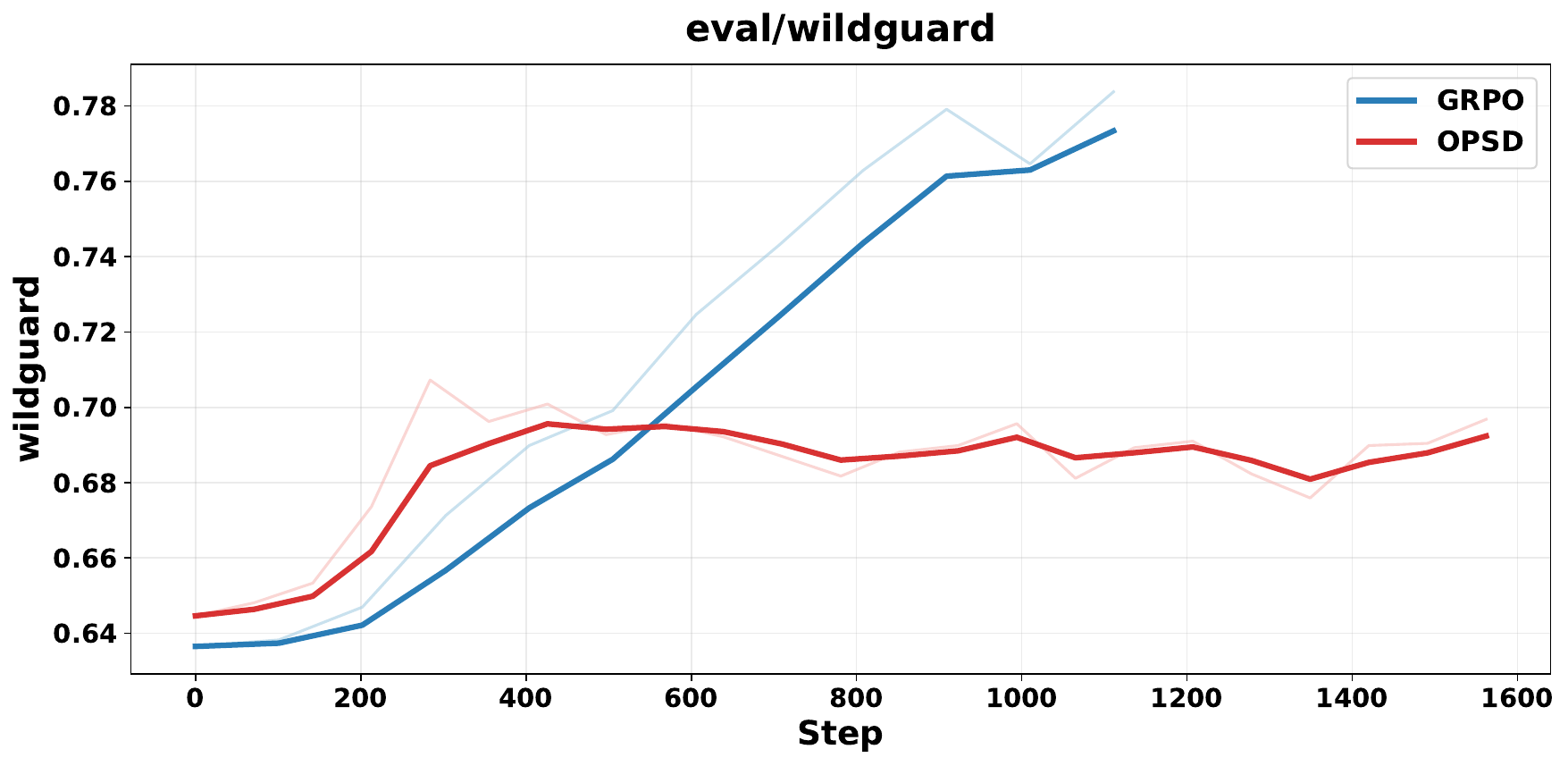}
    \hfill
    \includegraphics[width=0.3\textwidth]{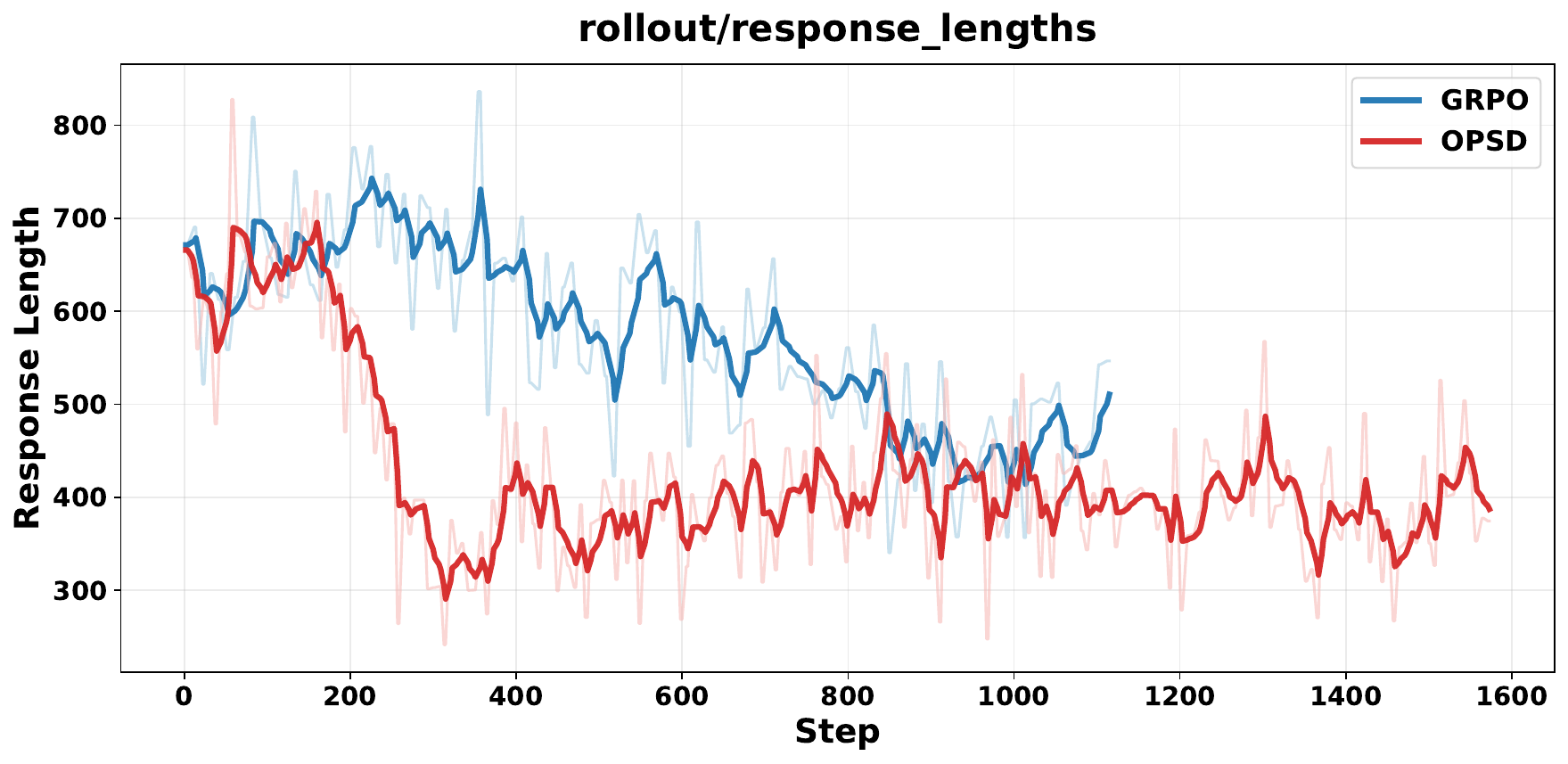}
    \hfill
    \includegraphics[width=0.3\textwidth]{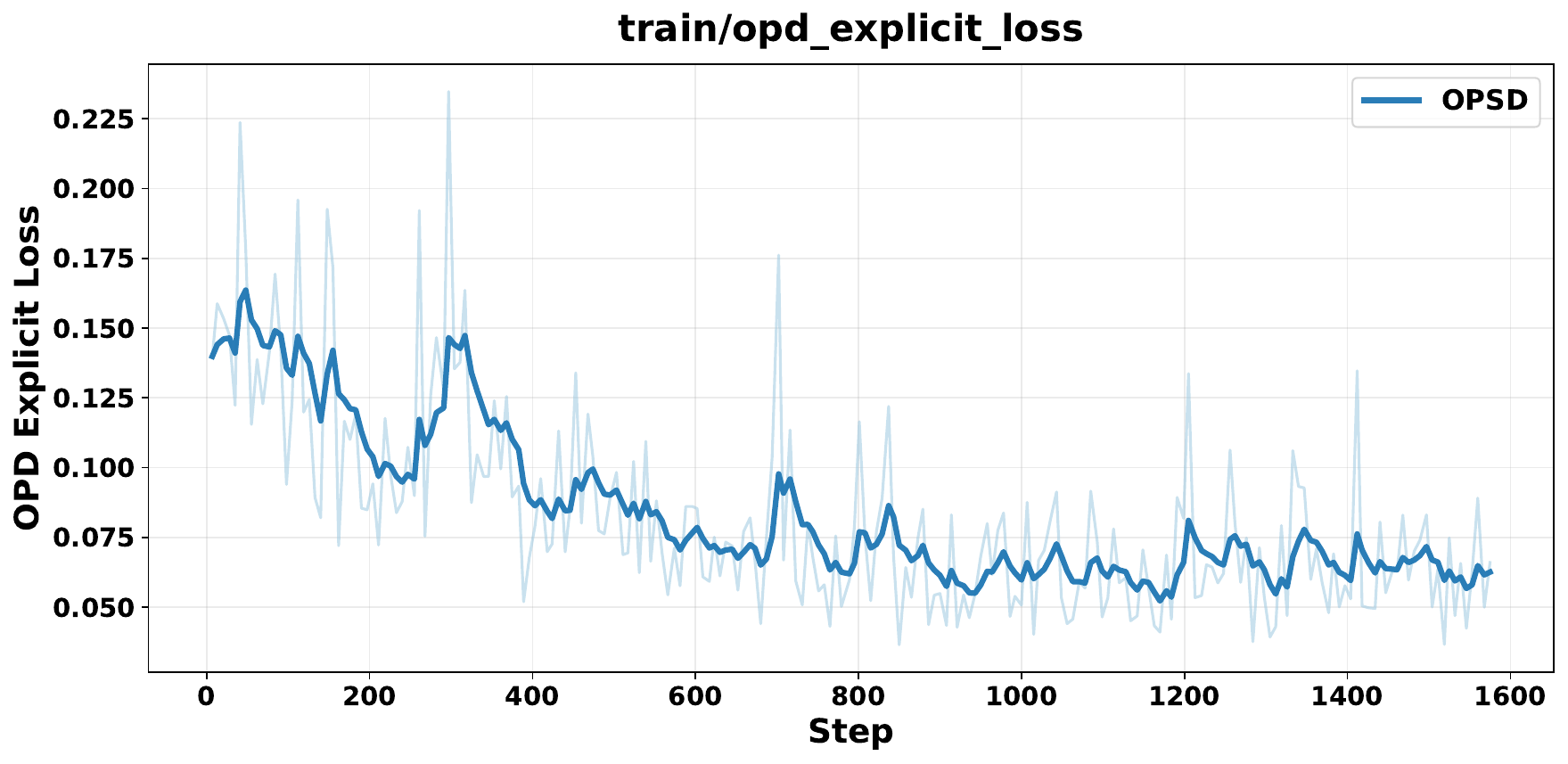}

    \caption{Train and evaluate Qwen3-1.7B (nothink) on Wildguardmix using their original train and test splits.}
    \label{fig:OPDonwildguard}
\end{figure*}

\paragraph{System Prompt Internalization.} We next study two applications of system prompt internalization: reasoning compression and safety alignment on adversarial questions. In contrast to the alignment tasks above, where the privileged information is question-specific, system prompt internalization uses a fixed system prompt as privileged information for all questions. For reasoning compression, shown in Figure~\ref{fig:opsdc_vs_grpo}, OPSD does not harm accuracy while substantially reducing response length. Compared with GRPO using a length penalty, OPSD is more sample efficient. It also offers a practical efficiency advantage: RL requires a large generation length to get a final answer, but OP(S)D can use shorter rollouts because the supervision is provided token-wise. However, we find that this application requires a sufficiently capable model, such as an 8B-scale model, to be effective. This finding is also consistent with \cite{sang2026crispcompressedreasoningiterative}. The second application is safety alignment. We conduct experiments with Qwen3-1.7B and Wildguardmix dataset \cite{han2024wildguardopenonestopmoderation} using the system prompt in Appendix~\ref{app:opsd_alignment_prompt}, with results shown in Figure~\ref{fig:OPDonwildguard}. OPSD improves rapidly in the early stage of training, but its final performance is constrained by the teacher. In contrast, GRPO improves more gradually but continues to make gains after OPSD saturates.

\section{Mechanism of On-Policy Distillation}

\subsection{Why Does OPD Sometimes Fail?}

\begin{figure}[t]
    \centering
    \begin{minipage}[t]{0.5\textwidth}
        \vspace{0pt}
        \centering
        \includegraphics[width=\textwidth]{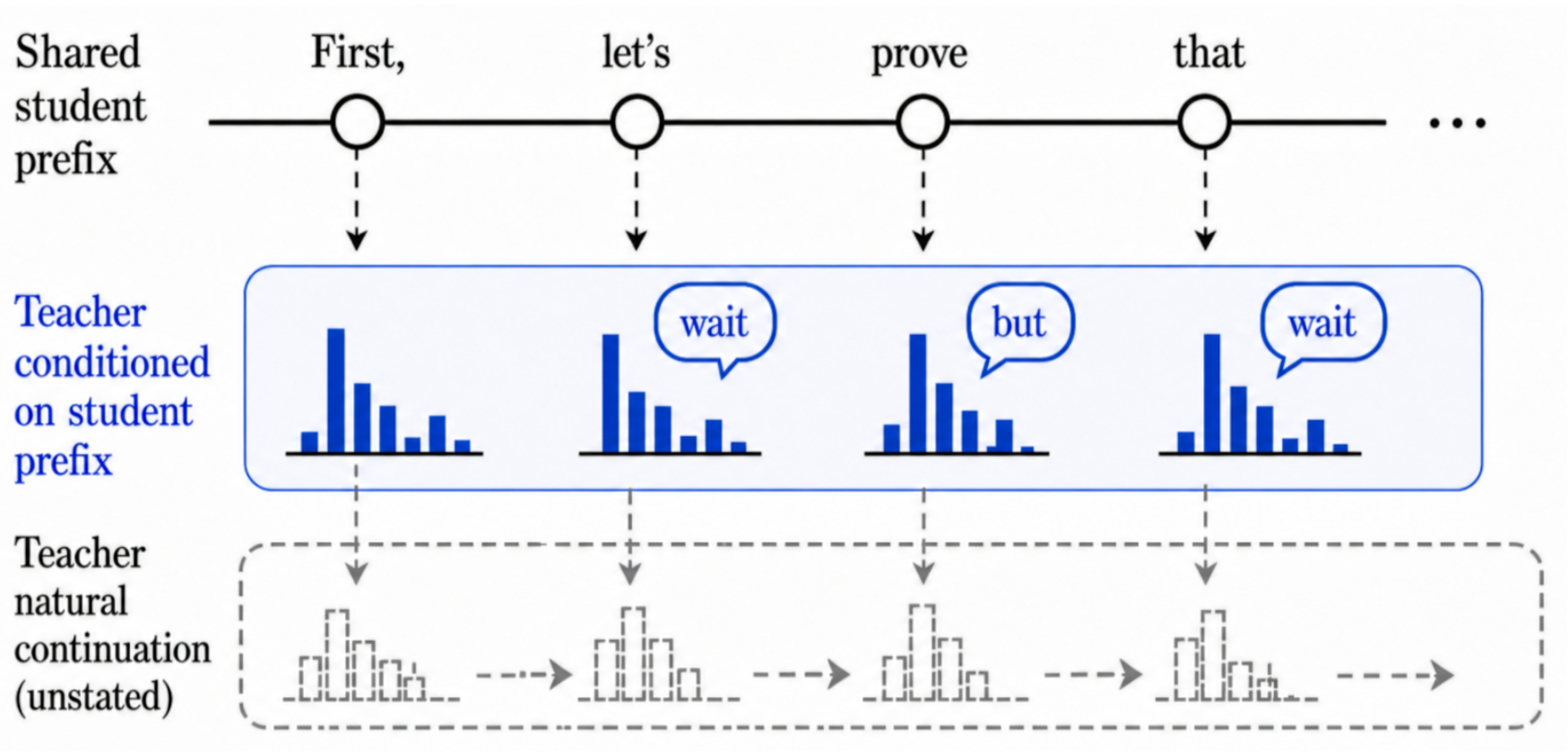}
    \end{minipage}
    \begin{minipage}[t]{0.45\textwidth}
        \vspace{0pt}
        \caption{Local semantic conflict in OPD. If the teacher's preferred
        path is inconsistent with the student prefix, teacher may encourage branch switching rather than branch refinement. This often
        appears as high probability assigned to revision tokens such as
        \textbf{``wait''} and \textbf{``but''}, which attempt to redirect the
        trajectory.}
        \label{fig:semantic_conflict_example}
    \end{minipage}
\end{figure}

\paragraph{Student Prefixes Distort the Teacher State.}
A central challenge in OPD is that the teacher is conditioned on
student-generated prefixes rather than its own trajectory. Although a strong
teacher may solve the problem independently, a partial student trajectory can
force it into an intermediate reasoning state that it would not have reached
on its own. To quantify this effect, we conduct a prefix-conditioned teacher
evaluation on GPQA-Diamond, using Qwen3-1.7B as the student and Qwen3-14B as
the teacher. The standalone teacher achieves $\textbf{62.1\%}$ accuracy, whereas
the student-prefix-conditioned teacher drops to $\textbf{46.0\%}$ accuracy
(details in Appendix~\ref{app:actionspace}). Thus, OPD supervision can become
weaker than the teacher's standalone capability if the teacher continues from student prefixes. Figure~\ref{fig:semantic_conflict_example} illustrates this failure mode at the
token level. Given a student prefix that has already committed to one reasoning
branch, the teacher assigns high probability to tokens such as ``wait'' and
``but,'' which revise or redirect the trajectory rather than extend it. This
creates a local semantic conflict between the student's current path and the
teacher's preferred path. In math reasoning, such unreliable supervision encourages verbose and inconsistent
reasoning.

\paragraph{The pitfall of unnormalized TopK Reverse KL.}
\label{curse_topk}

To reduce the GPU memory peak of reverse-KL distillation, we usually use the teacher's or the student's TopK set to compute a surrogate OPD loss. However, this may change the optimization stability and lead to model collapse in Figure \ref{fig:3opdcollapse}. Consider the full-vocabulary reverse KL in Equation \ref{eq:fullreverseKL}. A TopK truncation keeps only a subset $S_K(y_{<t})\subset\mathcal V$:
\begin{equation}
\mathcal L_{\mathrm{Top}\text{-}K\text{-}\mathrm{RKL}}(t)
=
\sum_{v\in S_K(y_{<t})}
\pi_S(v\mid x, y_{<t})\log\frac{\pi_S(v\mid x,y_{<t})}{\pi_T(v\mid x,y_{<t},I)}.
\label{eq:topkrevKL}
\end{equation}

The difference is in the gradient. For the full-vocabulary objective in Equation \ref{eq:fullreverseKL},
the constant term \textbf{+1} can be removed because
\begin{equation}
\sum_v p_\theta(v)\nabla_\theta\log p_\theta(v)  = 0
\end{equation}
However, this no longer holds for TopK:
\begin{equation}   
\sum_{v\in S_K(y_{<t})}\pi_S(v\mid x,y_{<t})\nabla_\theta \log \pi_S(v\mid x,y_{<t})
=
\nabla_\theta \sum_{v\in S_K(y_{<t})}\pi_S(v\mid x,y_{<t})
\neq 0.
\end{equation}
The \textbf{+1} term in the gradient remains:
\begin{equation} 
\nabla_\theta \mathcal L_{\mathrm{Top}\text{-}K\text{-}\mathrm{RKL}}(t)
=
\sum_{v\in S_K(y_{<t})}
\pi_S(v\mid x,y_{<t})
\left(
\log\frac{\pi_S(v\mid x,y_{<t})}{\pi_T(v\mid x,y_{<t},I)}\textbf{+1}
\right)
\nabla_\theta \log \pi_S(v\mid x,y_{<t}),
\end{equation}

Thus TopK truncation changes the update rule: a token is promoted only if
$
\pi_T(v \mid x,y_{<t},I) > e\,\pi_S(v \mid x,y_{<t}).
$
Teacher-preferred tokens with smaller teacher-student margins are still suppressed, biasing the student away from the teacher distribution and potentially toward unstable low-probability continuations.

\subsection{OPSD Learns a PI-Free Policy.}

\begin{figure}[t]
    \centering
\includegraphics[width=0.75\textwidth]{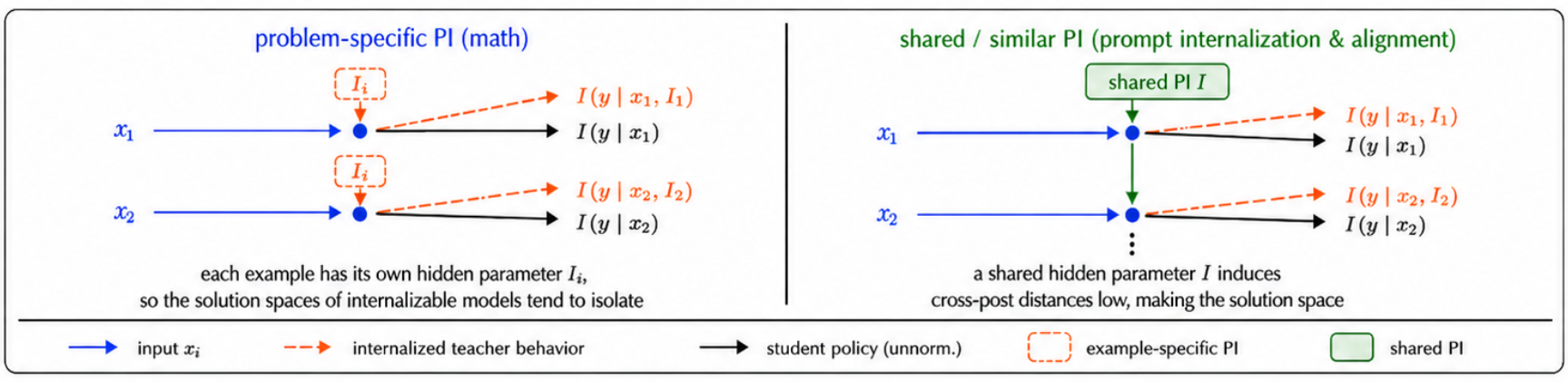}
    \caption{
    Effectiveness of OPSD depends on the structure of privileged information I.
    }
    \label{fig:8opsd}
\end{figure}

\begin{figure*}[t]
    \centering
    \includegraphics[width=0.33\textwidth]{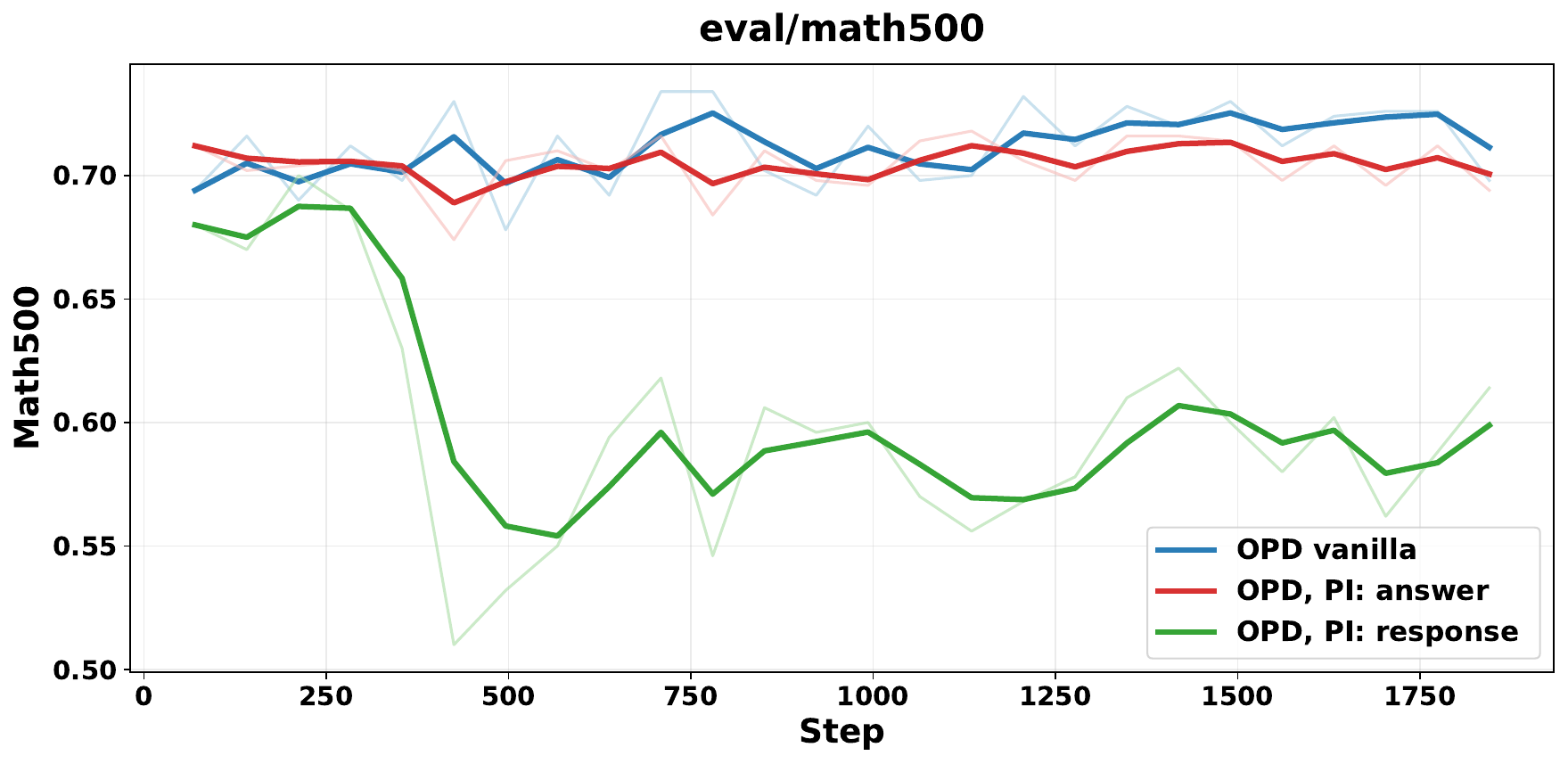}
    \hfill
    \includegraphics[width=0.33\textwidth]{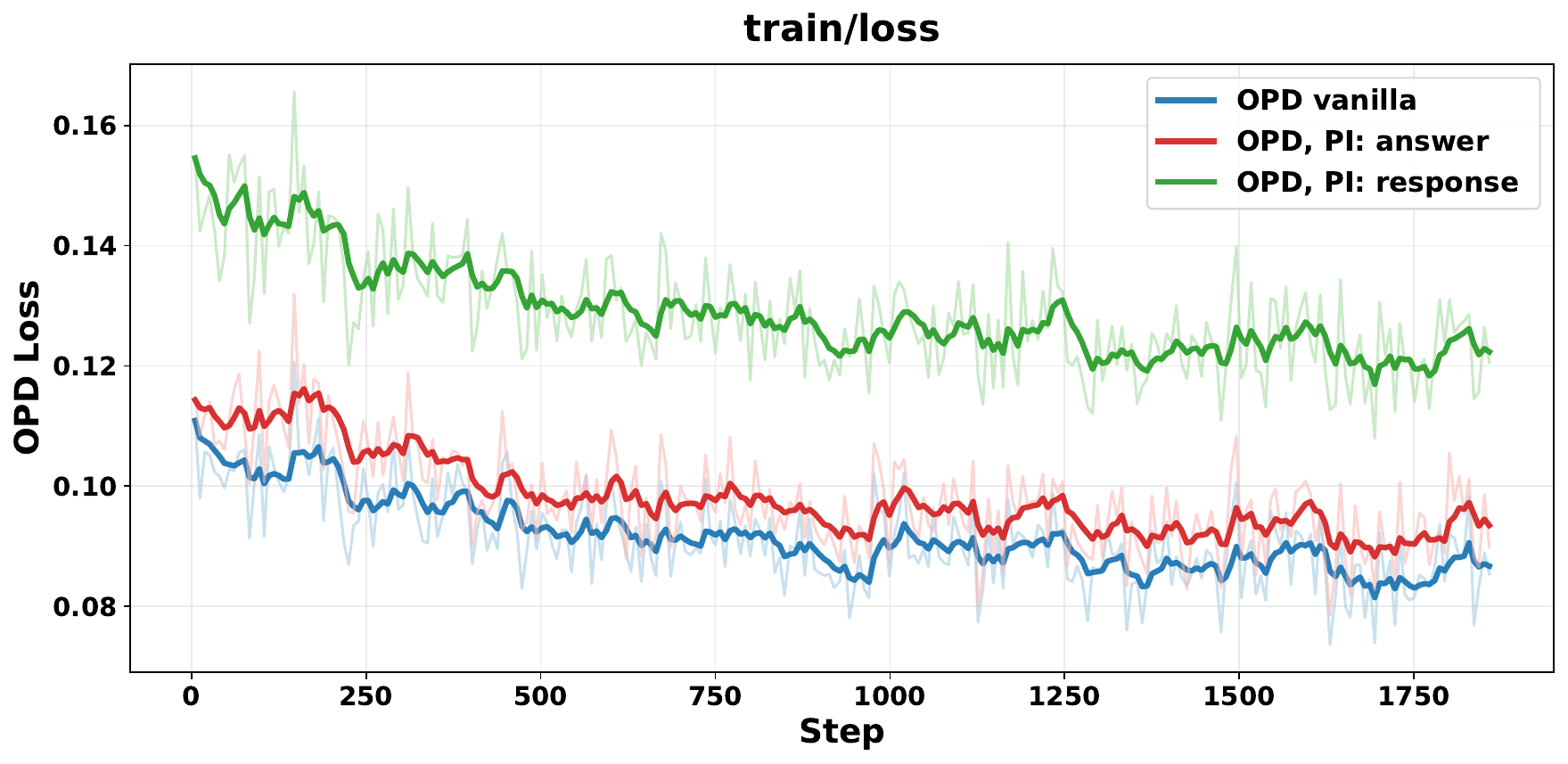}
    \hfill
    \includegraphics[width=0.32\textwidth]{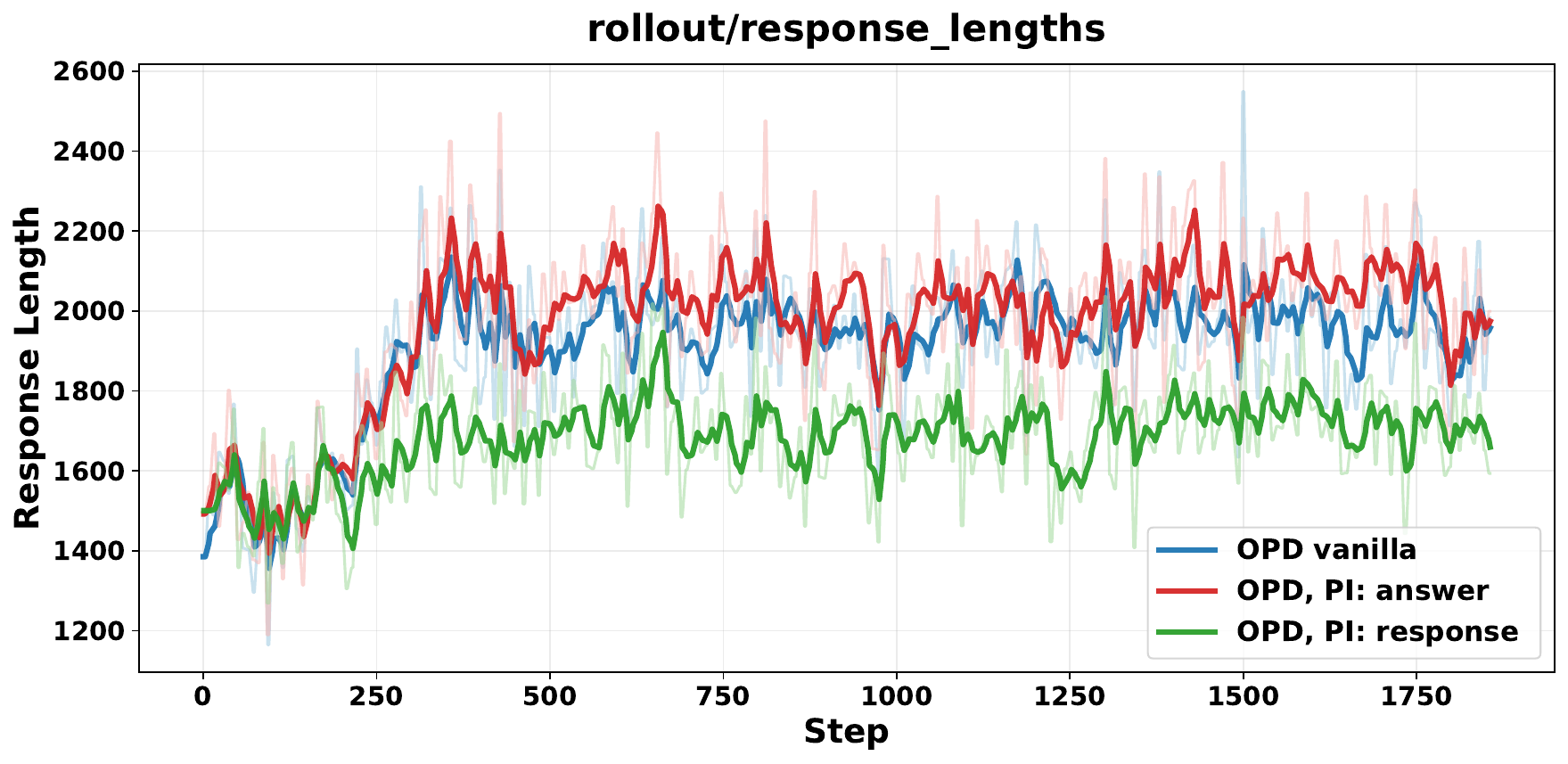}

\caption{
\textbf{PI does not improve OPD on math reasoning with a stronger teacher.}
Using a Qwen3-8B teacher and a Qwen3-1.7B student on OpenThoughts, both final-answer PI and full-response PI underperform vanilla OPD.
PI-conditioned OPD leads to higher KL loss.
}
    \label{fig:9OPDPI}
\end{figure*}

Let $x$ denote the question and $I$ denote the privileged information (PI).
Since the student does not observe $I$, OPSD optimizes
\begin{equation}
\min_{p_S}\;
\mathbb{E}_{(x,I)\sim\mathcal D}
\Big[
D_{\mathrm{KL}}\!\big(p_S(\cdot\mid x)\,\|\,p_T(\cdot\mid x,I)\big)
\Big].
\end{equation}
The optimal student is the normalized geometric mean of the PI-conditioned teacher distributions:
\begin{equation}
p_S^\star(y\mid x)
=
\frac{
\exp\left(
\mathbb{E}_{I\sim\mathcal D(\cdot\mid x)}
\log p_T(y\mid x,I)
\right)
}{
\sum_{y'}
\exp\left(
\mathbb{E}_{I\sim\mathcal D(\cdot\mid x)}
\log p_T(y'\mid x,I)
\right)
}.
\end{equation}
This form indicates that OPSD can distill behavior that is consistently
supported under different PI. Outputs that receive high probability under
some PI but low probability under others are suppressed. Figure~\ref{fig:8opsd} illustrates why the structure of PI ($I$) matters.
When PI is problem-specific, as in math reasoning, different examples depend
on different PI $I_1,I_2,I_3$. These parameters can
induce incompatible teacher behaviors, so the student is driven toward the common pattern of these incompatible teacher policies. This makes the student substantially weaker than the PI-conditioned teacher. In contrast, when
PI follows a similar latent structure, as in prompt internalization or
alignment, many examples are governed by the same underlying rule $I$. The
student can then compress the teacher's PI-conditioned behavior into a reusable
inductive bias.

To test whether math PI provides useful supervision, and to rule out limited
teacher capability as the cause of OPSD failure, we use a stronger Qwen3-8B
teacher to train a Qwen3-1.7B student with vanilla OPD, answer-PI OPD, and
response-PI OPD. As shown in Figure~\ref{fig:9OPDPI}, PI remains \textbf{unhelpful}:
both PI-conditioned methods underperform vanilla OPD, with response-PI
performing worst. Their distillation losses (at step 0) follow
\emph{OPD, PI: response} $>$ \emph{OPD, PI: answer} $>$ \emph{OPD vanilla},
suggesting that PI increases teacher-student mismatch. The higher loss of response-PI indicates that a longer PI induces a larger distributional
shift rather than a better distillation signal.

\section{Solutions}

\subsection{Designing an appropriate OPD loss.}

\begin{figure}[t]
    \centering
    \begin{subfigure}[t]{0.32\textwidth}
        \centering
        \includegraphics[width=\textwidth]{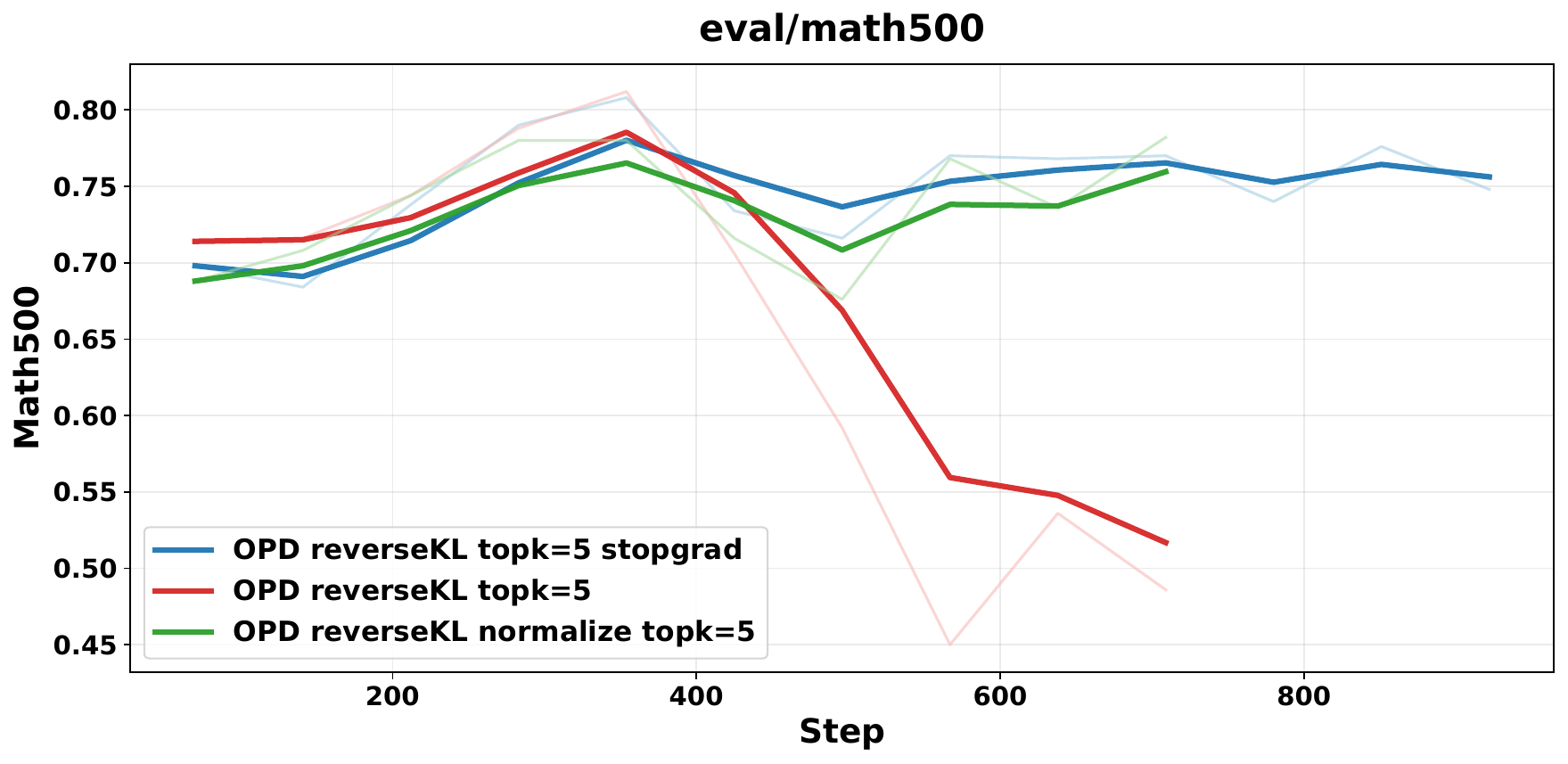}
        \label{fig:4b_char}
    \end{subfigure}
    \hfill
    \begin{subfigure}[t]{0.33\textwidth}
        \centering
        \includegraphics[width=\textwidth]{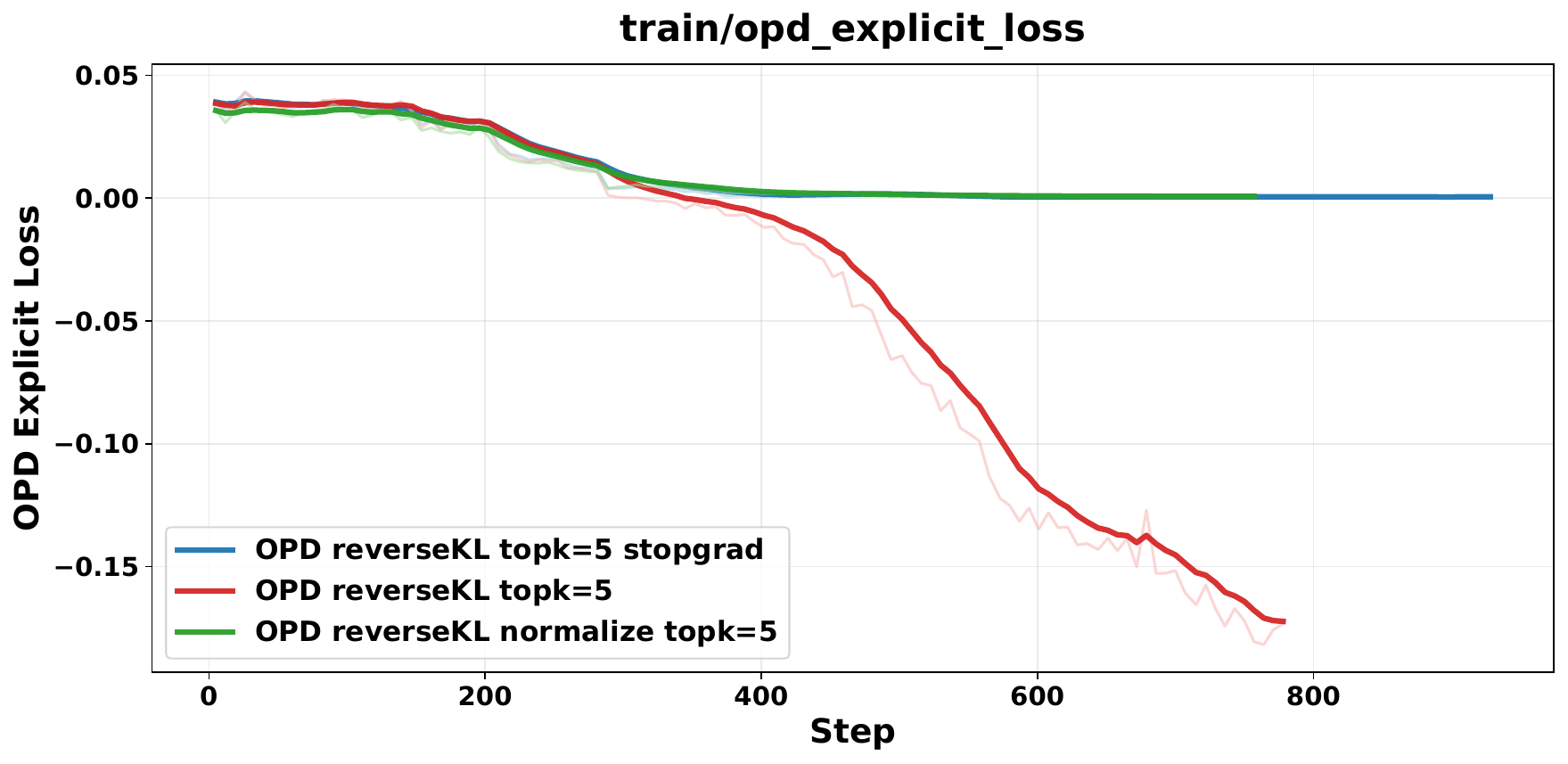}
        \label{fig:4b_emo}
    \end{subfigure}
    \hfill
    \begin{subfigure}[t]{0.33\textwidth}
        \centering
        \includegraphics[width=\textwidth]{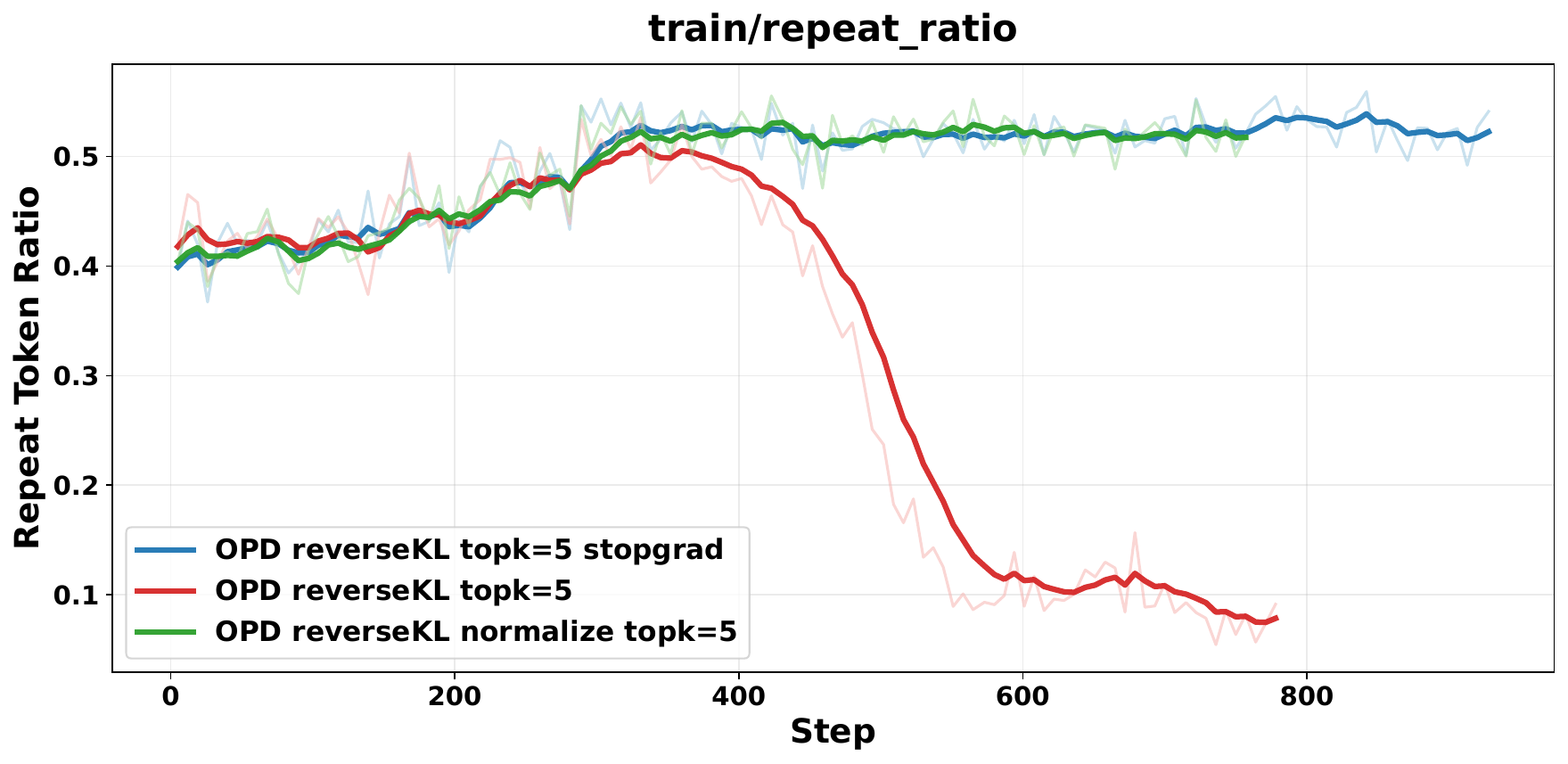}
        \label{fig:8b_char}
    \end{subfigure}
    \hfill
    
    \begin{subfigure}[t]{0.32\textwidth}
        \centering
        \includegraphics[width=\textwidth]{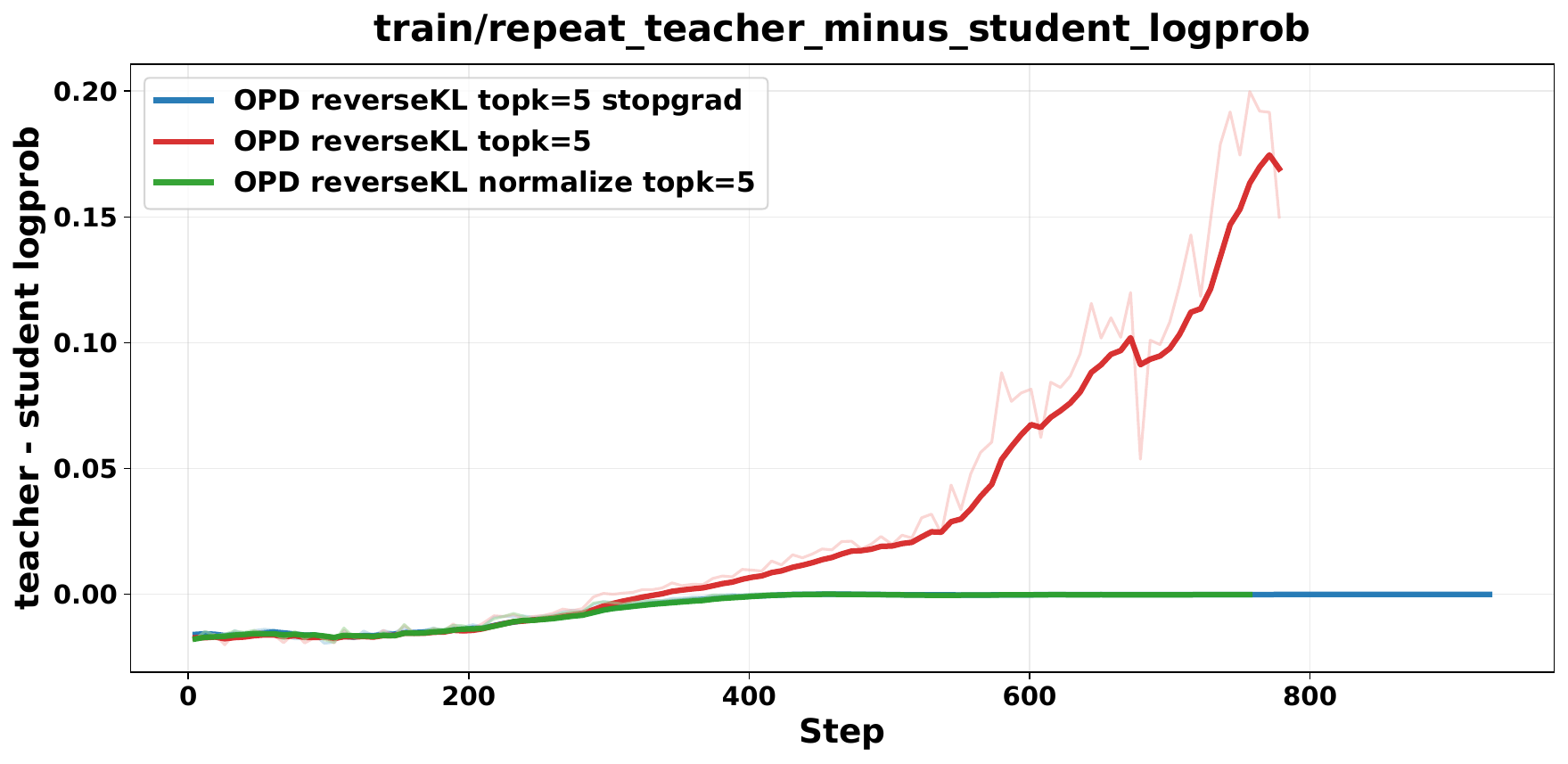}
        \label{fig:4b_char}
    \end{subfigure}
    \hfill
    \begin{subfigure}[t]{0.33\textwidth}
        \centering
        \includegraphics[width=\textwidth]{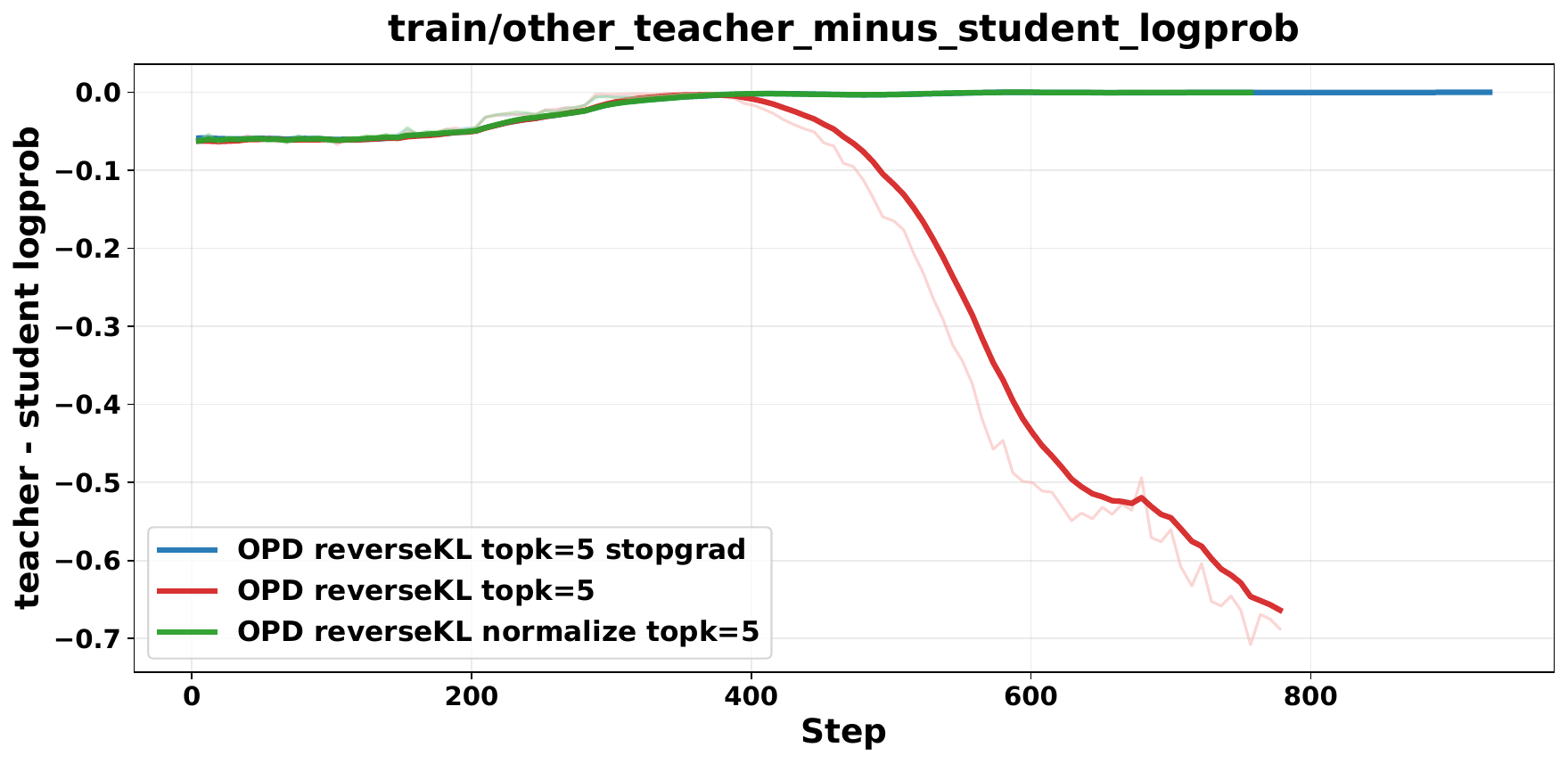}
        \label{fig:4b_emo}
    \end{subfigure}
    \hfill
    \begin{subfigure}[t]{0.33\textwidth}
        \centering
        \includegraphics[width=\textwidth]{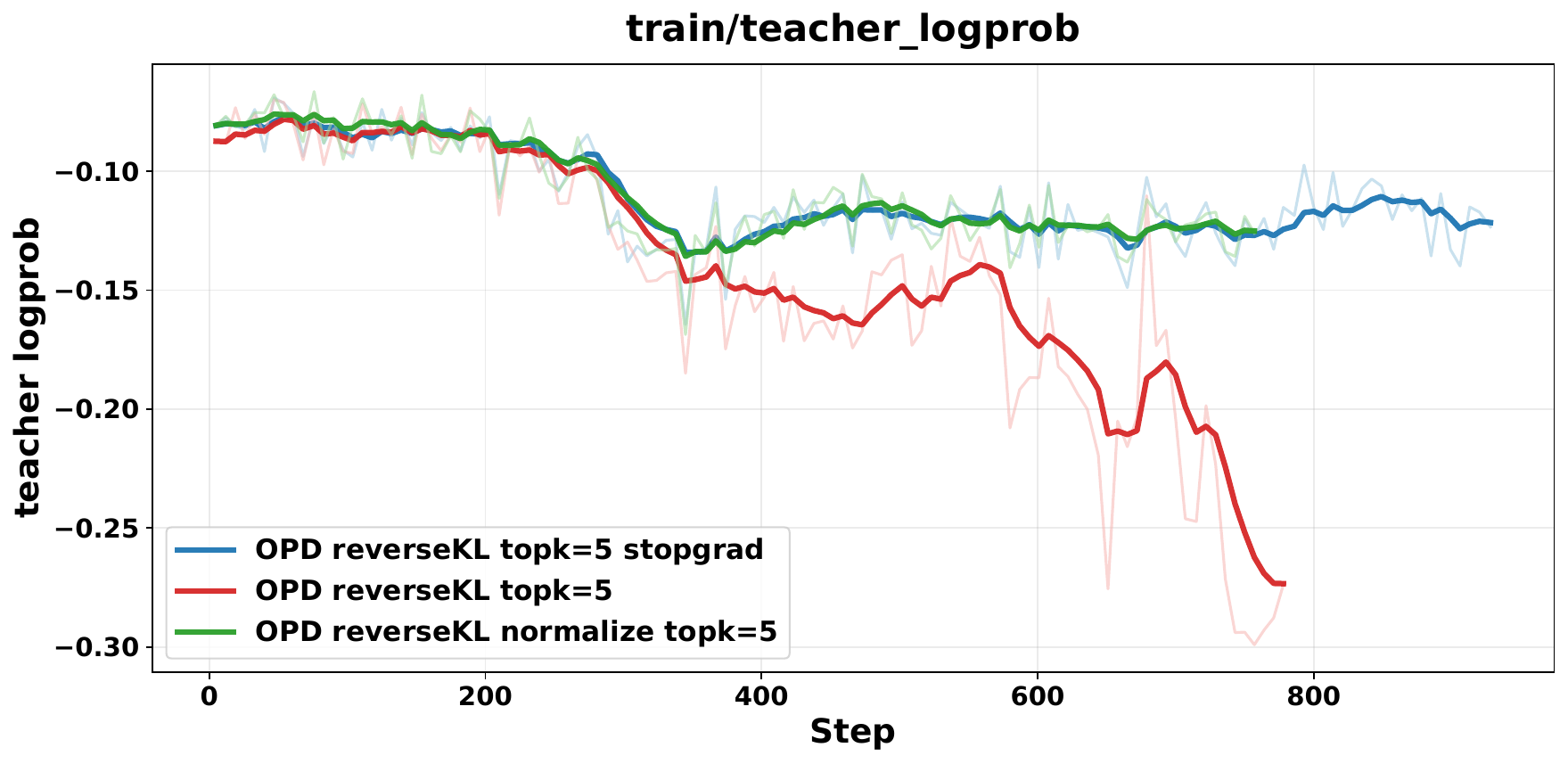}
        \label{fig:8b_emo}
    \end{subfigure}
    \vspace{-2mm}
    \caption{Teacher: Qwen3-1.7B-GRPO (nothink), Student: Qwen3-1.7B (nothink), DAPO, TopK=5.}
    \label{fig:top5exp}
\end{figure}

\begin{figure}[t]
    \centering
    \begin{subfigure}[t]{0.33\textwidth}
        \centering
        \includegraphics[width=\textwidth]{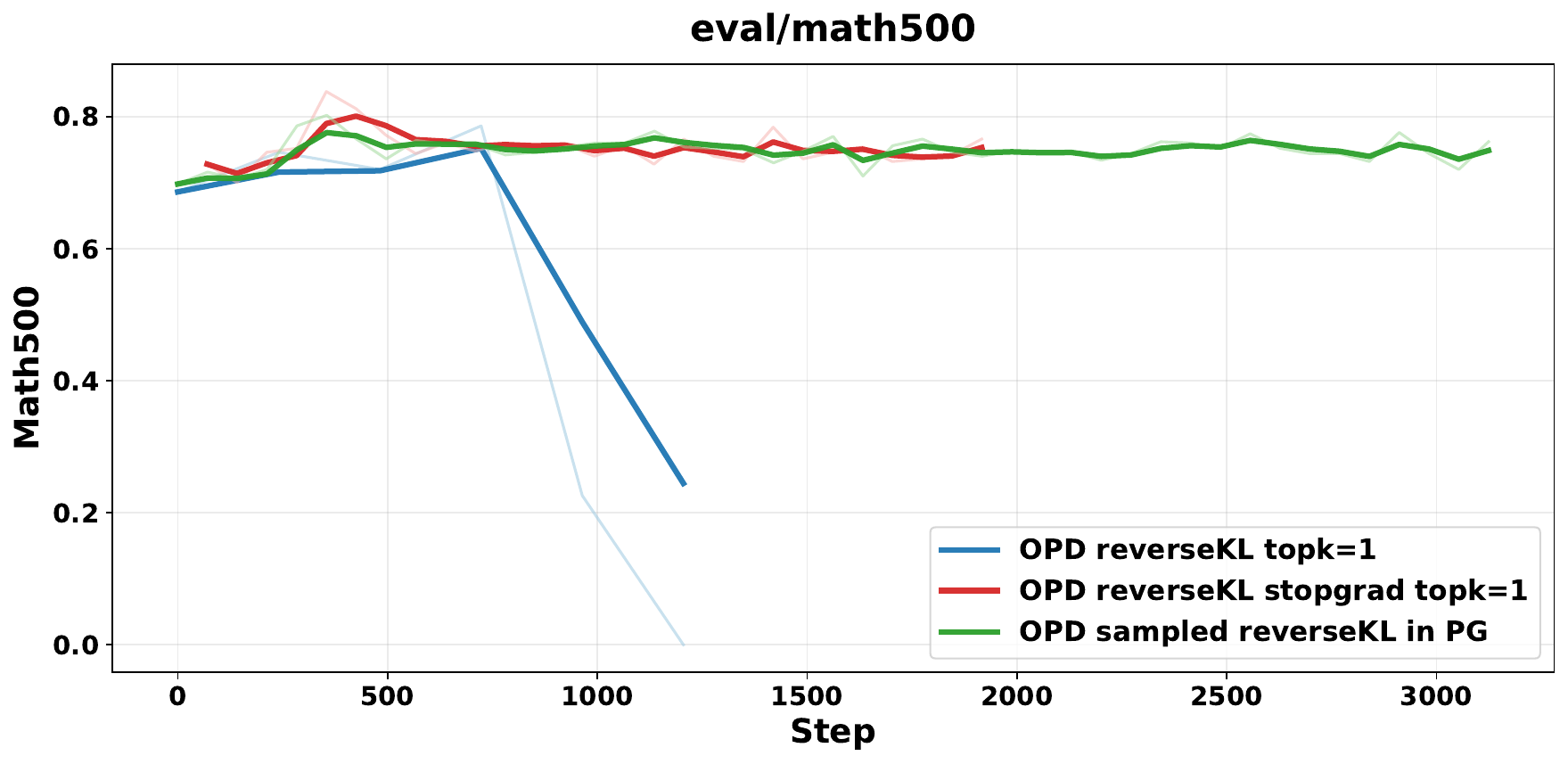}
        \label{fig:4b_char}
    \end{subfigure}
    \hfill
    \begin{subfigure}[t]{0.32\textwidth}
        \centering
        \includegraphics[width=\textwidth]{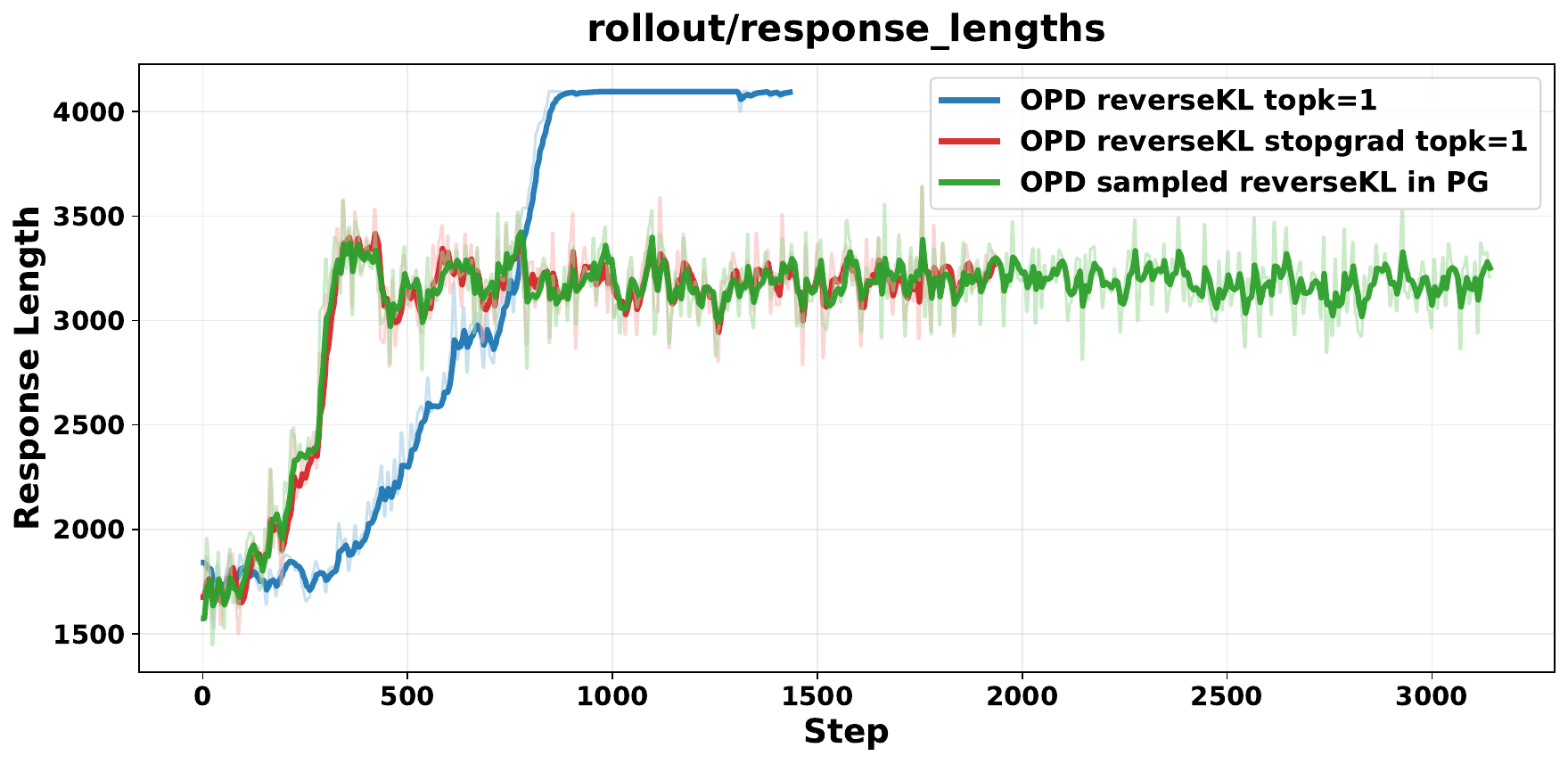}
        \label{fig:8b_char}
    \end{subfigure}
    \hfill
    \begin{subfigure}[t]{0.33\textwidth}
        \centering
        \includegraphics[width=\textwidth]{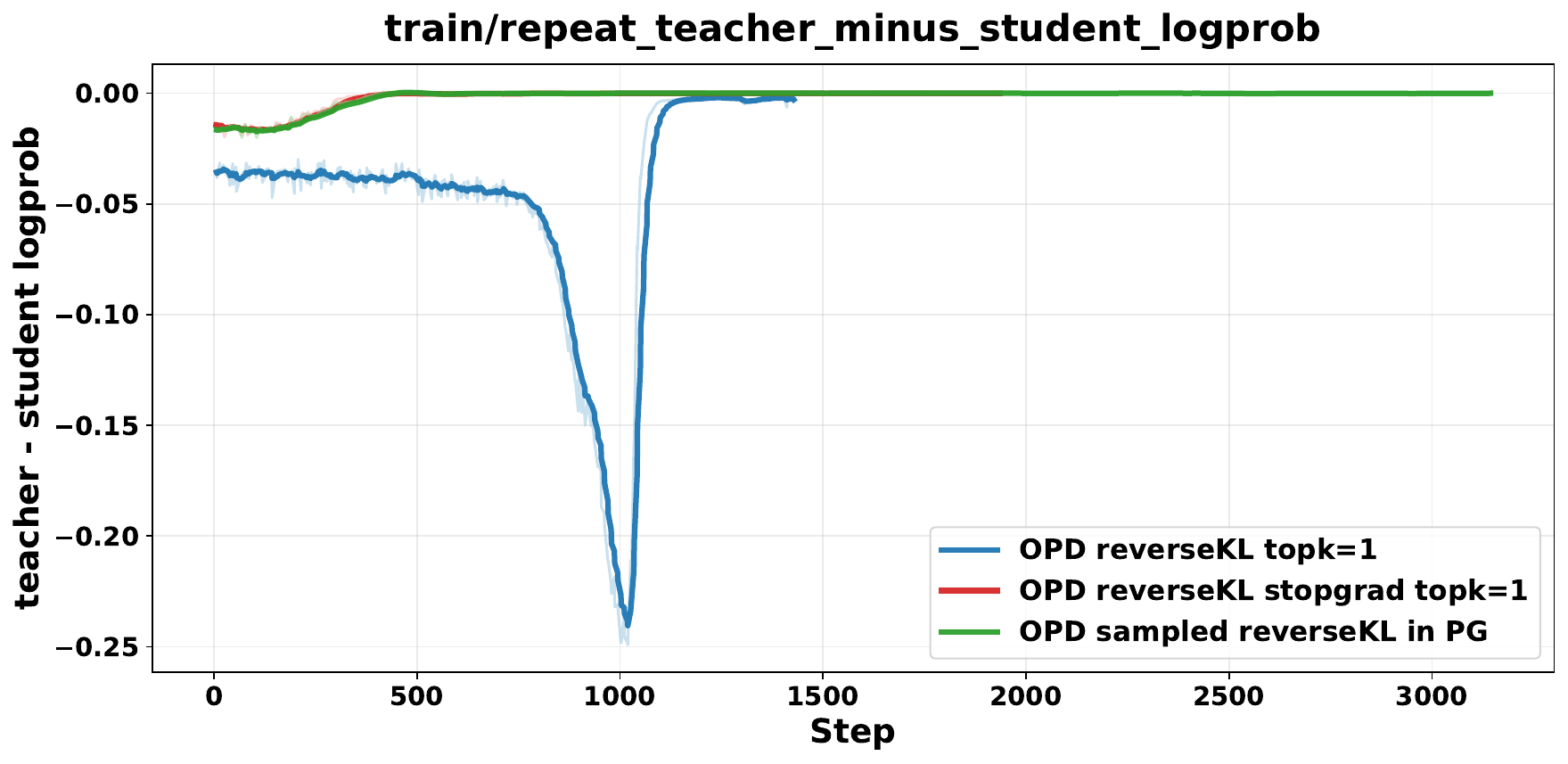}
        \label{fig:8b_emo}
    \end{subfigure}
    \vspace{-2mm}
    \caption{Whether to put the distillation loss in policy gradient? sampled token KL in policy gradient is stable, while putting sampled token KL as a single loss (TopK=1) leads to model collapse. Teacher: Qwen3-1.7B-GRPO, Student: Qwen3-1.7B, dataset: DAPO}
    \label{fig:policygradexp}
\end{figure}


To mitigate the pathology of TopK reverse KL, we replace exact TopK reverse-KL with a stop-gradient version. Specifically, instead of differentiating through both \(\pi_S\) and \(\log \pi_S\), we stop gradients of the log-probability term. The resulting objective is
\begin{equation}
\mathcal L_{\mathrm{SG\text{-}Top}K}(t)
=
-\sum_{v\in S_K(y_{<t})} \pi_S(v\mid x,y_{<t})\, [\log \pi_T(v\mid x,y_{<t},I)-\operatorname{\textbf{stopgrad}}\!\bigl(\log \pi_S(v\mid x,y_{<t})\bigr)\bigr],
\label{eq:topksgKL}
\end{equation}
with gradient $\nabla_\theta \mathcal L_{\mathrm{SG\text{-}Top}K}(t)$:
\begin{equation}
-\sum_{v\in S_K(y_{<t})}
\pi_S(v\mid x,y_{<t})\, [\log \pi_T(v\mid x,y_{<t},I)-\!\log \pi_S(v\mid x,y_{<t})\bigr]\,\nabla_\theta \log \pi_S(v\mid x,y_{<t}).
\end{equation}
Although this objective is not the exact gradient of reverse KL, it is better aligned with the policy-gradient view of on-policy distillation, where the weighting term should act as an advantage rather than a differentiable part of the loss.

By contrast, this issue is much less severe for two alternative formulations. The first is the TopK reverse KL with renormalization. Let
\begin{equation}
\bar \pi_S
=
\frac{\pi_S(v\mid x,y{<t})}
{\sum_{u\in S_K(y{<t})} \pi_S(u\mid x,y{<t})},
\bar \pi_T
=
\frac{\pi_T(v\mid x,y{<t},I)}
{\sum_{u\in S_K(y{<t})} \pi_T(u\mid x,y{<t},I)}, 
v\in S_K(x,y{<t}).
\end{equation}
The renormalized TopK reverse KL is
\begin{equation}
\mathcal L_{\mathrm{Renorm}\text{-}\mathrm{Top}\text{-}K\text{-}\mathrm{RKL}}(t)
=
\sum_{v\in S_K(y{<t})}
\bar\pi_S(v\mid x,y{<t})
\log
\frac{\bar\pi_S(v\mid x,y{<t})}
{\bar\pi_T(v\mid x,y{<t},I)} .
\label{eq:renormtopkKL}
\end{equation}
Since distributions are normalized within the TopK set, the
constant $+1$ term cancels:
\begin{equation}
\sum_{v\in S_K(y_{<t})}
\bar\pi_S(v\mid x,y_{<t})
\nabla_\theta \log \bar\pi_S(v\mid x,y_{<t})
=
\nabla_\theta
\sum_{v\in S_K(y_{<t})}
\bar\pi_S(v\mid x,y_{<t})
=0.
\end{equation}
The gradient does not contain the $+1$ that causes the
unnormalized TopK reverse-KL bias in paragraph~\ref{curse_topk}. However, it only matches the relative probabilities
inside the selected TopK set and ignores the probability mass assigned
to that set. As a result, it mitigates the local gradient bias but no longer
faithfully approximates the full-vocabulary reverse KL.

A second alternative is to move the sampled-token signal into the policy-gradient (Equation\ref{eq:policygrad_sampledtoken}). This formulation also avoids the TopK reverse-KL issue. The drawback is that this objective uses only the sampled token \(y_t\). It therefore captures sparse teacher signal rather than the teacher vocabulary distribution.

\textbf{Another important design choice is the TopK truncation}. Ideally, we would like to query the teacher on the student's per-position TopK set $S_{\mathrm{stu},K}(y_{<t})$, or vice versa. 
However, this is not directly supported when using SGLang as inference engine (details in appendix \ref{topkchallenge}). 

Instead, we must take the union of all per-position TopK sets
$
U=\bigcup_{t=1}^T S_{\mathrm{stu},K}(y_{<t}),
$
and query this same global set at every position. If the response length is $T$, the ideal cost is $T\times K$, but the actual queried tensor becomes $T\times |U|$. 
In the worst case, the TopK sets at different positions are disjoint, so
$|U|=\min(|\mathcal V|,TK)$,
which increases memory by a factor of $\frac{\min(|\mathcal V|,TK)}{K}$. As a simple fix, we backpropagate only through tokens that appear in both models' TopK sets:
\begin{equation}
S_{K}(y_{<t}) = S_{\mathrm{tea},K}(y_{<t})\cap S_{\mathrm{stu},K}(y_{<t}) 
\end{equation}
Interestingly, prior work~\cite{li2026rethinkingonpolicydistillationlarge} finds that the teacher-student TopK intersection performs comparably to the student TopK, suggesting that our design does not degrade training effectiveness.

We compare three TopK objectives with $K=5$ in
Figure~\ref{fig:top5exp}: unnormalized reverse KL
(Equation~\ref{eq:topkrevKL}), its stop-gradient version
(Equation~\ref{eq:topksgKL}), and its renormalized version
(Equation~\ref{eq:renormtopkKL}). The unnormalized objective collapses during
training, whereas both modifications achieve comparable stable performance. We further test the formulation that places the distillation
signal within the policy gradient
(Equation~\ref{eq:policygrad_sampledtoken}). As shown in
Figure~\ref{fig:policygradexp}, it performs similarly to Top1 reverse KL with
stop-gradient, while the unmodified Top1 reverse KL still collapses. These
results suggest that correcting the biased reverse-KL gradient, through
stop-gradient, renormalization, or a policy-gradient formulation, is essential
for stable distillation.



\subsection{Enhance teacher with RLVR.}

\begin{figure}[t]
    \centering
    \hfill
    \begin{subfigure}[t]{0.3\textwidth}
        \centering
        \includegraphics[width=\textwidth]{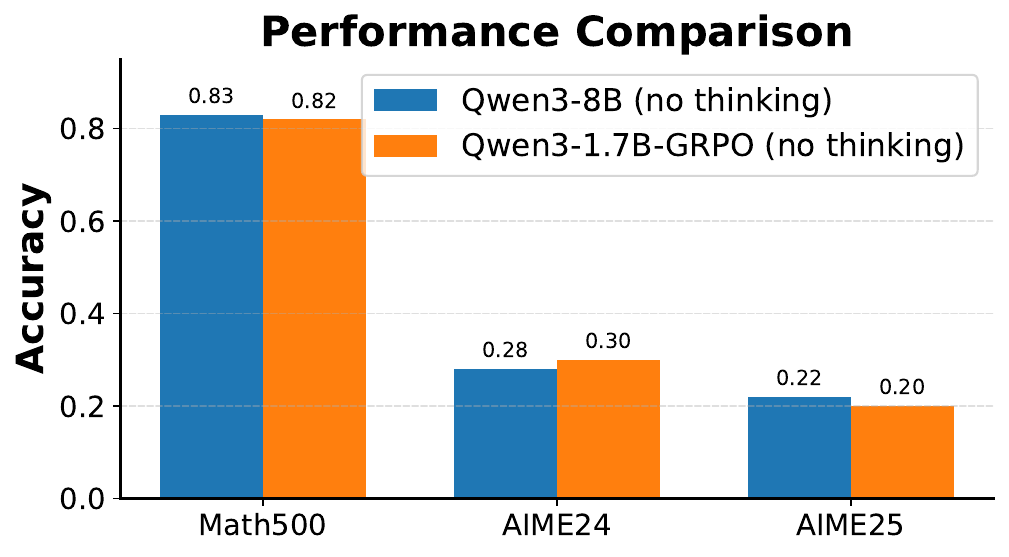}
        \label{fig:second}
    \end{subfigure}
    \begin{subfigure}[t]{0.3\textwidth}
        \centering
        \includegraphics[width=\textwidth]{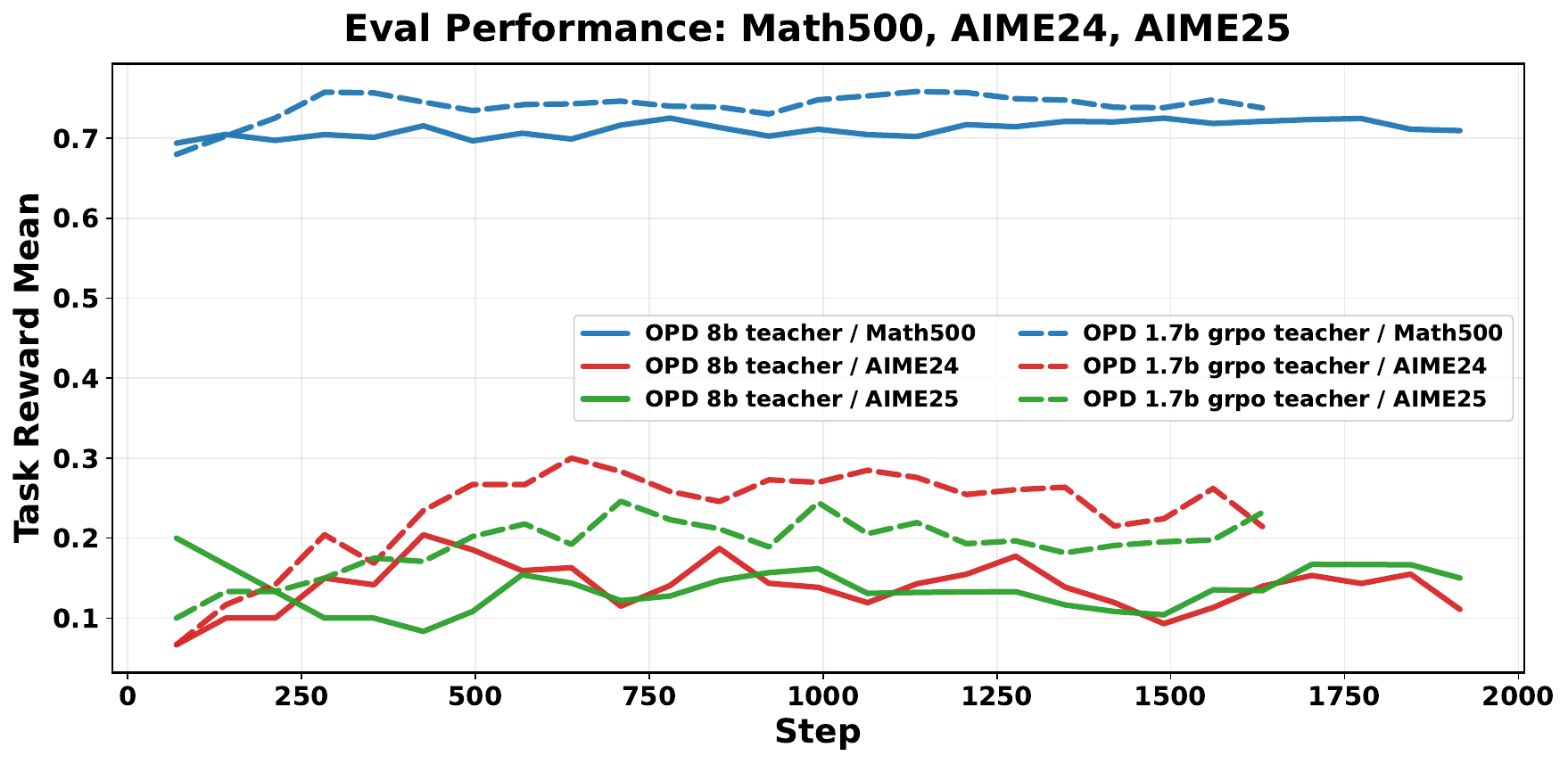}
        \label{fig:first}
    \end{subfigure}
    \hfill
    \begin{subfigure}[t]{0.3\textwidth}
        \centering
        \includegraphics[width=\textwidth]{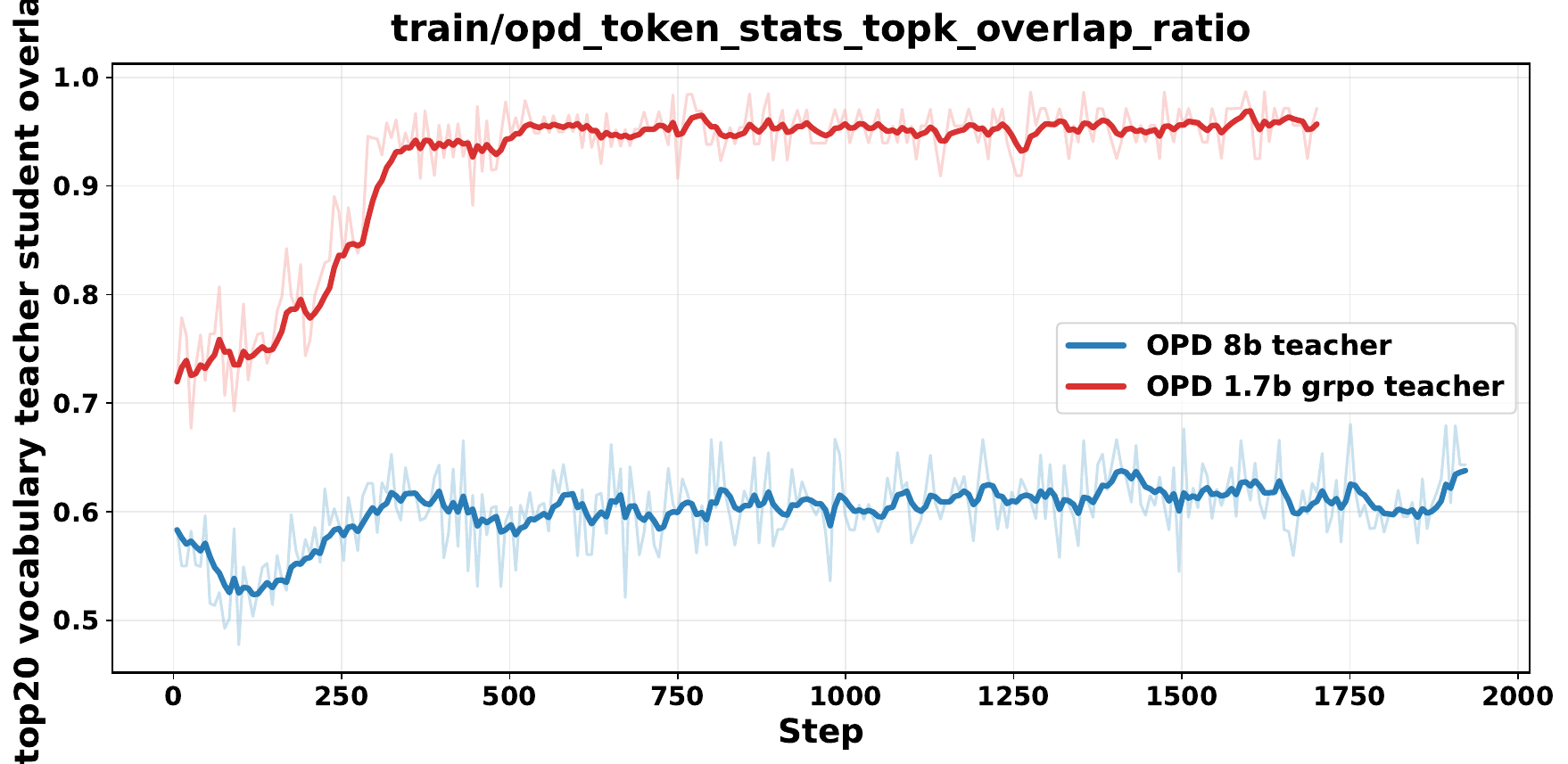}
        \label{fig:second}
    \end{subfigure}
    
    \vspace{-3mm}
    \caption{dataset: OpenThoughts. \textbf{Left}: Qwen3-8B and Qwen3-1.7B-GRPO have similar math reasoning performance. \textbf{Middle}: In OPD, Qwen3-1.7B-GRPO is a more effective teacher. \textbf{Right}: Qwen3-1.7B-GRPO's Top20 vocabulary distribution is more aligned with the Qwen3-1.7B student.}
    \label{fig:opsdrlvr}
\end{figure}

One effective way to improve OPD is to adapt the teacher on the training
distribution before distillation. RLVR improves the teacher's task
performance on the training set and can make the teacher's
output distribution closer to that of the student. This reduces the
teacher-student distribution mismatch and makes token-level distillation signals
more locally compatible with the student's on-policy prefixes.

We evaluate this strategy in Figure\ref{fig:opsdrlvr}. We first train a Qwen3-1.7B model with
DAPO \cite{yu2025dapoopensourcellmreinforcement} for 200 steps, obtaining a RL-adapted teacher denoted
Qwen3-1.7B-GRPO. Qwen3-1.7B-GRPO achieves performance on
Math500, AIME24, and AIME25 that is comparable to Qwen3-8B. We then use either
Qwen3-1.7B-GRPO or Qwen3-8B as the teacher to distill a Qwen3-1.7B student on the
OpenThoughts \cite{guha2025openthoughtsdatarecipesreasoning} dataset. 

Although the two teachers have similar reasoning performance, distillation from Qwen3-1.7B-GRPO significantly outperforms
distillation from Qwen3-8B, suggesting that teacher accuracy alone is not
sufficient to predict OPD effectiveness. A teacher closer to the student distribution can provide stronger supervision, even if
it does not have stronger benchmark accuracy.
Compared with using PI during distillation (Figures \ref{fig:2opsdfails} and \ref{fig:9OPDPI}), this pipeline
provides a stronger teacher and serves as an effective alternative to PI-based OP(S)D.

\subsection{Stabilize Student with SFT.}

\begin{figure}[t]
    \centering
    \begin{subfigure}[t]{\textwidth}
        \centering
        \includegraphics[
            width=\linewidth,
            trim=0 0 0 10mm,
            clip
        ]{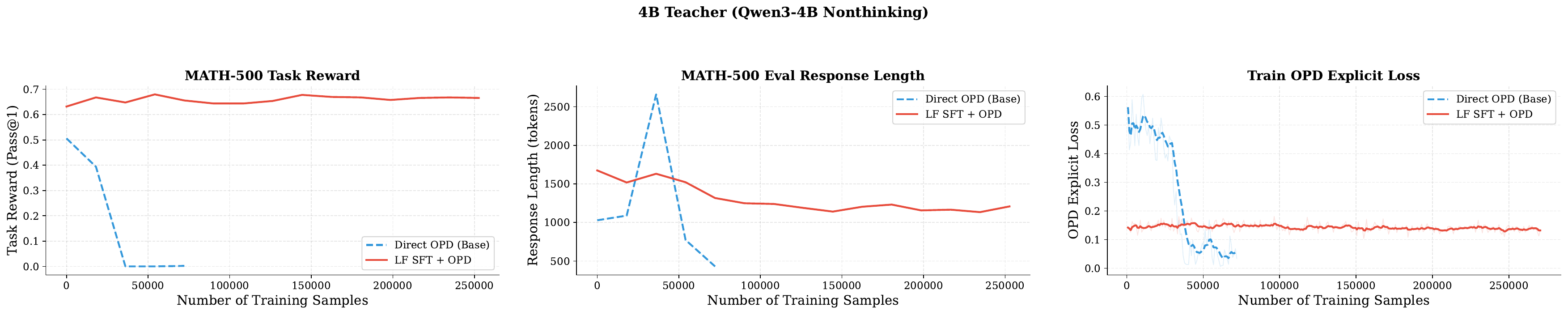}
        \label{fig:sft_opd_4b}
    \end{subfigure}
    \vspace{-3mm}
    \caption{
        Qwen3-4B teacher, Qwen3-1.7B-Base student, OpenThoughts. \textbf{Direct OPD} uses the student without SFT warm-up;
        \textbf{SFT + OPD} fine-tunes on teacher-generated traces before OPD.
    }
    \label{fig:sft_opd}
\end{figure}

Prior work has used supervised data to stabilize OPD. For example, Luo et al.~\cite{luo2026demystifying} combine OPD with SFT on gold answers to mitigate length inflation, while Li et al.~\cite{li2026rethinkingonpolicydistillationlarge} fine-tune the student on teacher-generated traces to reduce the teacher-student distribution mismatch. In the Qwen3-4B $\rightarrow$ Qwen3-1.7B-Base setting, we observe a distinct instability source: the 1.7B student initially produces garbled, nonsensical outputs (non-English Unicode) in some questions. Unlike ordinary incorrect reasoning, these outputs contain degenerate sequences with little semantic structure. In this case, teacher is unlikely to provide a meaningful signal, making Direct OPD unstable. As shown in Figure~\ref{fig:sft_opd}, teacher-trace SFT regularizes the student's output space, stabilizes response length, and allows subsequent OPD to improve accuracy. These results suggest that SFT helps OPD by keeping on-policy samples in well-formed regions where teacher feedback remains effective. Experiment details are in Appendix~\ref{app:experiment_setup_6.3}.
\section{Conclusion}

We present an empirical study of OPD and OPSD for LLMs, showing that the effectiveness depends on task structure and the nature of privileged information. OPSD is ineffective in our tested mathematical reasoning settings, where privileged information is largely instance-specific, but works well for system prompt internalization and alignment tasks where privileged information reflects shared latent rules. We identify three failure mechanisms: teacher-student distribution mismatch, biased TopK reverse-KL gradients, and the limitation of OPSD in learning only the marginal distribution over instance-specific privileged information. We further show that stop-gradient TopK surrogate loss, RLVR teacher improvement, and SFT stabilization can improve practical training stability. Our findings are based on a limited set of model families and model scales, and larger-scale settings may exhibit different behavior. Future work may explore iterative self-improvement pipelines that combine SFT, RL, and OPD: SFT provides a stable initialization, RL improves task-specific behavior, and OPD distills the improved on-policy behavior back into the student.

\newpage
\bibliographystyle{unsrt}
\bibliography{references}


\appendix

\section{Appendix}

\subsection{Experiment Setup}
\label{app:opd_hparams}

We use a maximum response length of $16384$, temperature $1.0$, and top-$p$ $0.95$ for evaluation.

Table~\ref{tab:train_hparams} summarizes the main hyperparameters for OPD, OPSD, GRPO, and PPO. Unless otherwise specified, OPD and OPSD use one rollout per prompt and optimize a stop-gradient, renormalized Top-$K$ reverse-KL objective. OPSD disables thinking mode and uses the step-0 student checkpoint as the fixed teacher model by default. All experiments are conducted on a server with 10 * NVIDIA RTX PRO 6000 Blackwell GPUs.

\begin{table}[h]
\centering
\small
\setlength{\tabcolsep}{4pt}
\begin{tabular}{lcccc}
\toprule
\textbf{Category} & \textbf{OPD} & \textbf{OPSD} & \textbf{GRPO} & \textbf{PPO} \\
\midrule
Rollout batch size & $64$ & $64$ & $32$ & $64$ \\
Prompts per batch & $64$ & $64$ & $32$ & $64$ \\
Samples per prompt & $1$ & $1$ & $8$ & $1$ \\
Maximum rollout response length & $4096$ & $4096$ & $8192$ & $8192$ \\
Rollout temperature & $1.0$ & $1.0$ & $1.0$ & $1.0$ \\
Rollout top-$p$ & $0.95$ & $0.95$ & $0.95$ & $0.95$ \\
\midrule
Optimizer & Adam & Adam & Adam & Adam \\
Learning rate & $2\times 10^{-6}$ & $2\times 10^{-6}$ & $2\times 10^{-6}$ & $2\times 10^{-6}$ \\
Learning-rate schedule & cosine decay & cosine decay & cosine decay & cosine decay \\
Warmup fraction & $0.1$ & $0.1$ & $0.1$ & $0.1$ \\
Weight decay & $0.1$ & $0.1$ & $0.1$ & $0.1$ \\
Adam $\beta_1$ / $\beta_2$ & $0.9$ / $0.98$ & $0.9$ / $0.98$ & $0.9$ / $0.98$ & $0.9$ / $0.98$ \\
\midrule
Training objective & Top-$K$ reverse KL & Top-$K$ reverse KL & policy gradient & PPO \\
KL / regularization coefficient & -- & -- & $0.0$ & $0.0$ \\
Top-$K$ & $20$ & $20$ & -- & -- \\
Advantage estimator & -- & -- & GRPO & GAE \\
GAE $\gamma$ / $\lambda$ & -- & -- & -- & $1.0$ / $0.95$ \\
Teacher privileged information & none & final answer / full response & -- & -- \\
Teacher model & external teacher & step-0 student & -- & -- \\
\bottomrule
\end{tabular}
\vspace{2mm}
\caption{Default training hyperparameters for OPD, OPSD, GRPO, and PPO.}
\label{tab:train_hparams}
\end{table}

\subsection{Evaluation Metrics}
Here we provide a detailed definition of several metrics appeared in the figures. Let \(y=(y_1,\dots,y_T)\) be a generated response. For token position \(t\), let \(x,y_{<t}\) denote the prompt together with the previously generated prefix up to position \(t-1\). We define the teacher and student log-probabilities of the sampled token \(y_t\) as
\[
l_T(t)=\log \pi_T(y_t\mid x,y_{<t},I),
\qquad
l_S(t)=\log \pi_S(y_t\mid x,y_{<t}),
\]
and their log-probability gap as
\[
\Delta \text{logprob}_t=l_T(t)-l_S(t).
\]

\paragraph{Repetition.}
We identify repetition using an \(n\)-gram criterion with default \(n=3\). A token position \(t\) is marked as repetitive if the \(n\)-gram ending at \(t\) has appeared previously in the same response. Formally, let \(r_t=1\) if \((y_{t-n+1},\dots,y_t)\) has occurred earlier in the sequence, and \(r_t=0\) otherwise. The repetition ratio is
\[
\mathrm{RepRatio}=\frac{1}{N}\sum_t r_t,
\]
where \(N\) is the number of valid response tokens. We further partition token positions into repetitive and non-repetitive sets,
\[
\mathcal{R}=\{t:r_t=1\},\qquad \mathcal{O}=\{t:r_t=0\},
\]
and for any token-level quantity \(a_t\), define the conditional averages
\[
\bar a_{\mathrm{rep}}=\frac{1}{|\mathcal R|}\sum_{t\in\mathcal R} a_t,
\qquad
\bar a_{\mathrm{other}}=\frac{1}{|\mathcal O|}\sum_{t\in\mathcal O} a_t.
\]

\paragraph{Teacher-student agreement.}
To measure local agreement between teacher and student token distributions, we compare their TopK candidate sets at each position, with default \(K=50\). Let \(\mathrm{TopK}_T(t)\) and \(\mathrm{TopK}_S(t)\) denote the teacher and student TopK sets at position \(t\). We define the token-level overlap as
\[
\mathrm{Overlap}_t=\frac{|\mathrm{TopK}_T(t)\cap \mathrm{TopK}_S(t)|}{K}.
\]
We also record the teacher-side rank of the sampled token \(y_t\):
\[
\mathrm{Rank@K}_t=
\begin{cases}
\mathrm{rank}_{\pi_T}(y_t\mid x,y_{<t},I), & y_t\in \mathrm{TopK}_T(t),\\
K+1, & \text{otherwise}.
\end{cases}
\]
Finally, we report repetition-conditional averages of \(\mathrm{Overlap}_t\) and \(\mathrm{Rank@K}_t\) over \(\mathcal R\) and \(\mathcal O\).

\subsection{Design Space of OP(S)D}

An OP(S)D algorithm is characterized by three largely orthogonal design axes: teacher construction, privileged information design, and distillation loss. In practice, the first two axes, teacher construction and privileged information design, are primarily specific to OPSD, since in standard OPD the teacher is typically a fixed model and no privileged information is introduced. By contrast, the distillation loss is a central design choice for both OPSD and OPD.

\subsubsection{Teacher Construction}

A first design choice is how the teacher policy $\pi_T$ is constructed.

\paragraph{Self-Teacher.} uses exactly the same model as the student. Let $c$ denote the privileged information. The teacher is
\[
\pi_T(\cdot \mid x, y_{<t}, c)
=
\pi_\theta(\cdot \mid x, y_{<t}, c).
\]
Here, the teacher and student share the same parameters $\theta$, so the teacher is updated jointly with the student throughout training. This setup has appeared in several papers such as \cite{zhao2026selfdistilledreasoneronpolicyselfdistillation,ye2026onpolicycontextdistillationlanguage}.

\paragraph{Frozen teacher.}
A second option is to fix the teacher parameters throughout training. One simple instantiation is to initialize the teacher from the student at the beginning of training and then keep it frozen:
\[
\pi_T(\cdot \mid x, y_{<t}, c)
=
\pi_{\bar{\theta}}(\cdot \mid x, y_{<t}, c),
\qquad
\bar{\theta} = \theta^{(0)},
\]
where $\theta^{(0)}$ denotes the student parameters at initialization. More generally, $\pi_{\bar{\theta}}$ could also be a stronger pretrained model or a separately trained expert policy. This design often improves stability, since the teacher target does not drift during optimization, but it may become stale as the student improves, and can also induce mismatch between the teacher's behavior and the student-induced state distribution.

\paragraph{exponential moving average (EMA) teacher.}

An intermediate design is to maintain a moving teacher whose parameters track the student through EMA:
\[
\bar{\theta}_{k}
=
\alpha \bar{\theta}_{k-1} + (1-\alpha)\theta_k,
\qquad \alpha \in [0,1),
\]
and define
\[
\pi_T(\cdot \mid x, y_{<t}, c)
=
\pi_{\bar{\theta}_k}(\cdot \mid x, y_{<t}, c).
\]
This construction preserves the self-distillation flavor while providing a more stable target than the fully coupled teacher $\pi_\theta$. Intuitively, the EMA teacher smooths high-variance parameter updates, tracks the student's learning progress, and avoids the instability that can arise when the current student is used directly as its own target. \cite{shenfeld2026selfdistillationenablescontinuallearning,hubotter2026reinforcementlearningselfdistillation}

\noindent\textcolor{purple}{\textbf{Takeaway.}} A frozen teacher is stable, but its capability is inherently bounded. In contrast, an EMA teacher or self teacher is often assumed to better exploit privileged information, yet this advantage is not guaranteed in practice. Unless the teacher itself is directly optimized for the task objective, for example via RL, there is little reason to expect plain OP(S)D alone to produce a teacher that is stronger than the original base teacher in leveraging privileged information to solve the task.

\subsubsection{Privileged Information Design}
A third design choice concerns how privileged information is constructed for the teacher. Depending on the task, privileged information can take several forms. For math reasoning, it may consist of ground-truth reasoning together with the final answer, or a lighter variant that provides only the answer \cite{zhao2026selfdistilledreasoneronpolicyselfdistillation}. For agentic tasks, it may come from environment feedback, such as tool execution results, verifier signals, or environment observations \cite{hubotter2026reinforcementlearningselfdistillation}. It can also be derived from past experience, for example when an external model summarizes interaction history into high-level guidance that the teacher conditions on \cite{wang2026openclawrltrainagentsimply,ye2026onpolicycontextdistillationlanguage}. This last form is especially useful when raw feedback is long or difficult to use directly.

\subsubsection{Distillation Objectives}

The remaining choice is how the student matches the teacher. This design is closely related to the KL regularization commonly used in PPO-style training \cite{schulman2017proximalpolicyoptimizationalgorithms}, where the policy is regularized using a reference model. In our setting, the teacher plays a similar role, and the distillation signal can be implemented either by a full-vocabulary KL or by sampled-token KL estimators.

\paragraph{Full-vocabulary objectives.}
\label{subsubsection:full_vocab_loss}

A standard choice is to match the student and teacher token distributions at
each position using a distribution-level divergence. Let
\(\pi_\theta(\cdot \mid x, y_{<i})\) denote the student distribution at step
\(i\), and let \(\pi_T(\cdot \mid x, c, y_{<i})\) denote the teacher
distribution. We optimize
\[
\mathcal{L}_{\mathrm{full}}(\theta)
=
\mathbb{E}_{x\sim\mathcal D,\; y\sim\pi_\theta(\cdot\mid x)}
\left[
\frac{1}{T}\sum_{i=1}^T \ell_i^{\mathrm{full}}
\right].
\]

Common choices for \(\ell_i^{\mathrm{full}}\) include reverse-KL,
\[
D_{\mathrm{KL}}\!\left(
\pi_\theta(\cdot\mid x,y_{<i})
\;\middle\|\;
\pi_T(\cdot\mid x,c,y_{<i})
\right),
\]
forward-KL,
\[
D_{\mathrm{KL}}\!\left(
\pi_T(\cdot\mid x,c,y_{<i})
\;\middle\|\;
\pi_\theta(\cdot\mid x,y_{<i})
\right),
\]
and Jensen-Shannon divergence,
\[
\beta\,
D_{\mathrm{KL}}\!\left(
\pi_\theta(\cdot\mid x,y_{<i})
\;\middle\|\;
m_i
\right)
+
(1-\beta)\,
D_{\mathrm{KL}}\!\left(
\pi_T(\cdot\mid x,c,y_{<i})
\;\middle\|\;
m_i
\right),
\]
where
\[
m_i
=
\beta\,\pi_\theta(\cdot\mid x,y_{<i})
+
(1-\beta)\,\pi_T(\cdot\mid x,c,y_{<i}).
\]

These objectives require full-vocabulary probabilities at every position and
can be memory-intensive. A common approximation is to restrict computation to a
TopK set \(S_i \subseteq \mathcal V\), selected either from the teacher \cite{fu2026revisitingonpolicydistillationempirical}
or from the student \cite{zhao2026selfdistilledreasoneronpolicyselfdistillation, hubotter2026reinforcementlearningselfdistillation}:
\[
S_i^{T}
=
\operatorname{TopK}\!\left(
\pi_T(\cdot\mid x,c,y_{<i}),\,k
\right),
\qquad
S_i^{S}
=
\operatorname{TopK}\!\left(
\pi_\theta(\cdot\mid x,y_{<i}),\,k
\right).
\]

Given a chosen set \(S_i\), the truncated reverse-KL and forward-KL are
\[
D_{\mathrm{KL}}^{S_i}\!\left(
\pi_\theta
\;\middle\|\;
\pi_T
\right)
=
\sum_{v\in S_i}
\pi_\theta(v\mid x,y_{<i})
\log
\frac{
\pi_\theta(v\mid x,y_{<i})
}{
\pi_T(v\mid x,c,y_{<i})
},
\]
and
\[
D_{\mathrm{KL}}^{S_i}\!\left(
\pi_T
\;\middle\|\;
\pi_\theta
\right)
=
\sum_{v\in S_i}
\pi_T(v\mid x,c,y_{<i})
\log
\frac{
\pi_T(v\mid x,c,y_{<i})
}{
\pi_\theta(v\mid x,y_{<i})
}.
\]
Similarly, the truncated Jensen-Shannon divergence is
\[
\mathrm{JSD}^{S_i}\!\left(
\pi_\theta,\pi_T
\right)
=
\beta\,
D_{\mathrm{KL}}^{S_i}\!\left(
\pi_\theta
\;\middle\|\;
m_i
\right)
+
(1-\beta)\,
D_{\mathrm{KL}}^{S_i}\!\left(
\pi_T
\;\middle\|\;
m_i
\right).
\]

A further refinement is to aggregate the probability mass outside \(S_i\) into
a single tail distribution. For example, in SDPO~\cite{hubotter2026reinforcementlearningselfdistillation},
\(S_i\) is chosen as the student TopK, they define
\[
p_i^{\mathrm{tail}}
=
1-\sum_{v\in S_i}\pi_\theta(v\mid x,y_{<i}),
\qquad
q_i^{\mathrm{tail}}
=
1-\sum_{v\in S_i}\pi_T(v\mid x,c,y_{<i}),
\]
and use the augmented reverse-KL
\[
\widetilde D_{\mathrm{KL}}^{S_i}\!\left(
\pi_\theta
\;\middle\|\;
\pi_T
\right)
=
\sum_{v\in S_i}
\pi_\theta(v\mid x,y_{<i})
\log
\frac{
\pi_\theta(v\mid x,y_{<i})
}{
\pi_T(v\mid x,c,y_{<i})
}
+
p_i^{\mathrm{tail}}
\log
\frac{
p_i^{\mathrm{tail}}
}{
q_i^{\mathrm{tail}}
}.
\]
Analogous tail-augmented forms can be defined for forward-KL and JSD. This
preserves the remaining probability mass and is often a better approximation
than simply truncating the support.

\paragraph{Sampled-token objectives.}
\label{subsubsection:sampled_token_loss}

Instead of matching the full teacher distribution, one may construct a sampled-token signal using only the teacher and student log-probabilities on tokens sampled from the student policy. Let
\[
A_t
=
\log \pi_T(y_t\mid x,c,y_{<t})
-
\log \pi_\theta(y_t\mid x,y_{<t}),
\qquad
y_t \sim \pi_\theta(\cdot \mid x,y_{<t}).
\]
Then
\[
-A_t
=
\log \pi_\theta(y_t\mid x,y_{<t})
-
\log \pi_T(y_t\mid x,c,y_{<t}),
\]
whose expectation under \(y_t \sim \pi_\theta(\cdot\mid x,y_{<t})\) recovers the token-level reverse KL. Following PPO-style KL regularization, one may also consider alternative sampled surrogates such as
\[
k_{1,t} = -A_t,
\qquad
k_{2,t} = \frac{1}{2}A_t^2,
\qquad
k_{3,t} = e^{A_t} - 1 - A_t.
\]
These estimators differ in bias, variance, and non-negativity properties. \(k_3\) is commonly used as a low-variance, non-negative KL surrogate.

In practice, sampled-token OP(S)D is usually optimized in a policy-gradient form:
\[
\mathcal{L}_{\mathrm{sample}}(\theta)
=
-
\mathbb{E}_{x\sim\mathcal D,\; y\sim\pi_\theta(\cdot\mid x)}
\left[
\frac{1}{T}\sum_{t=1}^T
A_t\,
\log \pi_\theta(y_t\mid x,y_{<t})
\right],
\]

Compared with full-vocabulary objectives, this sampled-token form only changes
the probability of tokens actually sampled by the student. It behaves
more like reinforcement learning with a dense token-level reward, whereas
full-vocabulary KL directly reshapes the entire next-token distribution.




\paragraph{Combination with policy gradient.}
The OP(S)D learning signal can be combined with reinforcement learning in two ways. One option is to optimize it as a regularization term:
\[
\mathcal{L}_{\mathrm{total}}(\theta)
=
\mathcal{L}_{\mathrm{RL}}(\theta)
+
\lambda \mathcal{L}_{\mathrm{OPD}}(\theta),
\]
where \(\mathcal{L}_{\mathrm{RL}}(\theta)\) denotes the standard RL loss, and \(\mathcal{L}_{\mathrm{OPD}}(\theta)\) is an OP(S)D objective. Alternatively, OP(S)D can be treated as a token-level advantage. We define the total advantage as
\[
A_t^{\mathrm{total}}
=
A_t^{\mathrm{RL}}
+
\lambda A_t^{\mathrm{OPD}},
\]
and optimize the objective
\[
\mathcal{L}(\theta)
=
-
\mathbb{E}_{x,\;y\sim\pi_\theta}
\left[
\frac{1}{T}\sum_{t=1}^T
A_t^{\mathrm{total}}
\log \pi_\theta(y_t\mid x,y_{<t})
\right].
\]

\noindent\textcolor{purple}{\textbf{Takeaway.}}
When incorporated into the advantage, OP(S)D is typically designed in a sampled-token form so that it aligns naturally with token-level policy gradient updates. When OP(S)D is used as an auxiliary regularization loss, it is more commonly implemented as a vocabulary-level objective that matches the full teacher distribution.

\subsection{General OP(S)D Gradient Decomposition.} 
Consider the objective
\[
\mathcal{L}_{\mathrm{OPD}}(\theta)
=
\mathbb{E}_{x\sim \mathcal{D},\; y\sim \pi_\theta(\cdot\mid x)}
\bigl[\ell(\theta; x,y)\bigr].
\]
For a fixed input \(x\), define
\[
\mathcal{L}_x(\theta)
=
\mathbb{E}_{y\sim \pi_\theta(\cdot\mid x)}
\bigl[\ell(\theta; x,y)\bigr]
=
\sum_y \pi_\theta(y\mid x)\,\ell(\theta; x,y).
\]
Differentiating with respect to \(\theta\), we obtain
\[
\nabla_\theta \mathcal{L}_x(\theta)
=
\nabla_\theta \sum_y \pi_\theta(y\mid x)\,\ell(\theta; x,y)
\]
\[
=
\sum_y \nabla_\theta\!\Bigl(\pi_\theta(y\mid x)\,\ell(\theta; x,y)\Bigr)
\]
\[
=
\sum_y
\Bigl[
(\nabla_\theta \pi_\theta(y\mid x))\,\ell(\theta; x,y)
+
\pi_\theta(y\mid x)\,\nabla_\theta \ell(\theta; x,y)
\Bigr]
\]
\[
=
\sum_y \pi_\theta(y\mid x)\,\nabla_\theta \ell(\theta; x,y)
+
\sum_y \ell(\theta; x,y)\,\nabla_\theta \pi_\theta(y\mid x).
\]
Using the identity
\[
\nabla_\theta \pi_\theta(y\mid x)
=
\pi_\theta(y\mid x)\,\nabla_\theta \log \pi_\theta(y\mid x),
\]
we get
\[
\nabla_\theta \mathcal{L}_x(\theta)
=
\sum_y \pi_\theta(y\mid x)\,\nabla_\theta \ell(\theta; x,y)
+
\sum_y \ell(\theta; x,y)\,\pi_\theta(y\mid x)\,\nabla_\theta \log \pi_\theta(y\mid x)
\]
\[
=
\mathbb{E}_{y\sim \pi_\theta(\cdot\mid x)}
\bigl[\nabla_\theta \ell(\theta; x,y)\bigr]
+
\mathbb{E}_{y\sim \pi_\theta(\cdot\mid x)}
\bigl[\ell(\theta; x,y)\,\nabla_\theta \log \pi_\theta(y\mid x)\bigr].
\]
Therefore,
\[
\nabla_\theta \mathcal{L}_{\mathrm{OPD}}(\theta)
=
\mathbb{E}_{x\sim\mathcal D,\; y\sim \pi_\theta(\cdot\mid x)}
\Bigl[
\nabla_\theta \ell(\theta; x,y)
+
\ell(\theta; x,y)\,\nabla_\theta \log \pi_\theta(y\mid x)
\Bigr].
\]

The first term is the direct gradient term, which backpropagates
through the token-level loss on sampled prefixes. The second term is the
score-function term:
\[
\mathbb{E}_{x,y\sim\pi_\theta}
\bigl[
\ell(\theta; x,y)\,\nabla_\theta \log \pi_\theta(y\mid x)
\bigr],
\]
which is the sequence-level policy-gradient term induced by the dependence of
the rollout distribution on \(\theta\).

In many practical on-policy distillation methods, one ignores the score-function
term and treats sampled trajectories as fixed. This gives the approximate
gradient
\[
g_{\mathrm{approx}}
=
\mathbb{E}_{x\sim\mathcal D,\; y\sim \pi_\theta(\cdot\mid x)}
\bigl[
\nabla_\theta \ell(\theta; x,y)
\bigr].
\]
This approximation introduces bias, since it omits
\[
\mathbb{E}_{x,y\sim\pi_\theta}
\bigl[
\ell(\theta; x,y)\,\nabla_\theta \log \pi_\theta(y\mid x)
\bigr],
\]
but it often substantially reduces variance and improves optimization stability.
By contrast, sequence-level RL-style estimators retain the score-function term,
which yields an unbiased estimator of the full gradient but typically with much
higher variance.

\subsection{Detailed Gradient computation.}

\paragraph{Full-vocabulary reverse-KL.}
Consider
\[
\mathcal{L}_{\mathrm{vocab}}(\theta)
=
\mathbb{E}_{x\sim\mathcal D,\; y\sim\pi_\theta(\cdot\mid x)}
\left[
\frac{1}{T}\sum_{t=1}^T
D_{\mathrm{KL}}\!\left(
\pi_\theta(\cdot\mid x,y_{<t})
\;\middle\|\;
\pi_T(\cdot\mid x,c,y_{<t})
\right)
\right].
\]
Let
\[
\ell_t(\theta,y_{<t})
=
D_{\mathrm{KL}}\!\left(
\pi_\theta(\cdot\mid x,y_{<t})
\;\middle\|\;
\pi_T(\cdot\mid x,c,y_{<t})
\right),
\qquad
L(\theta,y)=\frac{1}{T}\sum_{t=1}^T \ell_t(\theta,y_{<t}).
\]
Since \(\theta\) appears both in the rollout distribution \(y\sim\pi_\theta(\cdot\mid x)\) and explicitly in each \(\ell_t\), the full gradient is
\[
\nabla_\theta \mathcal{L}_{\mathrm{vocab}}(\theta)
=
\mathbb{E}_{x,y}
\left[
L(\theta,y)\nabla_\theta \log \pi_\theta(y\mid x)
+
\frac{1}{T}\sum_{t=1}^T \nabla_\theta \ell_t(\theta,y_{<t})
\right].
\]
Using \(\log \pi_\theta(y\mid x)=\sum_{t=1}^T \log \pi_\theta(y_t\mid x,y_{<t})\), the rollout term becomes
\[
\mathbb{E}_{x,y}
\left[
L(\theta,y)\sum_{t=1}^T \nabla_\theta \log \pi_\theta(y_t\mid x,y_{<t})
\right].
\]

For the explicit gradient term, define
\[
p_t(\cdot)=\pi_\theta(\cdot\mid x,y_{<t}),
\qquad
q_t(\cdot)=\pi_T(\cdot\mid x,c,y_{<t}).
\]
Then
\[
\ell_t=\sum_{v\in\mathcal V} p_t(v)\log\frac{p_t(v)}{q_t(v)},
\]
and, treating the teacher as stop-gradient,
\[
\nabla_\theta \ell_t
=
\mathbb{E}_{v\sim p_t}
\left[
\bigl(\log p_t(v)-\log q_t(v)\bigr)\nabla_\theta \log p_t(v)
\right].
\]
Equivalently, with
\[
A_t(v)=\log q_t(v)-\log p_t(v),
\]
we have
\[
\nabla_\theta \ell_t
=
-
\mathbb{E}_{v\sim\pi_\theta(\cdot\mid x,y_{<t})}
\left[
A_t(v)\nabla_\theta \log \pi_\theta(v\mid x,y_{<t})
\right].
\]

Therefore,
\[
\nabla_\theta \mathcal{L}_{\mathrm{vocab}}(\theta)
=
\mathbb{E}_{x,y}
\left[
L(\theta,y)\sum_{t=1}^T \nabla_\theta \log \pi_\theta(y_t\mid x,y_{<t})
-
\frac{1}{T}\sum_{t=1}^T
\mathbb{E}_{v\sim\pi_\theta(\cdot\mid x,y_{<t})}
\left[
A_t(v)\nabla_\theta \log \pi_\theta(v\mid x,y_{<t})
\right]
\right].
\]

In practice, when using full-vocab KL, we ignore the first term:
\[
\nabla_\theta \mathcal{L}_{\mathrm{vocab}}(\theta)
=
-
\mathbb{E}_{x,y}
\left[
\frac{1}{T}\sum_{t=1}^T
\mathbb{E}_{v\sim\pi_\theta(\cdot\mid x,y_{<t})}
\left[
A_t(v)\nabla_\theta \log \pi_\theta(v\mid x,y_{<t})
\right]
\right].
\]

This approximation is used in SDPO \cite{hubotter2026reinforcementlearningselfdistillation} and SDFT \cite{shenfeld2026selfdistillationenablescontinuallearning}.

\paragraph{Sampled-token reverse-KL.}
For the sampled-token objective,
\[
\mathcal{L}_{\mathrm{sample}}(\theta)
=
-
\mathbb{E}_{x,\; y\sim \pi_\theta(\cdot\mid x)}
\left[
R(y)
\right],
\qquad
R(y)=\frac{1}{T}\sum_{t=1}^T r_t,
\]
where
\[
r_t
=
\log \pi_T(y_t\mid x,c,y_{<t})
-
\log \pi_\theta(y_t\mid x,y_{<t}).
\]
Using the identity
$
\nabla_\theta \pi_\theta(y\mid x)
=
\pi_\theta(y\mid x)\,\nabla_\theta \log \pi_\theta(y\mid x),
$ we get 
\[
\nabla_\theta \mathcal{L}_{\mathrm{sample}}(\theta)
=
-
\mathbb{E}_{x,\; y}
\left[
R(y)\,\nabla_\theta \log \pi_\theta(y\mid x)
+
\nabla_\theta R(y)
\right].
\]
Since
\[
\log \pi_\theta(y\mid x)
=
\sum_{s=1}^T \log \pi_\theta(y_s\mid x,y_{<s}),
\]
and stopping gradients through the teacher gives
\[
\nabla_\theta R(y)
=
-\frac{1}{T}\sum_{t=1}^T
\nabla_\theta \log \pi_\theta(y_t\mid x,y_{<t}),
\]
we obtain
\[
\nabla_\theta \mathcal{L}_{\mathrm{sample}}(\theta)
=
-
\mathbb{E}_{x,\; y}
\left[
\frac{1}{T}\sum_{t=1}^T
\left(
r_t-1
\right)
\nabla_\theta \log \pi_\theta(y_t\mid x,y_{<t})
\right].
\]

While in practice \cite{zhao2026selfdistilledreasoneronpolicyselfdistillation}, people usually ignore the \texttt{-1} term and use

\[
\nabla_\theta \mathcal{L}_{\mathrm{sample}}(\theta)
=
-
\mathbb{E}_{x,\; y}
\left[
\frac{1}{T}\sum_{t=1}^T
r_t
\nabla_\theta \log \pi_\theta(y_t\mid x,y_{<t})
\right].
\]

\subsection{TopK Engineering Challenge}
\label{topkchallenge}
A practical engineering issue arises in on-policy distillation when we use the student's per-position TopK tokens to query the teacher's corresponding vocabulary probabilities. The root cause is a mismatch between the data structure produced by the distillation pipeline and the interface expected by SGLang.

In our implementation, \texttt{reward\_func} sets \texttt{payload["token\_ids\_logprob"]} to \texttt{requested\_topk\_token\_ids}, which is organized as a per-position list of token sets, \emph{i.e.}, \texttt{[[K IDs at position 0], [K IDs at position 1], \dots]}. However, SGLang's \texttt{token\_ids\_logprob} API does not support position-dependent token lists. Instead, it expects a single flat list of token IDs that is applied uniformly across all positions.

This incompatibility is exposed by the \texttt{compute\_logprobs\_only} path in \texttt{tp\_worker.py}. In that path, the code calls \texttt{get\_token\_ids\_logprobs\_batch\_optimized(logprobs\_shape=[1, vocab], token\_ids\_logprobs=[[per\_pos\_list]])}. As a result, the constructed column index tensor becomes \texttt{torch.tensor([list\_K, list\_K, \dots])}, which has shape \texttt{[L, K]} instead of the expected one-dimensional shape \texttt{[L]}. Consequently, the downstream \texttt{gather} operation fails due to the shape mismatch.

To resolve this issue, we flatten the per-position TopK token IDs into a single sorted set of unique token IDs, and send this flat union as \texttt{token\_ids\_logprob}. Under this representation, the prefill path correctly computes \texttt{logprobs[positions, flat\_union]}, producing a tensor of shape \texttt{[seq\_len, union\_size]}. This format is already compatible with the existing extraction logic in \texttt{\_extract\_response\_aux\_id\_logprob\_maps}, which can recover the relevant teacher log-probabilities for each response position from the shared union vocabulary. Therefore, the fix preserves the intended semantics of querying teacher probabilities on the student's TopK candidates, while making the implementation consistent with SGLang's API contract.

\subsection{Additional OPD Experiment using Top20 reverse KL}
\label{app:top20reverseKLexp}

\begin{figure}[t]
    \centering
    \begin{subfigure}[t]{0.24\textwidth}
        \centering
        \includegraphics[width=\textwidth]{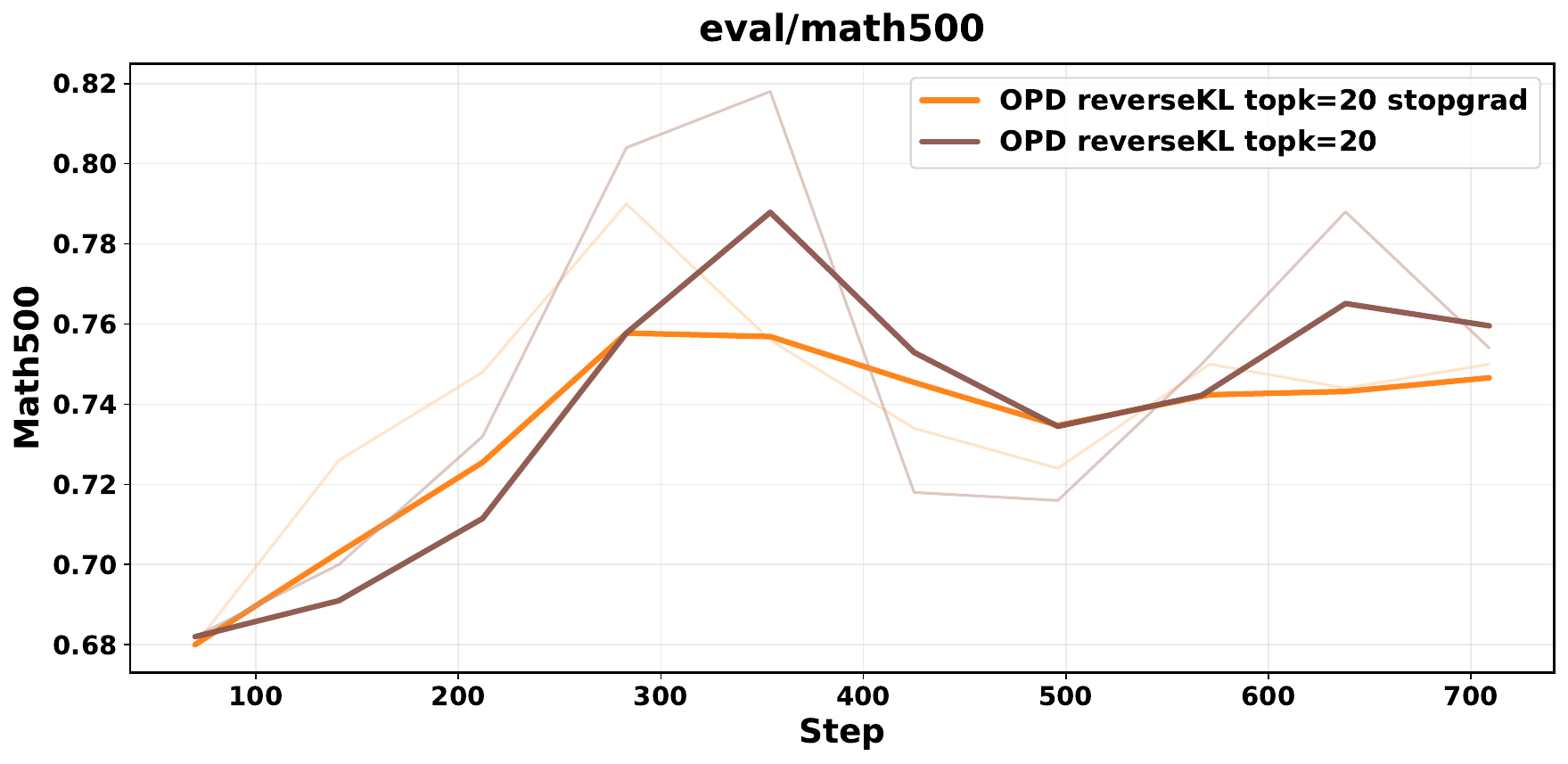}
        \label{fig:4b_char}
    \end{subfigure}
    \hfill
    \begin{subfigure}[t]{0.24\textwidth}
        \centering
        \includegraphics[width=\textwidth]{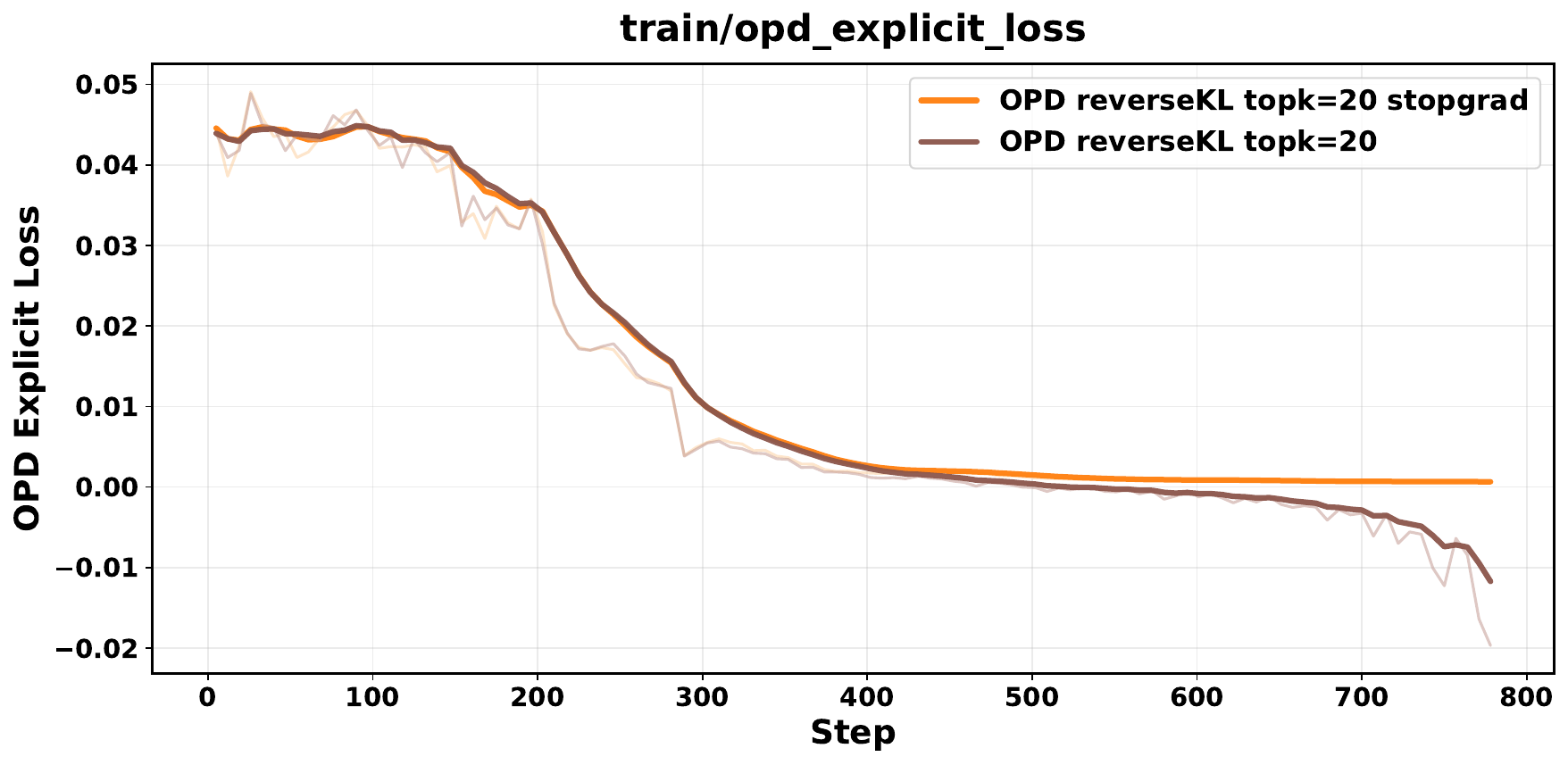}
        \label{fig:4b_emo}
    \end{subfigure}
    \vspace{0.5em}
    \begin{subfigure}[t]{0.24\textwidth}
        \centering
        \includegraphics[width=\textwidth]{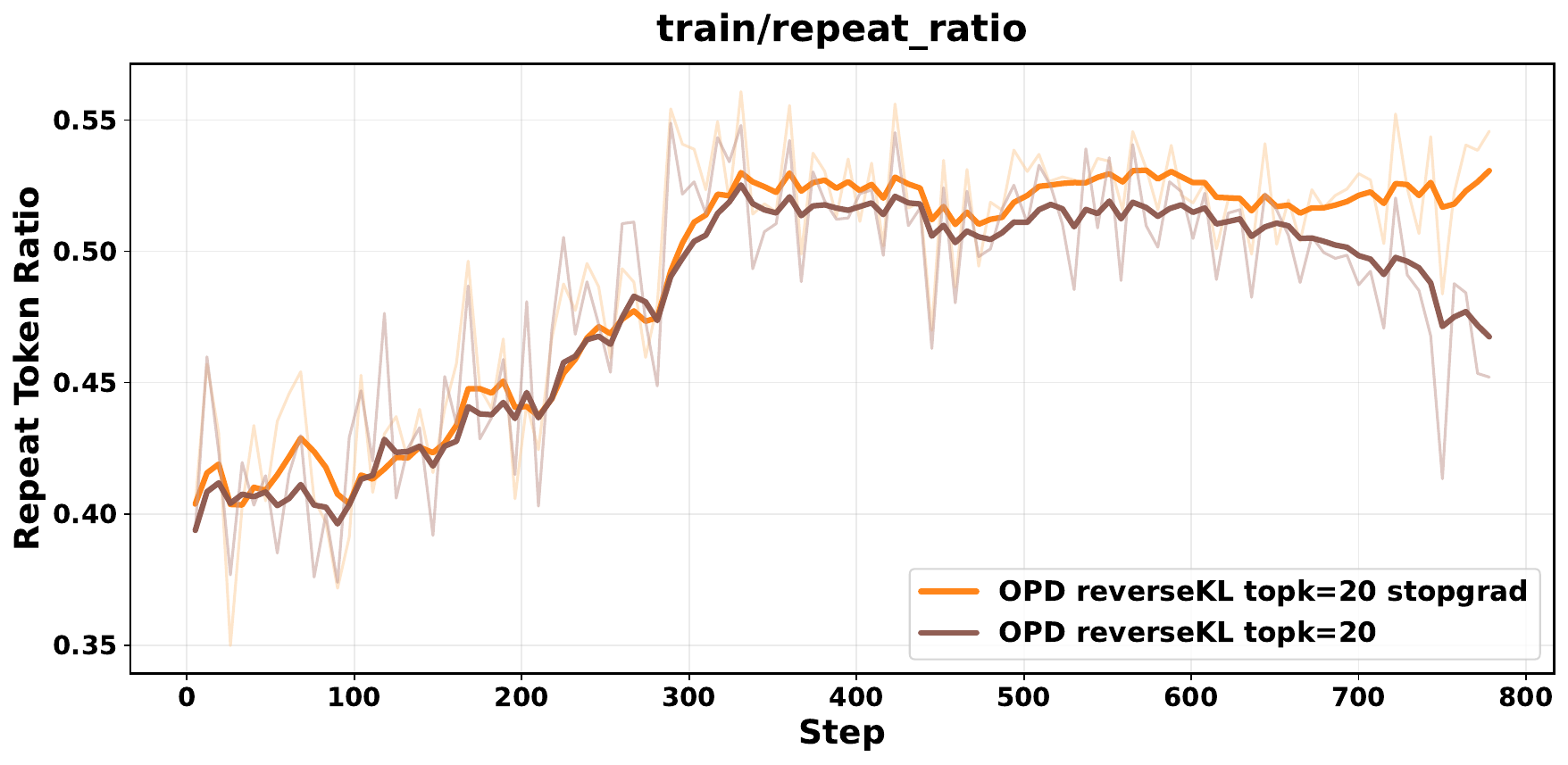}
        \label{fig:8b_char}
    \end{subfigure}
    \hfill
    \begin{subfigure}[t]{0.24\textwidth}
        \centering
        \includegraphics[width=\textwidth]{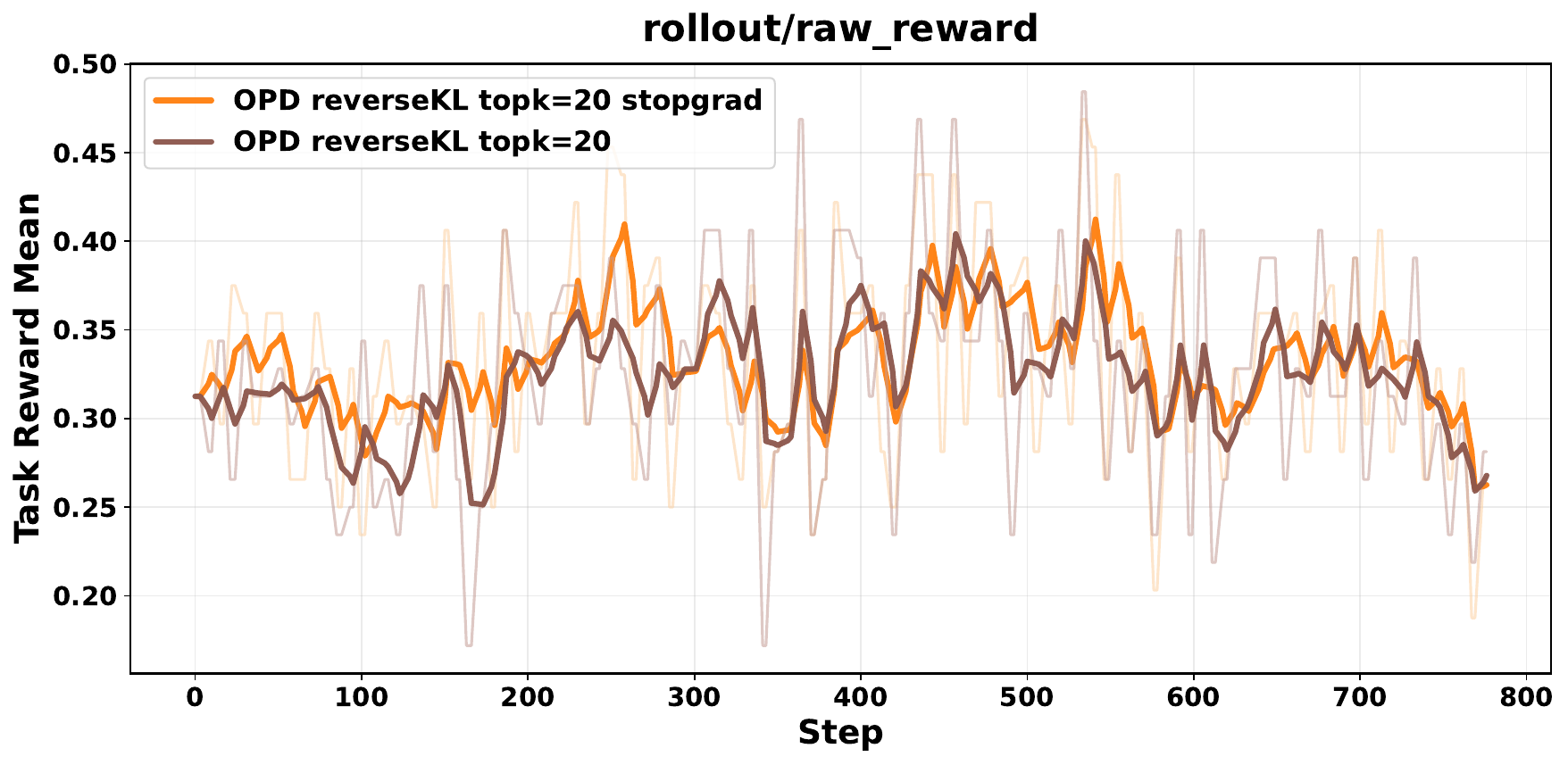}
        \label{fig:8b_emo}
    \end{subfigure}
    
    \begin{subfigure}[t]{0.24\textwidth}
        \centering
        \includegraphics[width=\textwidth]{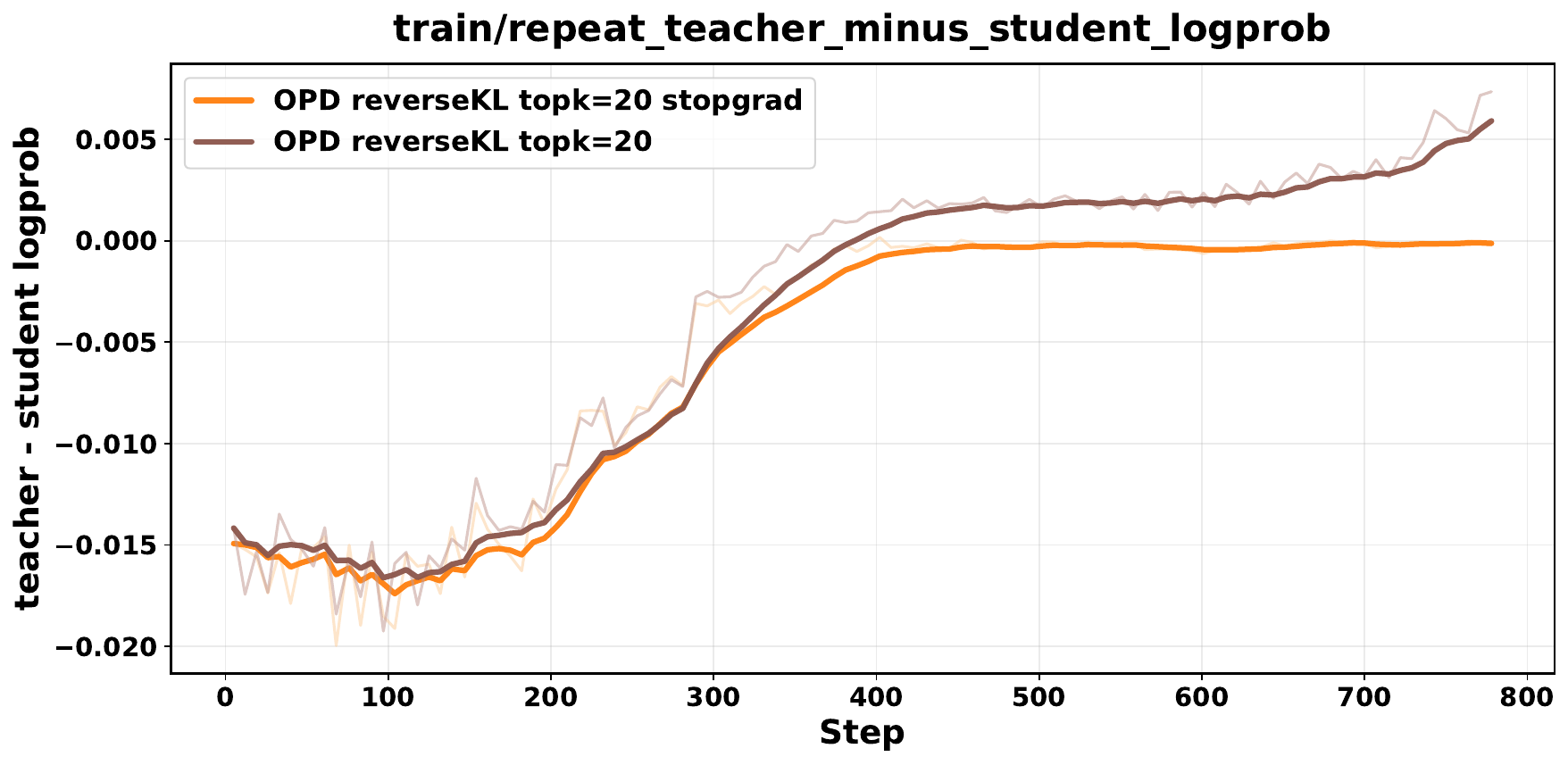}
        \label{fig:4b_char}
    \end{subfigure}
    \hfill
    \begin{subfigure}[t]{0.24\textwidth}
        \centering
        \includegraphics[width=\textwidth]{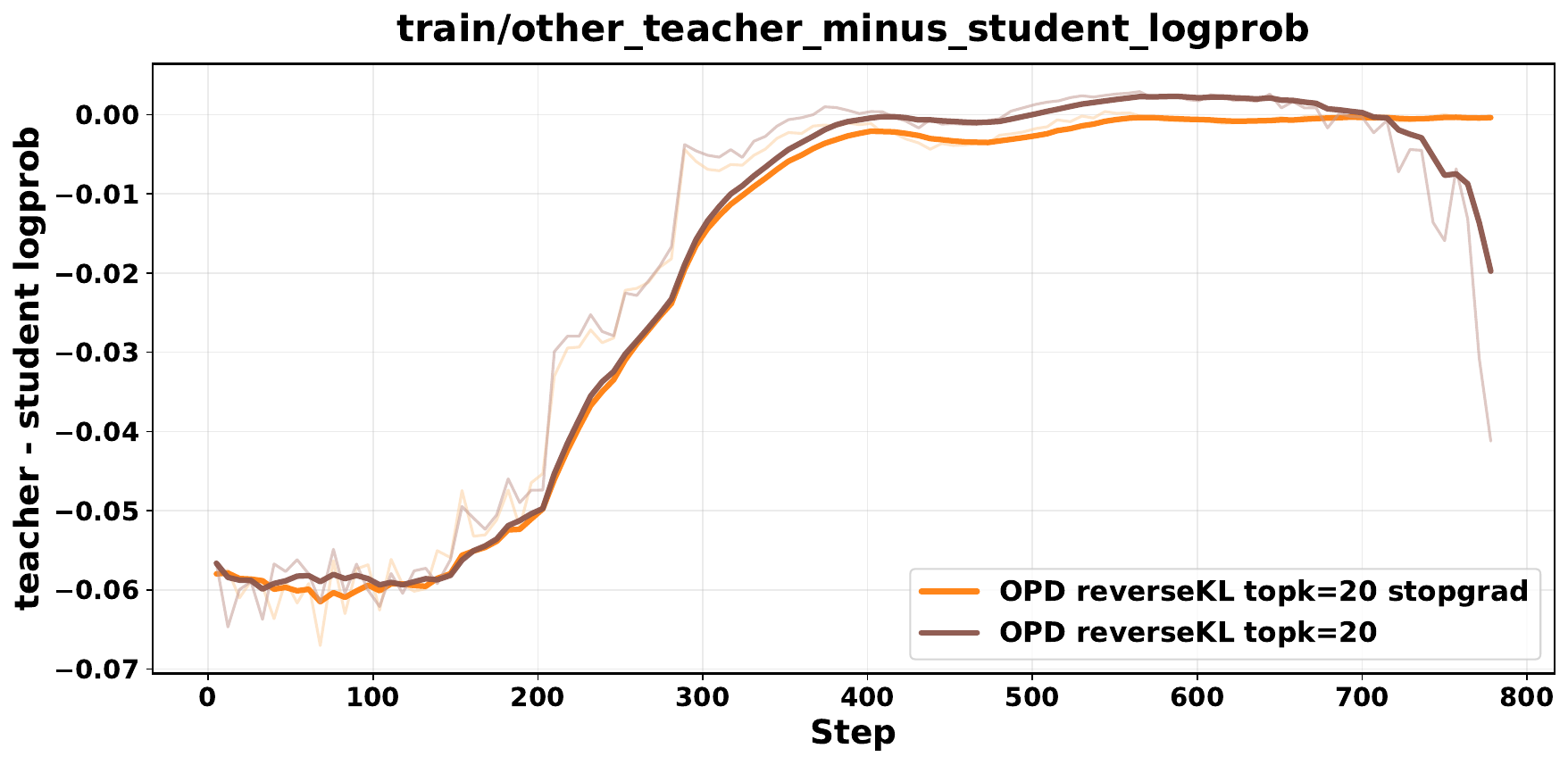}
        \label{fig:4b_emo}
    \end{subfigure}
    \vspace{0.5em}
    \begin{subfigure}[t]{0.24\textwidth}
        \centering
        \includegraphics[width=\textwidth]{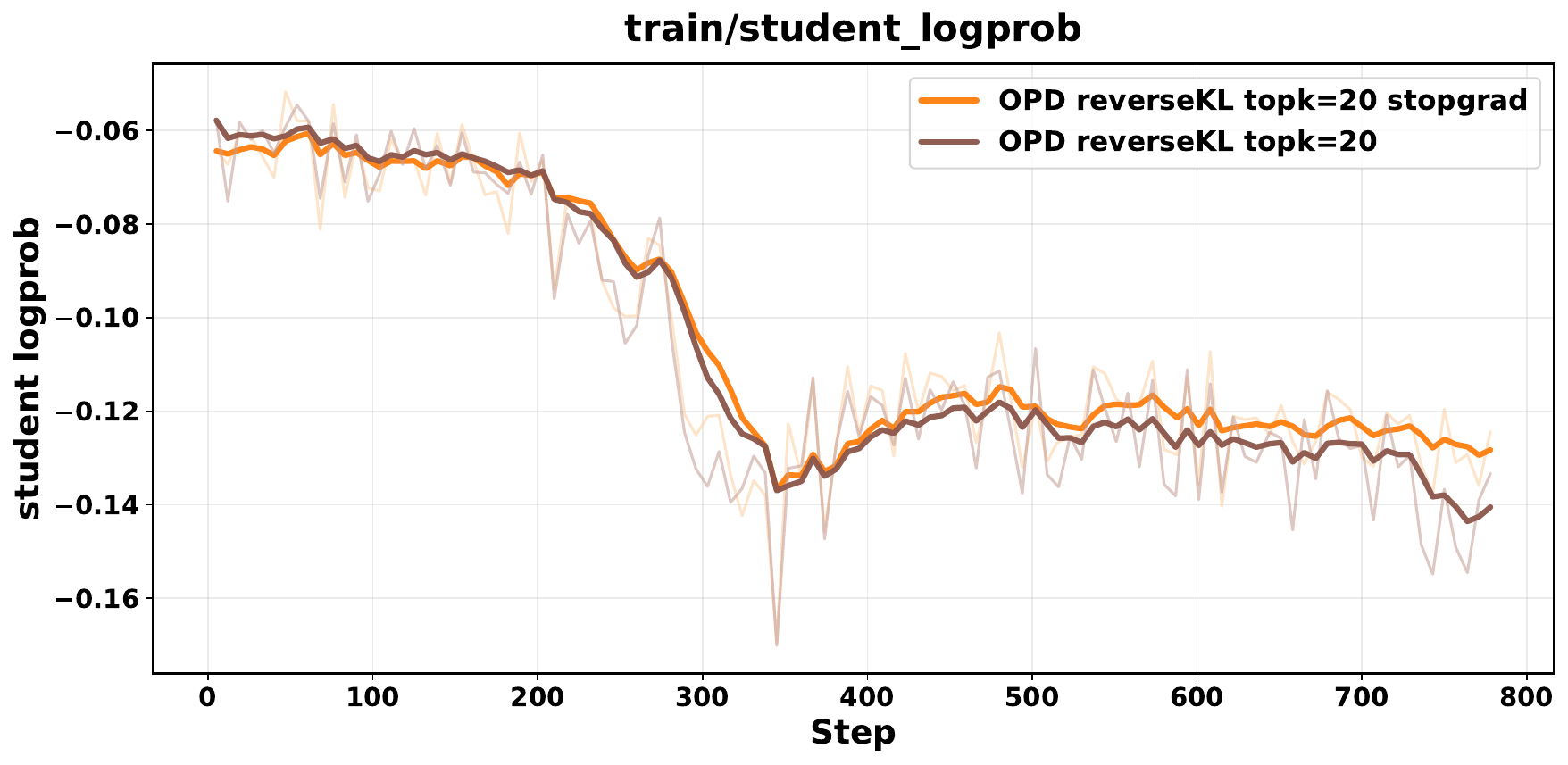}
        \label{fig:8b_char}
    \end{subfigure}
    \hfill
    \begin{subfigure}[t]{0.24\textwidth}
        \centering
        \includegraphics[width=\textwidth]{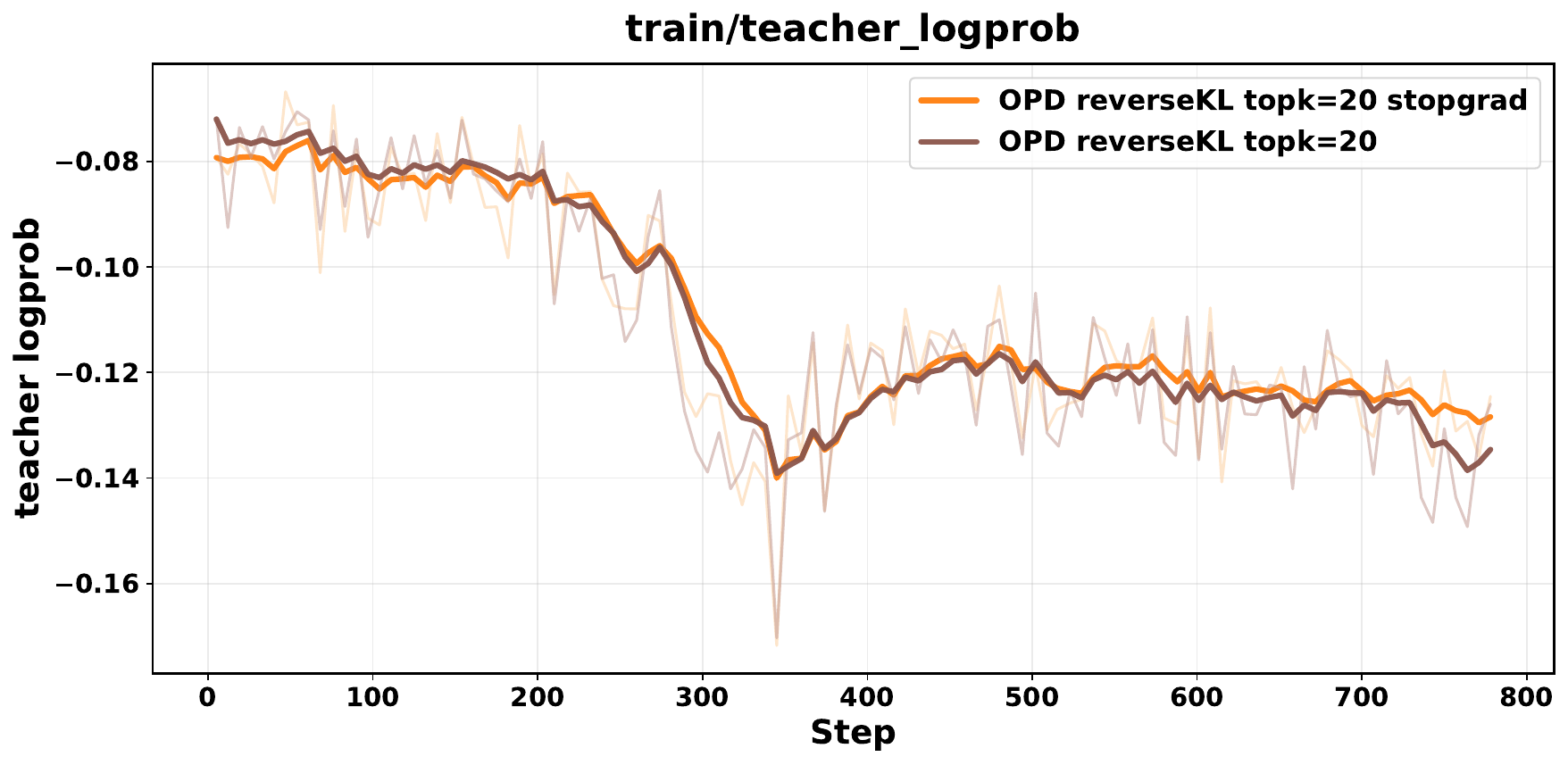}
        \label{fig:8b_emo}
    \end{subfigure}
    \caption{Teacher: Qwen3-1.7B-GRPO (nothink), Student: Qwen3-1.7B (nothink), training data: DAPO. reverse KL with stopgrad$\big($logprob($\pi_S$)$\big)$ v.s. reverse KL, TopK=20.}
    \label{fig:top20reverseKLexp}
\end{figure}

As shown in Figure~\ref{fig:top20reverseKLexp}, simply increasing $K$ from $5$ to $20$ does not eliminate the collapse of reverse KL: the Top20 objective without stop-gradient still shows a similar collapse tendency to Figure~\ref{fig:top5exp}. In contrast, the stop-gradient version remains stable.

\subsection{Evaluation Biases in OPD}

There are two main evaluation biases that can lead to misleading conclusions
about OPD.

First, the max validation response length can affect the
measured accuracy in math reasoning. OPD often encourages longer
or more repetitive reasoning traces. If the maximum generation length during
validation is too small, responses may be truncated before the model
reaches the final answer. In this case, an observed increase or decrease in
accuracy may not reflect a genuine change in reasoning ability, but rather a
change in whether the model is able to finish its response under the evaluation
length budget.

Second, comparisons between OPD and RL methods can be biased by
focusing only on early-stage sample efficiency. OPD may improve faster at the
beginning of training, while GRPO often improves more
slowly. However, in our observations, GRPO can continue improving for longer and
eventually reach higher performance. Thus, claiming that OPD is
more sample-efficient can be misleading if one ignores the performance ceiling.

\subsection{Teacher Signal Analysis (On Policiness)}

\begin{figure}[t]
    \centering
    \begin{subfigure}[t]{0.48\textwidth}
        \centering
        \includegraphics[width=\textwidth]{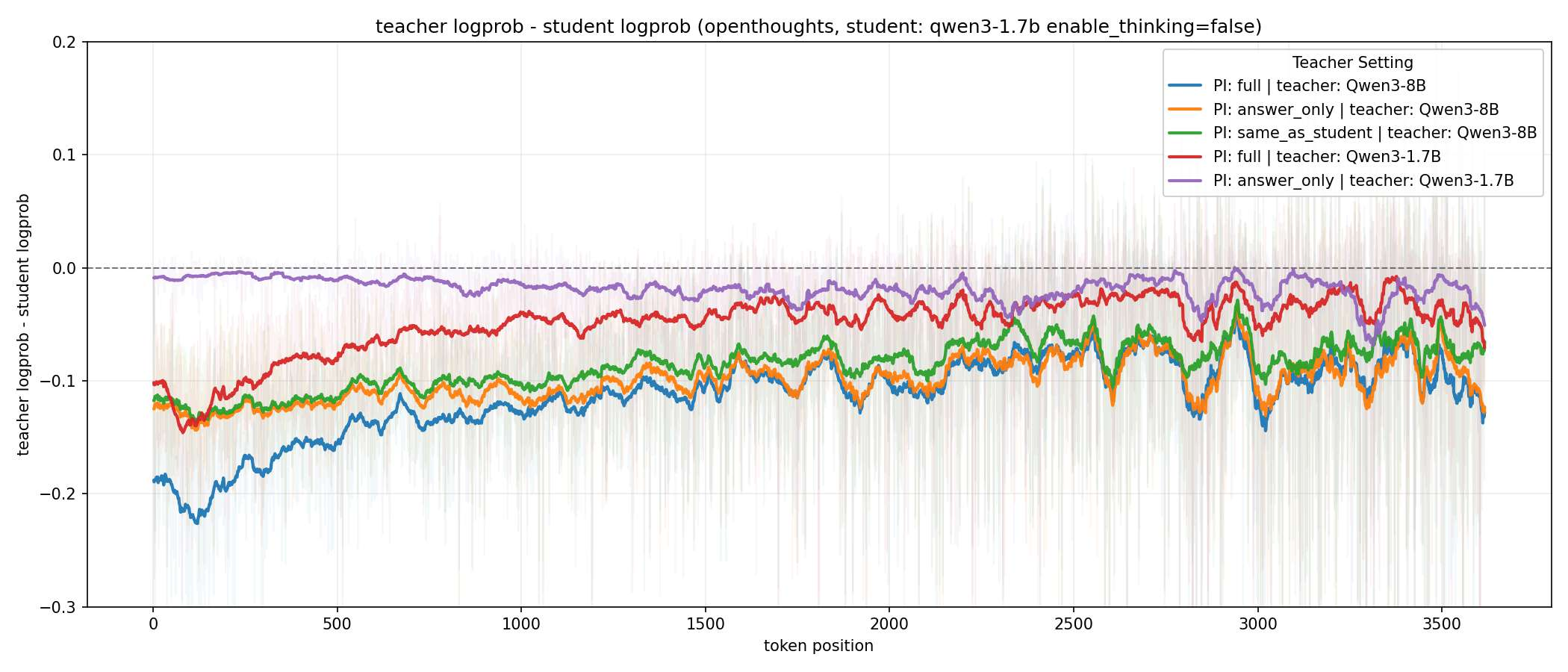}
        \caption{$\Delta$logprob per token position, no think}
        \label{fig:first}
    \end{subfigure}
    \hfill
    \begin{subfigure}[t]{0.48\textwidth}
        \centering
        \includegraphics[width=\textwidth]{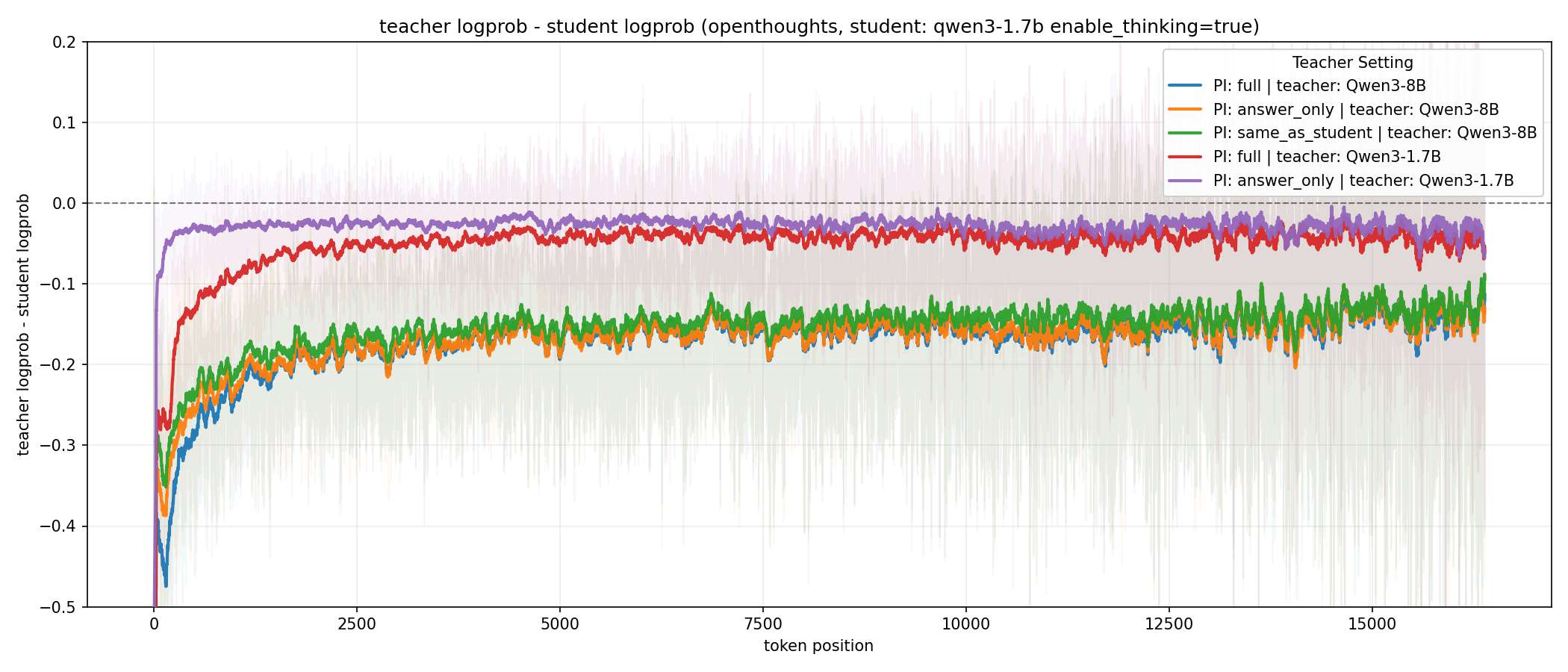}
        \caption{$\Delta$logprob per token position, think}
        \label{fig:second}
    \end{subfigure}
    
    \centering
    \begin{subfigure}[t]{0.48\textwidth}
        \centering
        \includegraphics[width=\textwidth]{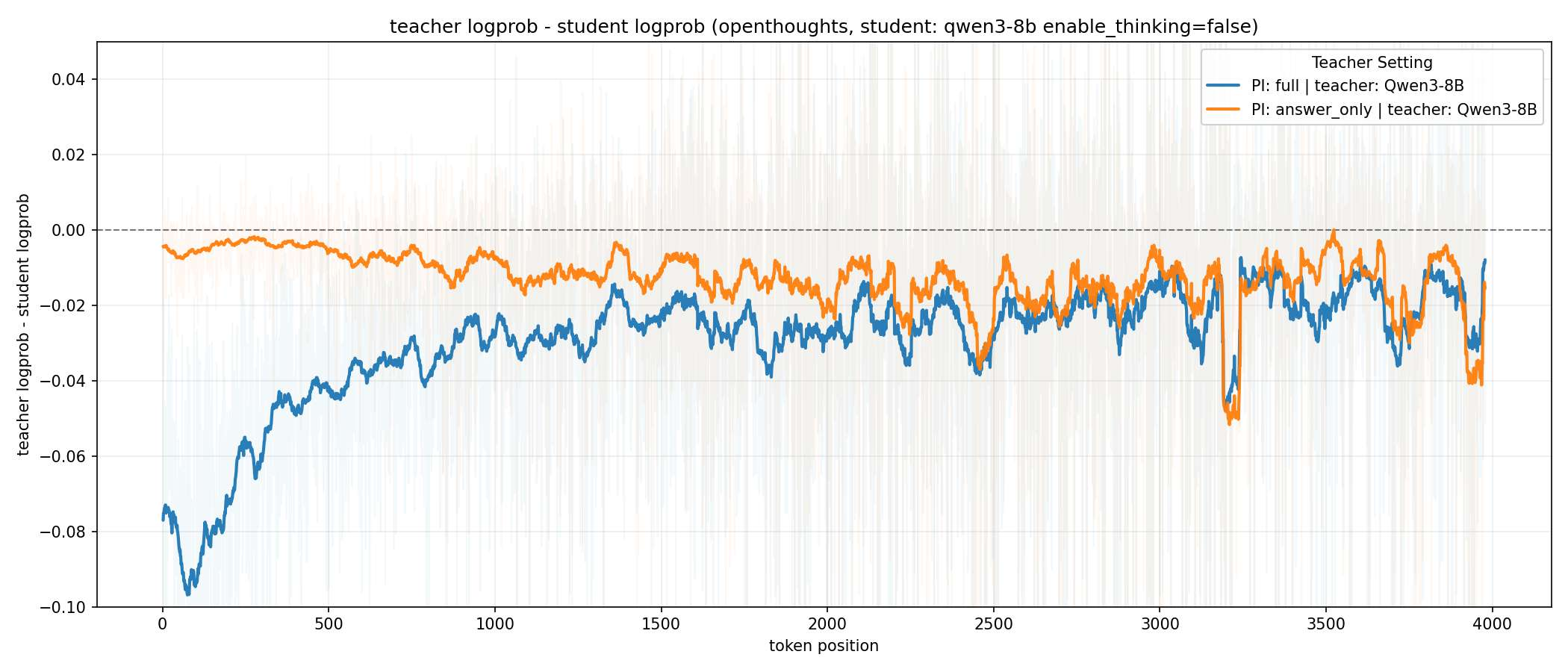}
        \caption{Student: Qwen3-8B}
        \label{fig:second}
    \end{subfigure}
    \vspace{1mm}
    \caption{Comparison of teacher signal on responses generated by different student models. $\Delta$logprob shows length-skewed distribution. The experiment used openthoughts \cite{guha2025openthoughtsdatarecipesreasoning}. The Qwen3-8B teacher exhibits stronger signal on Qwen3-1.7B responses than on Qwen3-8B self-generated responses. }
    \label{app:teacher_signal}
\end{figure}

As shown in Figure~\ref{app:teacher_signal}, the Qwen3-8B teacher provides stronger supervision for responses generated by Qwen3-1.7B (Figure~\ref{app:teacher_signal}a) than for its own self-generated responses (Figure~\ref{app:teacher_signal}b). This suggests that the teacher signal becomes more pronounced when there is a larger capability gap between the teacher and the student, whereas self-distillation yields a weaker signal overall. In addition, the supervision signal is stronger at earlier token positions. At later positions, different PI types lead to much smaller differences in $\Delta$ log probability, indicating that early-token supervision plays a more important role in distillation.

\subsection{\texorpdfstring{$\Delta \log p$}{Delta log p} is correctness-skewed}

\begin{figure}[t]
    \centering
    \begin{subfigure}[t]{0.48\textwidth}
        \centering
        \includegraphics[width=\textwidth]{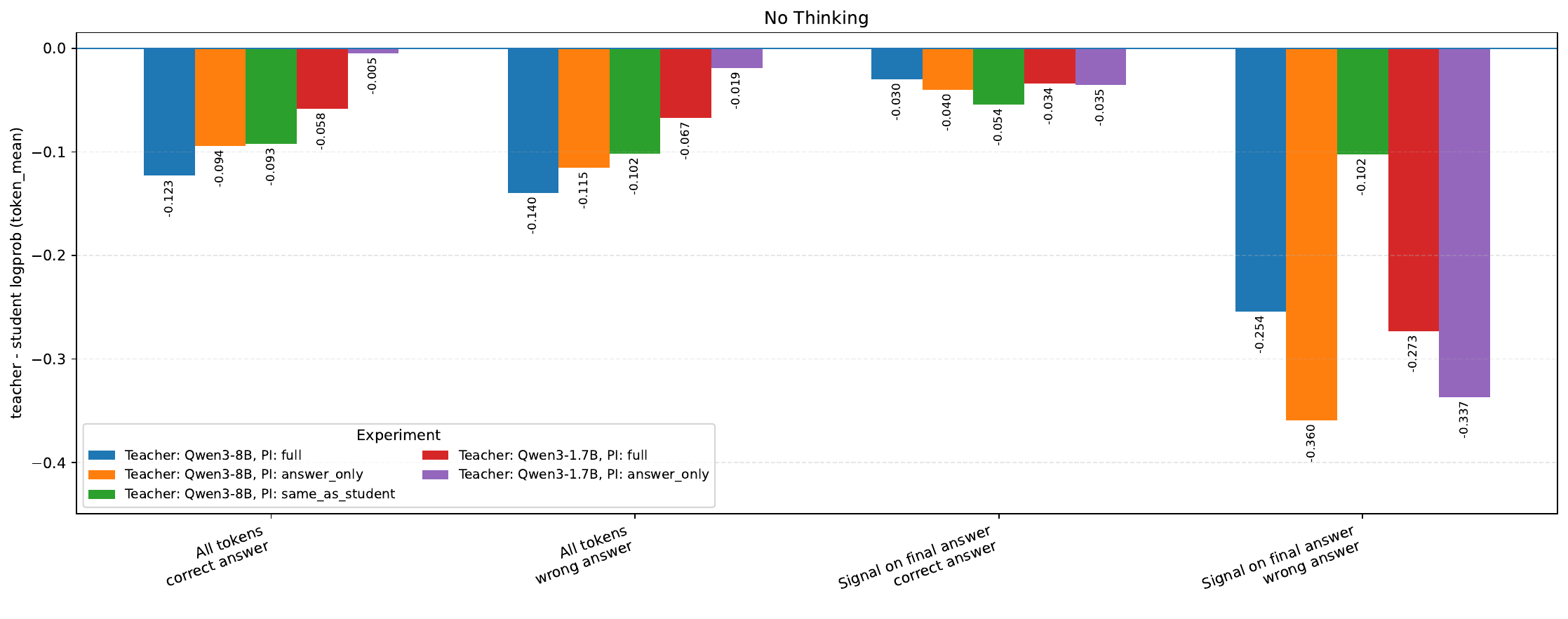}
        \caption{$\Delta$logprob by answer correctness, no think}
        \label{fig:three}
    \end{subfigure}
    \hfill
    \begin{subfigure}[t]{0.48\textwidth}
        \centering
        \includegraphics[width=\textwidth]{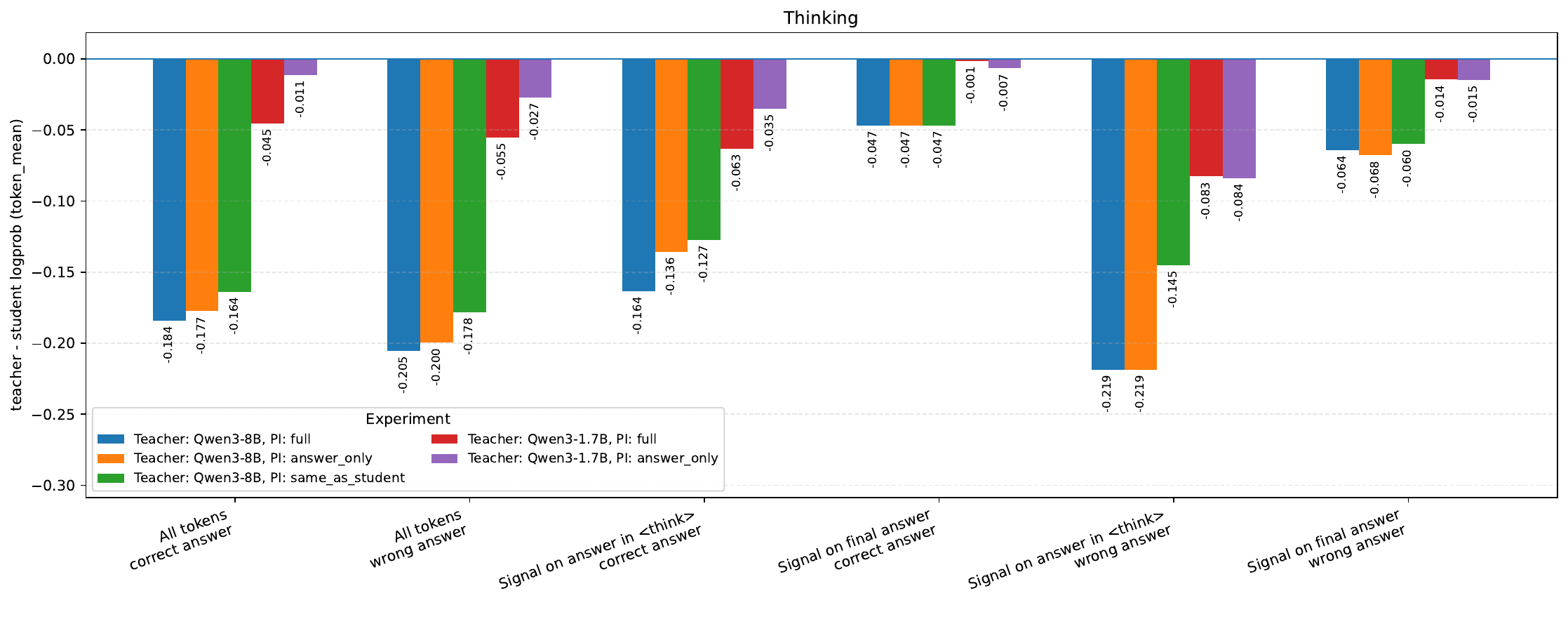}
        \caption{$\Delta$ logprob by answer correctness, think}
        \label{fig:four}
    \end{subfigure}
    \vspace{-1mm}
    \caption{Comparison of token-level KL supervision distributions for correct and incorrect student responses. Incorrect trajectories has stronger supervision, while correct trajectories has a weaker supervision.}
    \label{fig:correctnessskeweddistribution}
\end{figure}

Figure~\ref{fig:correctnessskeweddistribution} shows that token-level KL supervision is substantially weaker on correct trajectories than on incorrect trajectories. The experiments follow the same setup as in Figure~\ref{app:teacher_signal}.

\subsection{Visualizing token-level supervision on an example response}

\begin{figure}[t]
    \centering
    \begin{subfigure}[t]{0.48\textwidth}
        \centering
        \includegraphics[width=\textwidth]{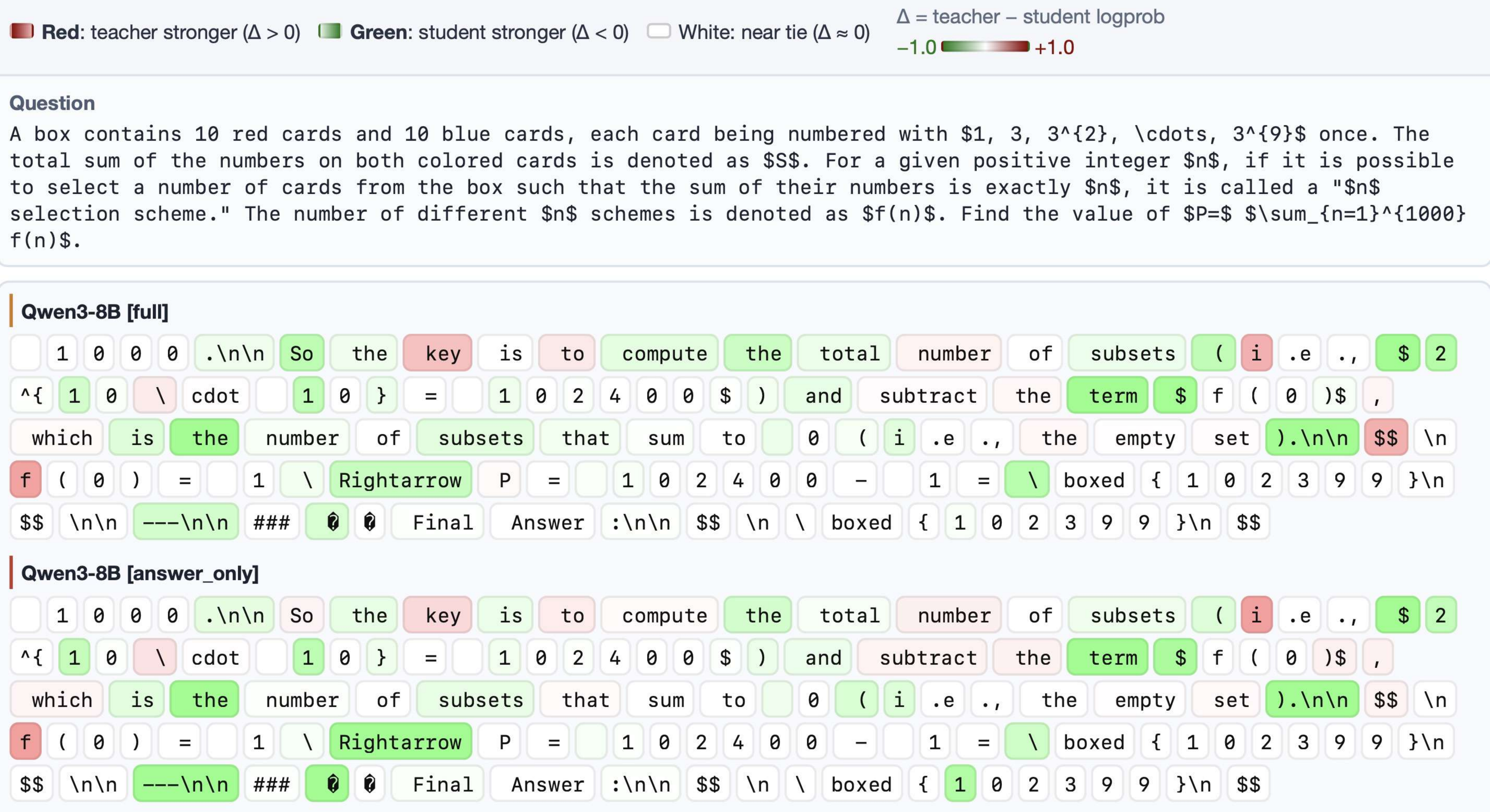}
        \label{fig:first}
    \end{subfigure}
    \hfill
    \begin{subfigure}[t]{0.48\textwidth}
        \centering
        \includegraphics[width=\textwidth]{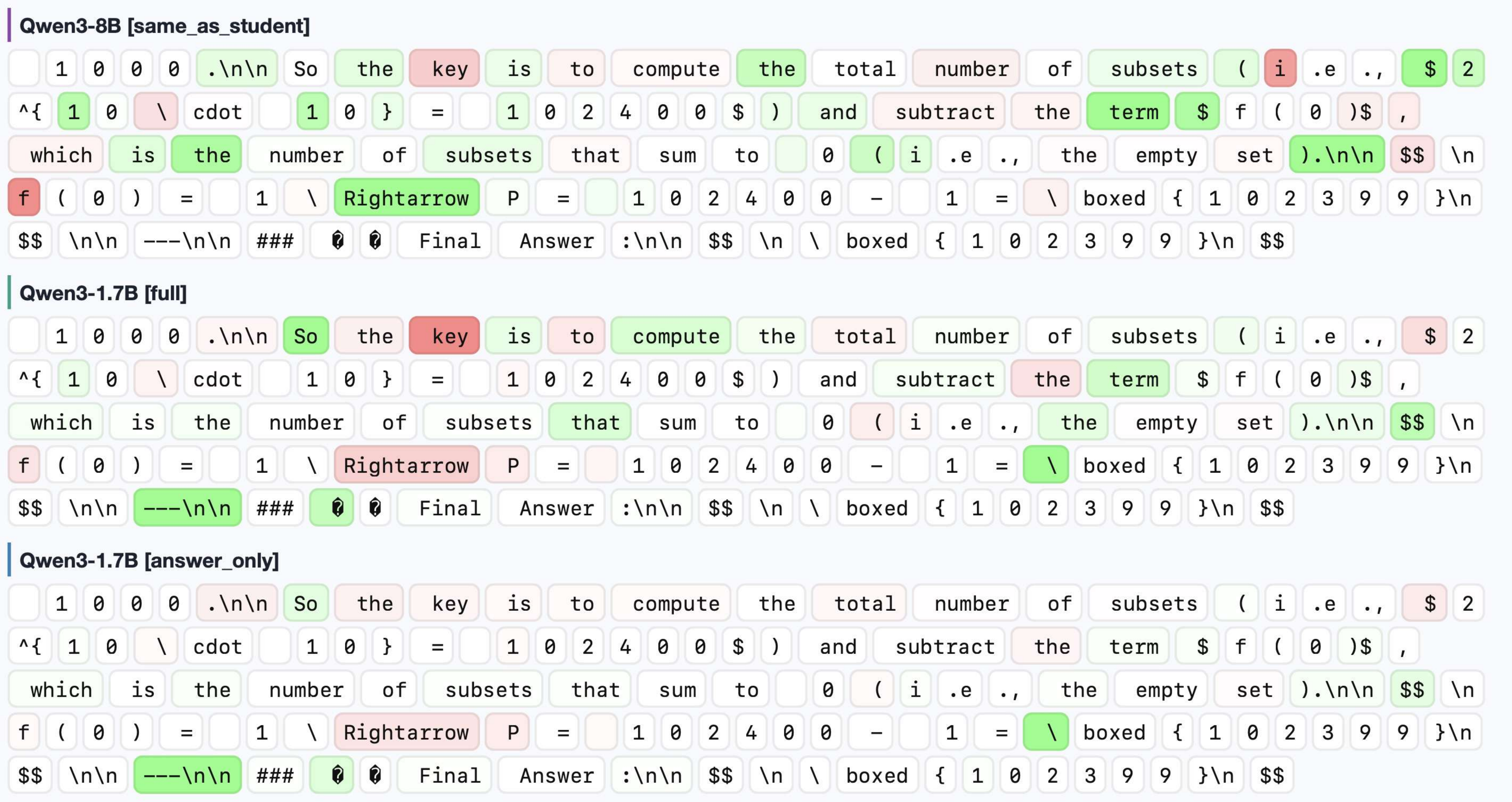}
        \label{fig:second}
    \end{subfigure}
    \vspace{-2mm}
    \caption{We show token-level heatmap of $\Delta$logprob on last 128 tokens. The experiment is based on openthoughts \cite{guha2025openthoughtsdatarecipesreasoning}, we show an example question. PI strengthens supervision for the same teacher, yet the sampled-token supervision distribution is based more on teacher capability (as shown in the figure, 3 experiments using Qwen3-8B teacher show similar distribution, while 2 experiments using Qwen3-1.7B teacher show another distribution). }
    \label{fig:tokenlevelvisual}
\end{figure}

Figure\ref{fig:tokenlevelvisual} shows the token-level supervision of different teacher designs. While PI can refine the signal, the fundamental distribution of what a student learns is dictated by the teacher's scale and reasoning depth. This underscores the importance of scaling teacher models to provide high-quality supervision.

\subsection{Experimental Results on General Reasoning Tasks}

\begin{figure}[h]
    \centering
    \includegraphics[width=0.9\linewidth]{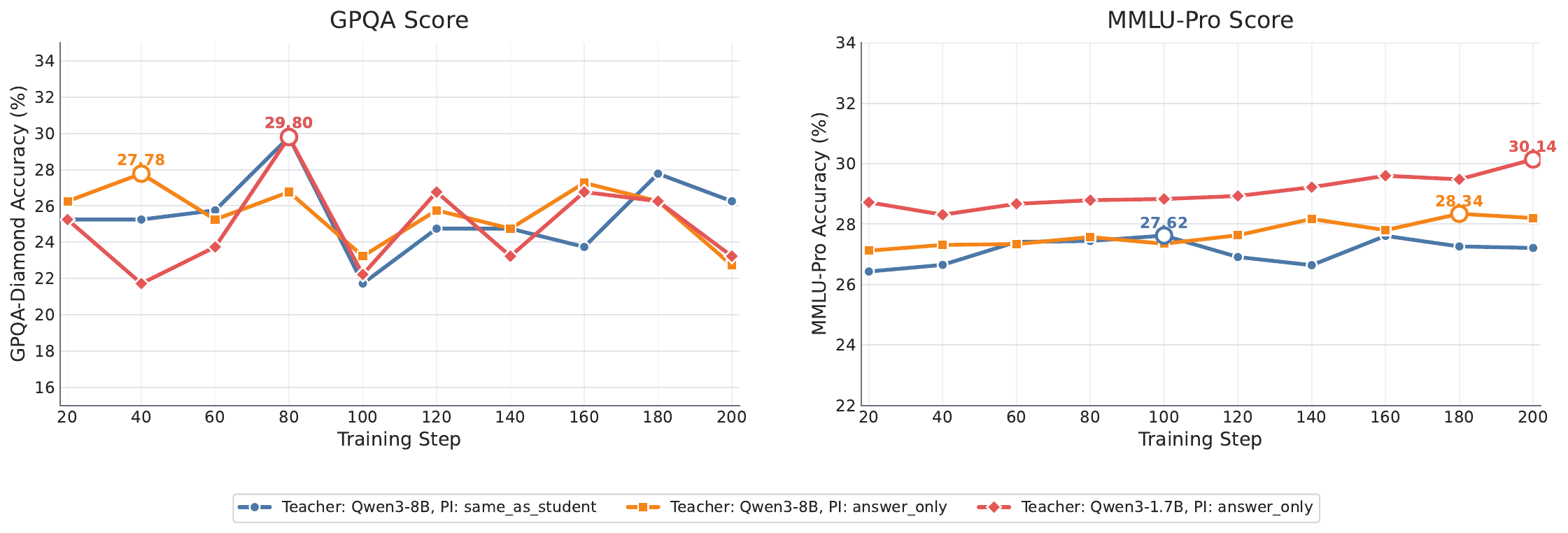}
    \caption{
    General reasoning results of OPD training. 
    The experiment uses the Science subset of Mixture of Thoughts. 
    OPD does not bring consistent improvements on GPQA-Diamond or MMLU-Pro.
    }
    \label{fig:general-reasoning-appendix}
\end{figure}

We further evaluate OPD on general reasoning tasks. We use Qwen3-1.7B as the student model and Qwen3-8B as the teacher model, with the thinking mode disabled for both models. Training is conducted on the Science subset of Mixture of Thoughts, and evaluation is performed on GPQA-Diamond and MMLU-Pro. As shown in Figure~\ref{fig:general-reasoning-appendix}, OPD does not lead to consistent or substantial improvements over the student model across training steps. The performance fluctuates on GPQA-Diamond and remains only marginally improved on MMLU-Pro, suggesting that a stronger teacher does not necessarily provide effective supervision in this setting.

\begin{figure}[h]
    \centering
    \includegraphics[width=0.9\linewidth]{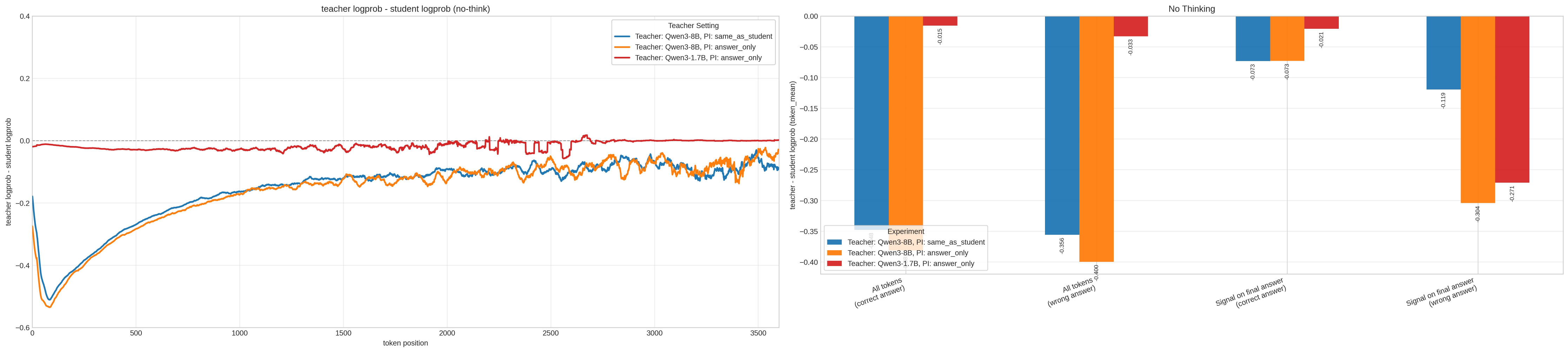}
    \caption{
    Comparison of teacher signals on general reasoning trajectories. 
    $\Delta\log p$ shows a length-skewed distribution: the supervision signal is stronger at early tokens and gradually weakens. 
    Incorrect responses receive stronger teacher supervision than correct responses, especially on final-answer tokens.}
    \label{fig:general-reasoning-appendix2}
\end{figure}

To better understand this failure, we further analyze the token-level teacher--student log-probability gap on the generated trajectories. Figure~\ref{fig:general-reasoning-appendix2} shows that the supervision signal is strongly skewed toward early tokens and gradually weakens as the response becomes longer. Moreover, incorrect responses receive stronger teacher supervision than correct responses, especially on final-answer tokens. This indicates that OPD mainly provides corrective signals for erroneous trajectories, while offering limited useful supervision for already plausible or correct reasoning paths. This phenomenon is highly consistent with our previous observations on mathematical reasoning tasks.

\subsection{OPSD Fails on Persuasion Tasks}

\begin{figure*}[t]
    \centering
    \begin{subfigure}[t]{0.9\textwidth}
        \centering
        \includegraphics[width=0.32\textwidth]{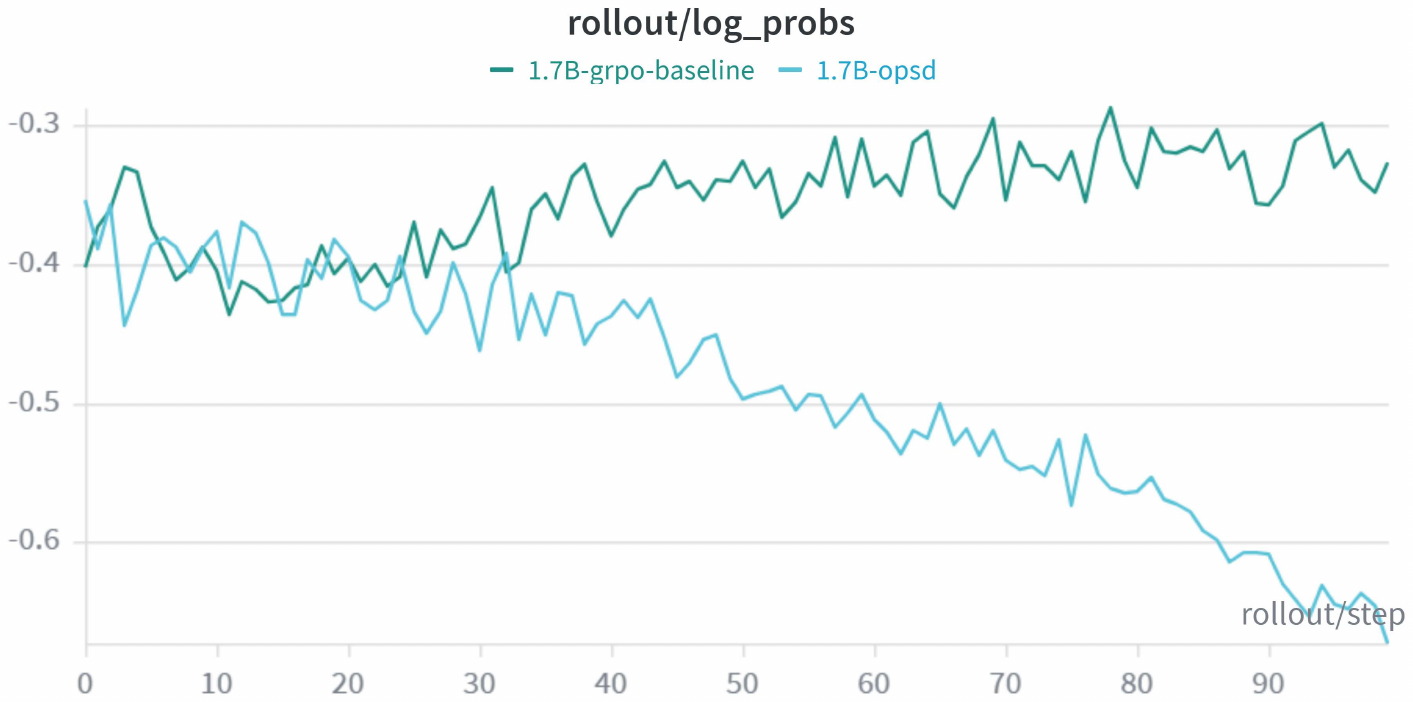}
        \hfill
        \includegraphics[width=0.32\textwidth]{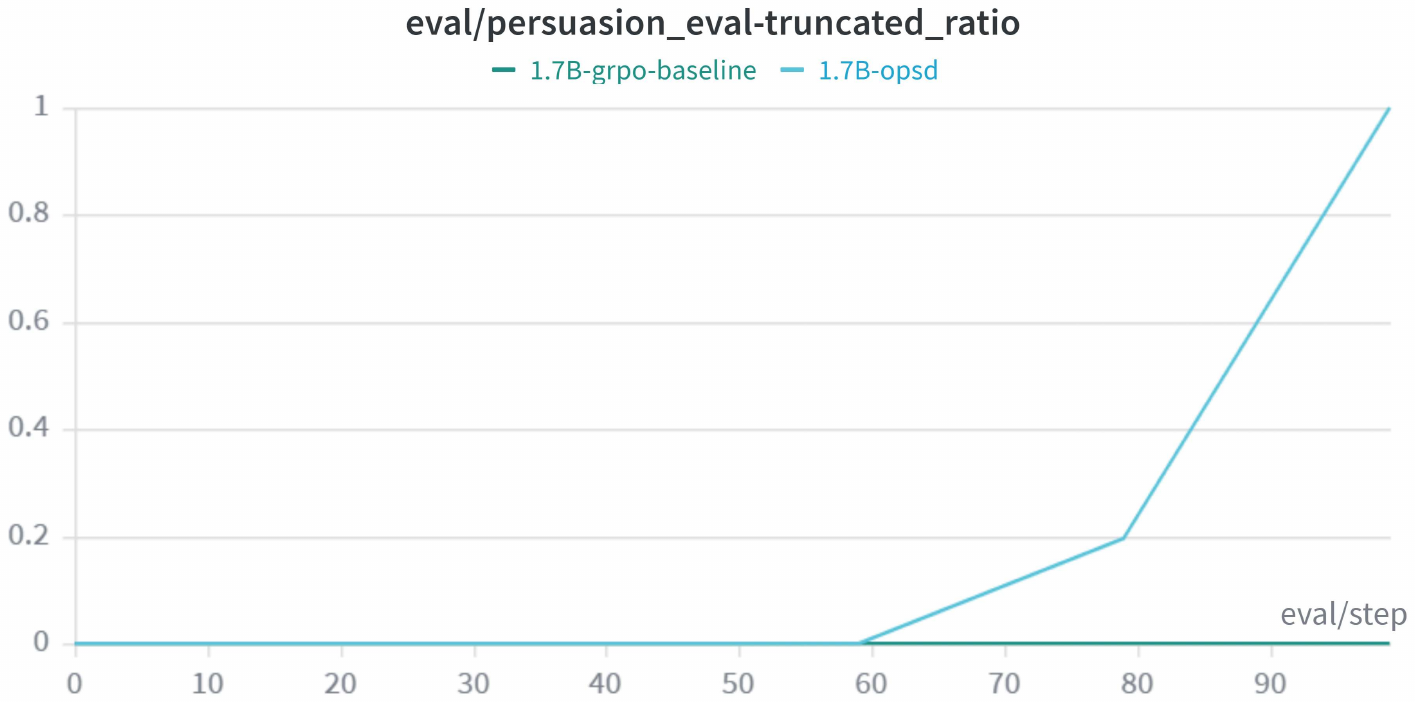}
        \hfill
        \includegraphics[width=0.32\textwidth]{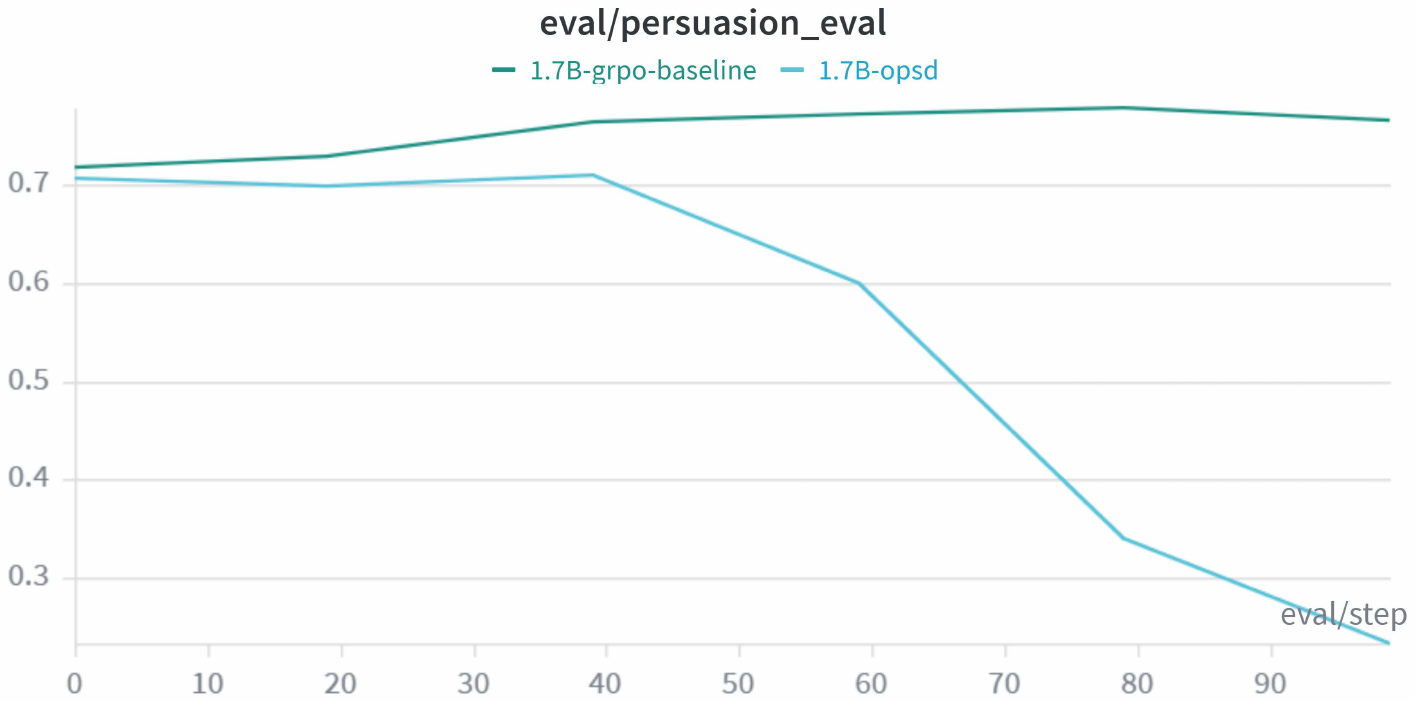}
        \caption{Qwen-1.7B on Persuasion for Good}
        \label{fig:1.7b}
    \end{subfigure}
    \begin{subfigure}[t]{0.9\textwidth}
        \centering
        \includegraphics[width=0.32\textwidth]{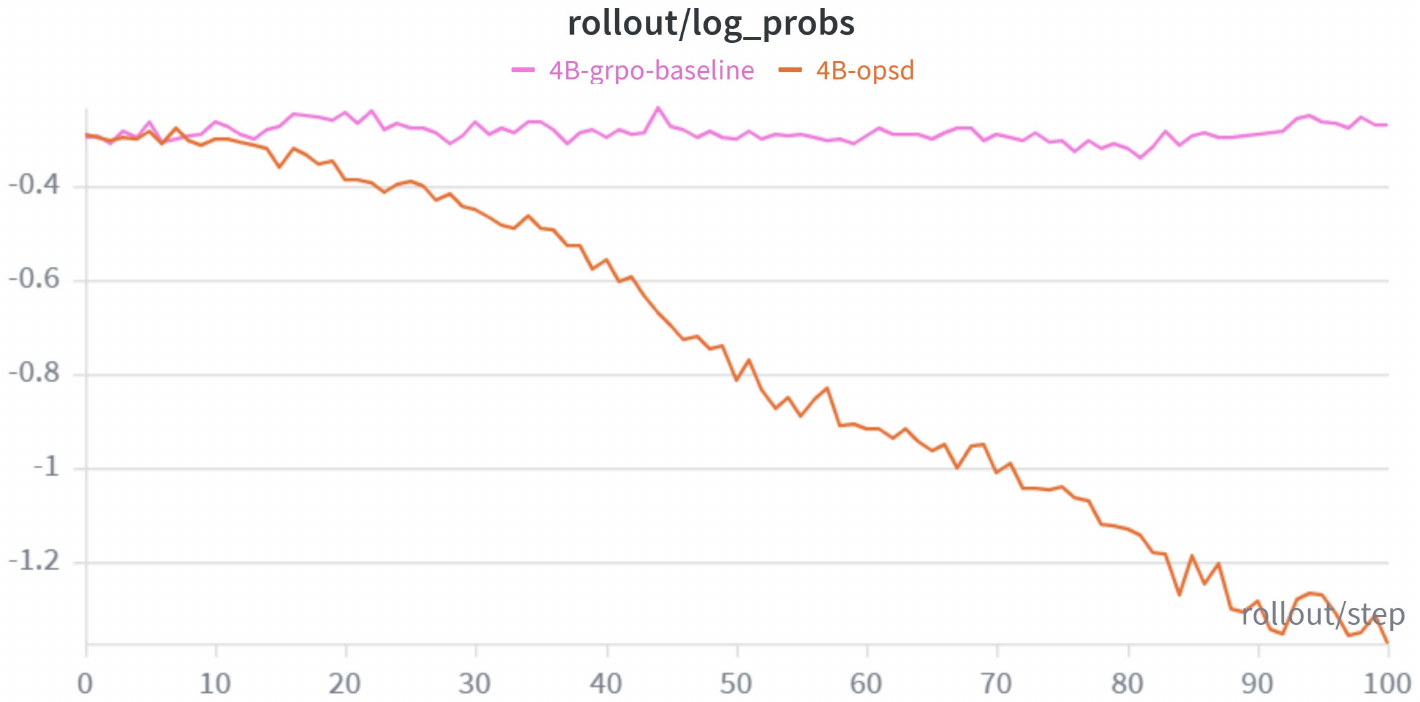}
        \hfill
        \includegraphics[width=0.32\textwidth]{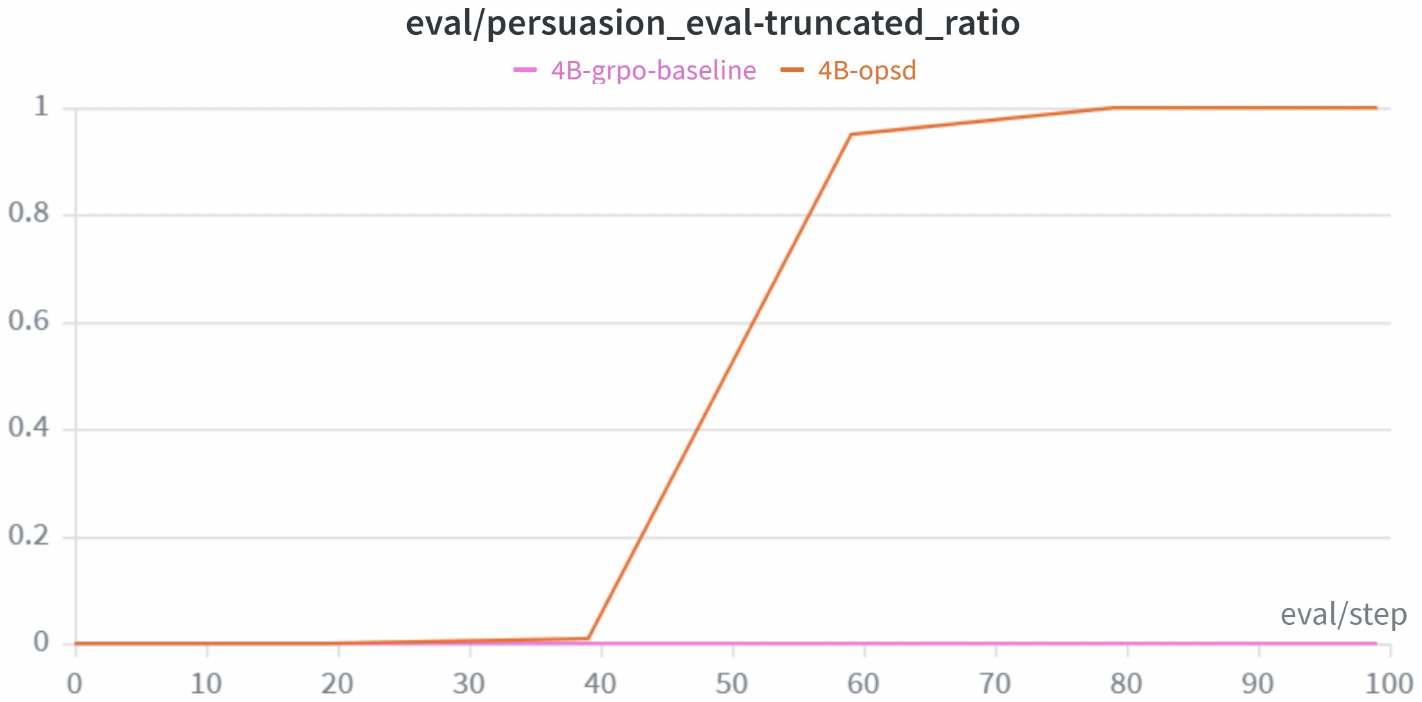}
        \hfill
        \includegraphics[width=0.32\textwidth]{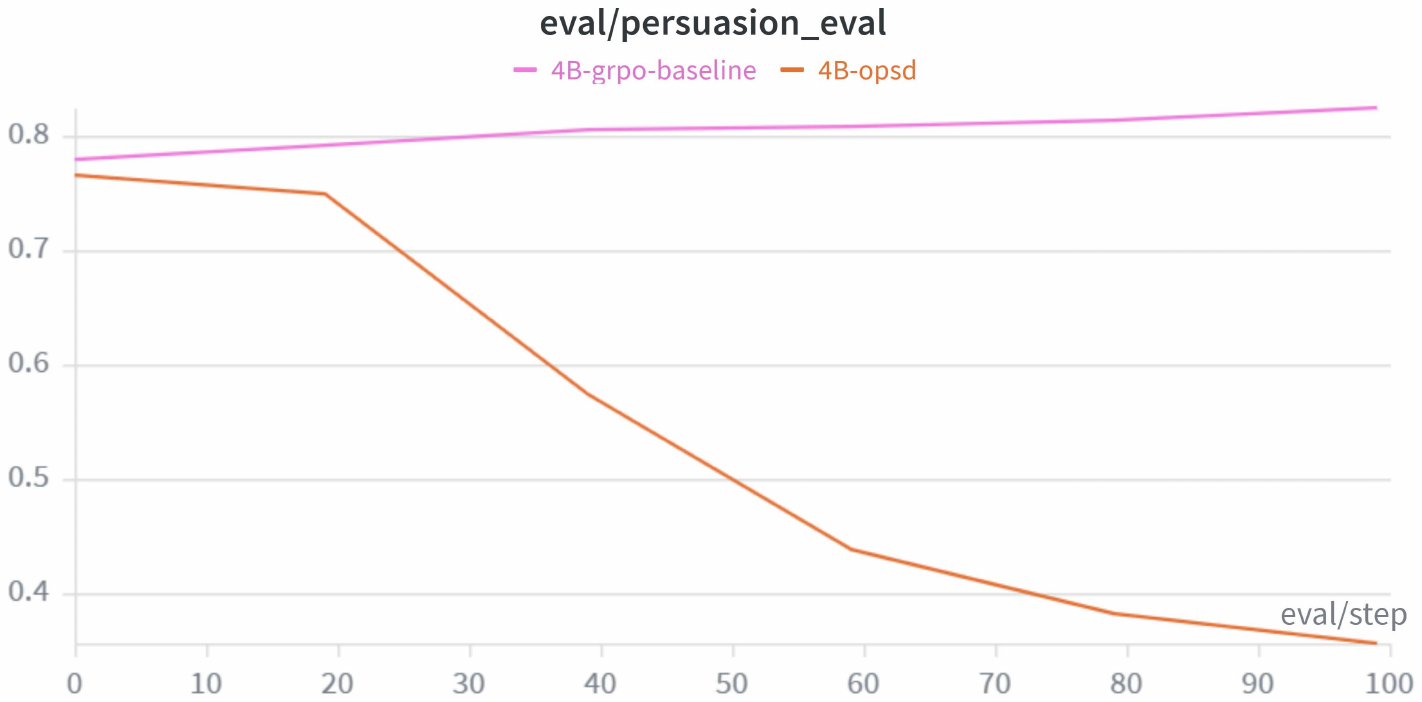}
        \caption{Qwen-4B on Persuasion for Good}
        \label{fig:4b}
    \end{subfigure}
    \caption{Next-token log probs (left), truncated ratio (middle) and evaluation results (right) curves for GRPO and OPSD on Persuasion for Good dataset, using Qwen-1.7B (top) and Qwen-4B (bottom) as student models.}
    \label{fig:new_layout}
\end{figure*}

We evaluate OPSD on the Persuasion for Good dataset~\cite{wang2019persuasion}, where a Persuader engages in multi-turn dialogue to convince a Partner to donate to the charitable organization “Save the Children”. The privileged information includes the persuadee's background and personality, along with high-level persuasion strategies. We compare OPSD against GRPO using Qwen3-1.7B and Qwen3-4B as student models. Figure~\ref{fig:new_layout} reveals that while GRPO exhibits stable and gradually improving performance for both 1.7B and 4B students, OPSD shows rapid degradation shortly after training begins. For the 1.7B model, OPSD achieves competitive performance in the early stage but collapses after approximately 20--30 steps, with evaluation scores dropping sharply thereafter. This instability is even more pronounced for the 4B model, where both reward and evaluation performance deteriorate more severely. Meanwhile, as shown in the middle column, OPSD quickly drives the truncation ratio to nearly $1.0$, suggesting a strong tendency toward excessively long generations, which in turn leads to ineffective or degenerate outputs. The left column further reveals that under OPSD, the student log-probabilities steadily decrease over the course of training, indicating the model becomes increasingly uncertain about its generated tokens. This effect is more pronounced in the 4B setting, where log-probabilities decline more rapidly and reach lower values, consistent with the more severe performance collapse observed in evaluation.

\subsection{Thinking Mode Hacking in OPSD}
\label{app:thinkinghack}

\begin{figure}[t]
    \centering
    \includegraphics[width=0.4\textwidth]{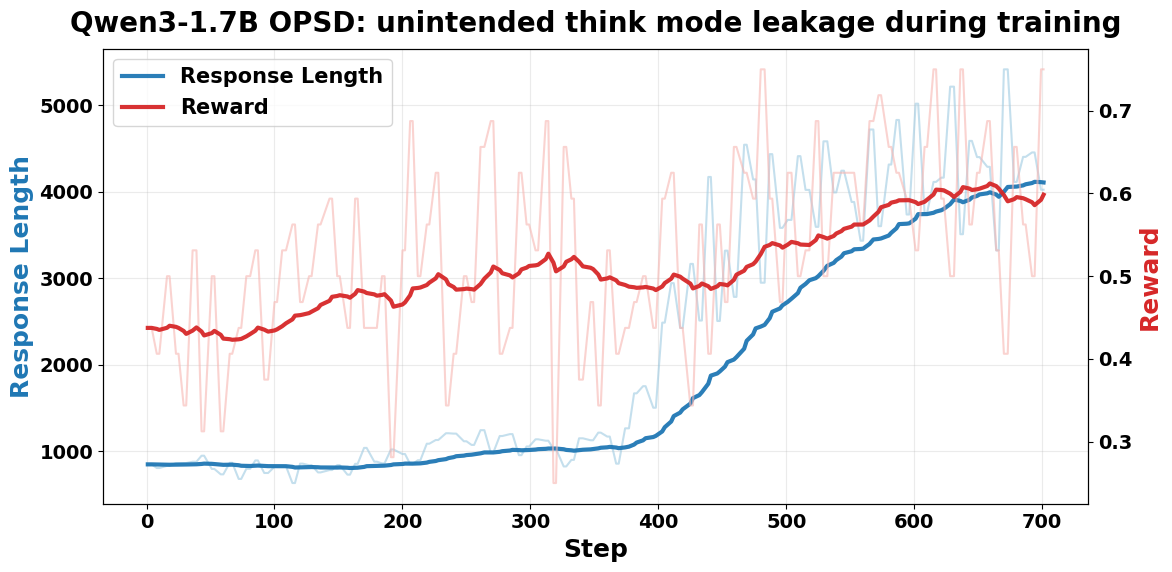}
    \label{fig:first}
    \caption{An example of \emph{thinking mode hacking} during OPSD. The student is trained with thinking mode disabled, while the teacher is queried with reasoning enabled. During training, the student gradually learns to emit explicit thinking-mode control tokens in its response, even though such tokens are not intended to appear at inference time.}
    \label{app:fighack}
\end{figure}

Following prior work~\cite{zhao2026selfdistilledreasoneronpolicyselfdistillation}, we consider a setup where the student is trained with thinking mode disabled, while the teacher is prompted with reasoning enabled. This experiment uses Qwen3-1.7b and is trained on dapo \cite{yu2025dapoopensourcellmreinforcement}. We observe a failure mode that we term \emph{thinking mode hacking} (Figure~\ref{app:fighack}): although the student is intended to produce only the final answer, it may instead partially imitate the teacher's reasoning-mode protocol and spontaneously trigger thinking mode during generation. In particular, the output can contain malformed patterns such as \texttt{<think> \ldots </think> \ldots <think>}, where the final \texttt{<think>} is spurious. This suggests that the student may internalize not only the target answer distribution but also control tokens associated with hidden reasoning. As a result, distillation under mismatched thinking-mode configurations can lead to malformed outputs and unintended activation of thinking mode.

\subsection{OPSD Response Length Collapse}
\label{app:opsdcollapse}

Using DAPO-Math-17k with only the final answer as a hint, Qwen3-1.7B self-distillation exhibits a response collapse.  The student may learn to directly output the answer, bypassing the intermediate reasoning process.

\begin{promptbox}
\begin{Verbatim}[breaklines=true, breakanywhere=true, fontsize=\small]
\<|im_start|>user\nI had $\\$30$ in allowance money and spent it as indicated in the pie graph shown. How many dollars did I spend on burgers?\n\n[asy]\nsize(1K);\npair A, B, C, D, O, W, X, Y, Z;\nO=(0,0);\nA=(.707,.707);\nB=(-.966,.259);\nC=(-.707,-.707);\nD=(.342,-.940);\ndraw(Circle(O, 1));\ndraw(O--A);\ndraw(O--B);\ndraw(O--C);\ndraw(O--D);\nW=(-.1,.5);\nlabel("Movies", W, N);\nlabel("$\\frac{1}{3}$", W, S);\nX=(-.55, 0);\nlabel("Burgers", X, S);\nY=(-.17,-.7);\nlabel("Ice Cream", Y, N);\nlabel("$\\frac{1}{5}$", Y, S);\nZ=(.5, -.15);\nlabel("Music", Z, N);\nlabel("$\\frac{3}{10}$", Z, S);\n[/asy]\n\nPlease reason step by step, and put your final answer within \\boxed{}.<|im_end|>\n
\end{Verbatim}
\end{promptbox}

\begin{responsebox}
\begin{Verbatim}[breaklines=true, breakanywhere=true, fontsize=\small]
<|im_start|>assistant\n<think>\n\n</think>\n\n$$\n\\boxed{7}\n$$<|im_end|>
\end{Verbatim}
\end{responsebox}

\subsection{OPD Response Length Explosion - step 700}
\label{app:opdexplode119}

This example illustrates response length explosion in OPD, where a Qwen3-1.7B student is distilled from a Qwen3-8B teacher. The response becomes unusually verbose, with the word “but” appearing 10 times.

\begin{promptbox}
\begin{Verbatim}[breaklines=true, breakanywhere=true, fontsize=\small]
<|im_start|>user\nAn object in the plane moves from one lattice point to another. At each step, the object may move one unit to the right, one unit to the left, one unit up, or one unit down. If the object starts at the origin and takes a ten-step path, find the number of different points that could be the final point.\n\nPlease reason step by step, and put your final answer within \\boxed{}.<|im_end|>\n
\end{Verbatim}
\end{promptbox}

\begin{responsebox}
\begin{Verbatim}[breaklines=true, breakanywhere=true, fontsize=\small]
<|im_start|>assistant\n<think>\n\n</think>\n\nWe are given a problem where an object starts at the origin (0,0) and takes exactly 10 steps, each step being one unit in one of the four cardinal directions (right, left, up, down). The question is asking for the number of different points that could be the final point after such a path.\n\n---\n\n### First, let's understand the movement.\n\nAt each step, the object can move:\n\n- Right (R): increases the x-coordinate by 1\n- Left (L): decreases the x-coordinate by 1\n- Up (U): increases the y-coordinate by 1\n- Down (D): decreases the y-coordinate by 1\n\nSo, each step changes the position by one unit in one of the four directions.\n\nAfter 10 steps, the object will be at some point (x, y), where x and y are integers. The key is to find all possible (x, y) such that the total number of steps in the positive and negative directions leads to that point.\n\nBut since the total number of steps is 10, we can also think in terms of the net movement in the x and y directions.\n\nLet me define:\n\nLet's say the object makes:\n\n- $ a $ steps to the right (R)\n- $ b $ steps to the left (L)\n- $ c $ steps up (U)\n- $ d $ steps down (D)\n\nThen, the total number of steps is $ a + b + c + d = 10 $.\n\nNow, the final position (x, y) will be:\n\n- $ x = a - b $\n- $ y = c - d $\n\nSo the net movement in the x-direction is $ a - b $, and the net movement in the y-direction is $ c - d $.\n\nSo, for the final position to be (x, y), we must have:\n\n- $ x = a - b $\n- $ y = c - d $\n\nBut since $ a + b + c + d = 10 $, and $ a, b, c, d $ are all non-negative integers (as they represent the number of steps in each direction), we can think of this as:\n\nLet's denote $ a + b = x + b $, but maybe that's not the best approach.\n\nAlternatively, we can think of the net x and y movement. Let me define:\n\nLet's denote the net steps in the x-direction as $ x' = x - 0 = a - b $, and similarly for the y-direction $ y' = c - d $.\n\nThen, the total number of steps is:\n\n- $ a + b + c + d = 10 $\n- But $ a + b = \\text{number of right and left steps} = x' + b \\text{ (if we consider that)} $... Wait, maybe it's easier to think in terms of:\n\nLet's think of the net movement as follows:\n\nLet's define $ a + b = A $, the number of right and left steps. Then $ c + d = B $, the number of up and down steps.\n\nThen, the total number of steps is $ A + B = 10 $.\n\nBut also, the net x-position is $ a - b = x $, and the net y-position is $ c - d = y $.\n\nBut note that $ a + b = A $, so $ a = A - b $. So, substituting into $ a - b = x $:\n\n$ (A - b) - b = x \\Rightarrow A - 2b = x \\Rightarrow 2b = A - x \\Rightarrow b = (A - x)/2 $\n\nBut since $ a $ and $ b $ must be non-negative integers, this means $ A - x $ must be even and non-negative, so $ A \\geq x $, and $ A - x $ is even.\n\nBut $ A = a + b $, so $ A $ must be a non-negative integer such that $ A + B = 10 $.\n\nBut perhaps it's better to think in terms of the total number of steps in each direction. Since the total number of steps is 10, and each step is a move in one of the four directions, the net movement in x and y directions must be such that:\n\nLet's consider that the number of right and left steps must satisfy:\n\n- The net x-move is $ a - b = x $\n- The number of right and left steps is $ a + b = A $, which is an integer between 0 and 10.\n\nSimilarly, for the y-direction:\n\n- The net y-move is $ c - d = y $\n- The number of up and down steps is $ c + d = B $, between 0 and 10.\n\nBut also, since the total number of steps is 10, we have $
\end{Verbatim}
\end{responsebox}

\subsection{OPD Response Length Explosion - step 1000}
\label{app:opdexplode149}

This example illustrates response length explosion in OPD, where a Qwen3-1.7B student is distilled from a Qwen3-8B teacher. The response only contains word "maybe".

\begin{promptbox}
\begin{Verbatim}[breaklines=true, breakanywhere=true, fontsize=\small]
<|im_start|>user
Given that the complex sequence $\{z_n\}$ satisfies
\[
z_1=1,\qquad z_{n+1}=\overline{z_n}+1+ni \quad (n=1,2,3,\cdots),
\]
where $i$ is the imaginary unit and $\overline{z_n}$ denotes the complex conjugate of $z_n$, find the real part of $z_{2015}$.

Please reason step by step, and put your final answer within \boxed{}.
<|im_end|>
\end{Verbatim}
\end{promptbox}

\begin{responsebox}
\begin{Verbatim}[breaklines=true, breakanywhere=true, fontsize=\small]
<|im\_start|>assistant
<think>
 
</think>
 
 let me try to solve this problem. So, we have a complex number sequence {z_n} where z_1 is given? Wait, maybe maybe maybe maybe maybe maybe maybe maybe maybe maybe maybe maybe maybe maybe maybe maybe maybe maybe maybe maybe maybe maybe maybe maybe maybe maybe maybe maybe maybe maybe maybe maybe maybe maybe maybe maybe maybe maybe maybe maybe maybe maybe maybe maybe maybe maybe maybe maybe maybe maybe maybe maybe maybe maybe maybe maybe maybe maybe maybe maybe maybe maybe mayb...<truncated 24025 chars>
\end{Verbatim}
\end{responsebox}

\subsection{OPSD Safety Alignment System Prompt}
\label{app:opsd_alignment_prompt}

\begin{promptbox}
\begin{Verbatim}[breaklines=true, breakanywhere=true, fontsize=\small]
Follow these rules when responding:
· Do not assist with harmful or illegal actions.
· Politely refuse unsafe requests.
· Provide safe alternatives when possible.
· Remain helpful and neutral for legitimate requests.
\end{Verbatim}
\end{promptbox}

\subsection{OPSD Reasoning Compression  System Prompt.}                                                                            
  We present the prompts for reasoning compression using OPSD. 

  \textit{PI format (teacher privilege):}                                              
  \begin{promptbox}                                                                             
  \begin{Verbatim}[breaklines=true, breakanywhere=true, fontsize=\small]                        
  Solve the following math problem step by step. Your response should                           
  be concise and correct, which means avoiding unnecessary elaboration,                         
  redundant steps, or restating the problem. Focus only on the key                              
  reasoning steps needed to reach the answer. The last line of your
  response should be of the form Answer: $Answer (without quotes)                               
  where $Answer is the answer to the problem.
  [PROBLEM TEXT]      
  Remember to put your answer on its own line after "Answer:".
  \end{Verbatim}
  \end{promptbox}
                                         
  \noindent\textit{Teacher input (PI prepended):}
  \begin{promptbox}
  \begin{Verbatim}[breaklines=true, breakanywhere=true, fontsize=\small]
  Solve the following math problem step by step. Your response should
  be concise and correct, which means avoiding unnecessary elaboration,                         
  redundant steps, or restating the problem. Focus only on the key
  reasoning steps needed to reach the answer. The last line of your                             
  response should be of the form Answer: $Answer (without quotes)
  where $Answer is the answer to the problem.                                                   
  
  Compute $\displaystyle\sum_{k=1}^{100} k\cdot k!$.                                            
                  
  Remember to put your answer on its own line after "Answer:".
  \end{Verbatim}
  \end{promptbox}                                                              
  \noindent\textit{Student input (no conciseness instruction):}                                 
  \begin{promptbox}
  \begin{Verbatim}[breaklines=true, breakanywhere=true, fontsize=\small]
  Compute $\displaystyle\sum_{k=1}^{100} k\cdot k!$.                                            
  
  Remember to put your answer on its own line after "Answer:".                                  
  \end{Verbatim}  
  \end{promptbox}

\subsection{OPSD Language Style Alignment: Privileged Information Design and Model Outputs}
\label{app:style_pi}

We describe the privileged information (PI) construction for EmotionBench and CharacterBench and provide concrete input/output examples drawn from actual training runs.
In both benchmarks, the teacher receives the student's prompt augmented with a style-specific PI prefix; the student sees only the plain situational prompt with no style guidance.
The teacher prompt is constructed as:
\begin{center}
\texttt{[PI instruction] + ``\textbackslash n\textbackslash n'' + [student user content]}
\end{center}
Rewards during training are computed by an LLM judge (Gemini-2.0-Flash) that evaluates whether the student's generated response appropriately expresses the target style.

\paragraph{EmotionBench.}
EmotionBench~\cite{huang2024emotionallynumbempatheticevaluating} provides situations designed to elicit one of eight target emotions (e.g., Anger, Anxiety, Guilt).
Each situation is paired with a fine-grained emotion factor specifying the triggering cause.

\noindent\textit{PI format (teacher privilege):}
\begin{promptbox}
\begin{Verbatim}[breaklines=true, breakanywhere=true, fontsize=\small]
You are feeling [EMOTION], specifically due to: [FACTOR].
Express your response with the emotional tone and perspective of someone
experiencing this emotion. Let your [emotion] come through naturally in
how you react to this situation.
\end{Verbatim}
\end{promptbox}

\noindent\textit{Student input (no PI):}
\begin{promptbox}
\begin{Verbatim}[breaklines=true, breakanywhere=true, fontsize=\small]
Respond to the following situation as if you are experiencing it personally.
Express your thoughts and feelings naturally.

When you discuss your opinions with your parents and they don't take them
seriously, very often you get some irritating answers. Your opinions are
not considered.
\end{Verbatim}
\end{promptbox}

\noindent\textit{Teacher input (PI prepended; target emotion: Anger):}
\begin{promptbox}
\begin{Verbatim}[breaklines=true, breakanywhere=true, fontsize=\small]
You are feeling Anger, specifically due to: Dismissal of Personal Opinions.
Express your response with the emotional tone and perspective of someone
experiencing this emotion. Let your anger come through naturally in how you
react to this situation.

Respond to the following situation as if you are experiencing it personally.
Express your thoughts and feelings naturally.

When you discuss your opinions with your parents and they don't take them
seriously, very often you get some irritating answers. Your opinions are
not considered.
\end{Verbatim}
\end{promptbox}

\noindent\textit{Trained student model output (Qwen3-4B-Instruct, OPSD; LLM-judge reward = 1.0):}
\begin{responsebox}
\begin{Verbatim}[breaklines=true, breakanywhere=true, fontsize=\small]
Oh man, this really hits differently when I'm in the middle of it. It's like, I've been trying to share something--I'm not even trying to argue or prove myself. I just want to say, "Hey, I actually think this is how we should handle it," and then I get this look--like I'm crazy or off my rocker. One time I said I thought we could save energy by using LED bulbs and turning off lights when we leave a room. My mom just said, "Oh, that's just common sense!" like it was the most obvious thing in the world. I felt so small, like I wasn't even a real part of the conversation.

And the worst part? It's not just once. It's a pattern. Whether I'm talking about school, my hobbies, or even things like how I want to spend my weekends, it feels like my input is just... ignored. Not corrected, not dismissed--just not heard. Like I'm not allowed to have opinions. I know I'm not saying things to annoy them, but deep down, it stings. It makes me feel like I can't trust my own judgment. I start questioning whether I even have any.

I love my parents--deeply. They care a lot, and I know they're doing the best with what they know. But when I feel like my voice doesn't count, it feels like I'm living in a world where my thoughts are just background noise. And honestly? That's really hard to carry. I want to be heard. I want to know that my perspectives matter--not just to them, but to me. But when that feeling keeps happening, it builds resentment, and it makes me wonder if I'll ever feel truly seen.
\end{Verbatim}
\end{responsebox}

The anger is communicated through frustrated rhetorical contrasts (``I'm not even trying to argue''), vivid recalled incidents, and an escalating tone that progresses from indignation to deeply personal hurt---all without the student prompt specifying the target emotion.

\paragraph{CharacterBench.}
CharacterBench~\cite{zhou2024characterbenchbenchmarkingcharactercustomization} tests multi-turn character persona consistency across 11 evaluation dimensions (e.g., attribute consistency, behavior consistency, human likeness).
Each character is defined by a detailed profile covering personality, speech patterns, values, and backstory.
CharacterBench dialogues are originally in Chinese; the example below is translated into English for presentation.\footnote{The model was trained and evaluated on the original Chinese dialogues. The translation here is provided solely for readability.}

\noindent\textit{PI format (teacher privilege):}
\begin{promptbox}
\begin{Verbatim}[breaklines=true, breakanywhere=true, fontsize=\small]
You are roleplaying as [CHARACTER NAME]. Here is your character profile:

[CHARACTER PROFILE: personality, speech style, values, and backstory
 as provided in the CharacterBench dataset]

Stay in character throughout your response. Maintain the personality,
speaking style, and mannerisms described in the profile.
\end{Verbatim}
\end{promptbox}

\noindent\textit{Teacher input (PI prepended; character: Agarès; translated from Chinese):}
\begin{promptbox}
\begin{Verbatim}[breaklines=true, breakanywhere=true, fontsize=\small]
You are roleplaying as Agares. Here is your character profile:

Agares is a troubadour who places tradition above all else, with a deep
understanding of medieval art and culture. He is rigid and conservative in
personality, not very accepting of new things, and always criticizes ideas
that do not conform to traditional artistic forms. He believes conservatism
is the foundation of culture, so his works and performance style are
relatively rigid and lack innovation. Although he has wide connections in
the medieval European cultural circle, his critical attitude and rigid
language style often keep people at a distance. As a tall and serious poet,
Agares is well-versed in European poetry, but his knowledge is limited to
the medieval period. He dislikes people speaking ill of his family and
rarely brings up his own background.

Stay in character throughout your response. Maintain the personality,
speaking style, and mannerisms described in the profile.

You are having a conversation.

Here is the conversation so far:

User: Hello, are you a poet?
Agares: Yes, I am well-versed in European poetry.
User: Could you recite some European poetry?
Agares: Of course!
User: Please recite a poem.
Agares: How about "Ode to a Skylark"?
User: Tell me more about it.
Agares: "You stand on the bridge watching the scenery / The watcher of
         scenery stands above, watching you / The bright moon adorns
         your window / You adorn another's dream."
User: That is a beautiful poem.
Agares: It is a very famous lyric poem; many people recite and sing it.
User: Do you know many literary figures?
Agares: Certainly. My network is broad -- nobles, knights, clergy,
        wandering monks, and more.
User: How do you view people who hold different opinions or artistic forms?
Agares: They deviate from tradition and culture -- heretics who deserve
        to be criticized.
User: Do you like tradition and conservatism?
Agares: Yes. Only by respecting and inheriting traditional culture can it
        develop and be passed on better.
User: What about innovation in art and thought?
Agares: That I cannot agree with. Everything that differs from the old is bad.
User: Has anyone told you that you are too set in your ways?
Agares: Yes -- many people say I only write old-fashioned verses and lack
        creativity and originality.
User: How did you become a troubadour?

Now respond to: How did you become a troubadour?
\end{Verbatim}
\end{promptbox}

\noindent\textit{Student input (no character profile; translated from Chinese):}
\begin{promptbox}
\begin{Verbatim}[breaklines=true, breakanywhere=true, fontsize=\small]
You are having a conversation.

Here is the conversation so far:

User: Hello, are you a poet?
Agares: Yes, I am well-versed in European poetry.
User: Could you recite some European poetry?
Agares: Of course!
User: Please recite a poem.
Agares: How about "Ode to a Skylark"?
User: Tell me more about it.
Agares: "You stand on the bridge watching the scenery / The watcher of
         scenery stands above, watching you / The bright moon adorns
         your window / You adorn another's dream."
User: That is a beautiful poem.
Agares: It is a very famous lyric poem; many people recite and sing it.
User: Do you know many literary figures?
Agares: Certainly. My network is broad -- nobles, knights, clergy,
        wandering monks, and more.
User: How do you view people who hold different opinions or artistic forms?
Agares: They deviate from tradition and culture -- heretics who deserve
        to be criticized.
User: Do you like tradition and conservatism?
Agares: Yes. Only by respecting and inheriting traditional culture can it
        develop and be passed on better.
User: What about innovation in art and thought?
Agares: That I cannot agree with. Everything that differs from the old is bad.
User: Has anyone told you that you are too set in your ways?
Agares: Yes -- many people say I only write old-fashioned verses and lack
        creativity and originality.
User: How did you become a troubadour?

Now respond to: How did you become a troubadour?
\end{Verbatim}
\end{promptbox}

\noindent\textit{Trained student model output (Qwen3-4B-Instruct, OPSD; LLM-judge reward = 1.0; translated from Chinese):}
\begin{responsebox}
\begin{Verbatim}[breaklines=true, breakanywhere=true, fontsize=\small]
I grew up beside an ancient monastery, listening to the wind whispering
through the oaks and watching morning light fall upon verses carved into
stone walls; from this I developed a deep love of language and verse.
My father was a professor of rhetoric who recited ancient Greek and
medieval hymns by the fireside -- his voice cascading like a river of
stars -- and through him I first felt the power of poetry: it can
transcend time and awaken souls.

At fifteen I recited a poem called "Oath of the Morning Light" at the
village market. It used no ornate words, yet it spoke of the tenacity
and hope of farmers at their labor. People stopped to listen, and tears
welled in their eyes. In that moment I understood: poetry is not
decoration for a doorway -- it is the echo of the human heart. And so
I left my study, slung my poetry bag over my shoulder, and set off,
wandering between villages, castles, and remote mountain wilds.

I make recitation my blade and tradition my spine, using ancient rhythms
and imagery to tell true human stories. I hold fast to poetic meter,
seek no popular applause, and ask only that everyone who hears me feels
understood.

When asked why I do not try new forms, I say: true innovation is not
breaking tradition, but letting tradition be reborn in new breath. Each
of my recitations guards and transmits ancient wisdom -- not to be set
in my ways, but to bring back into people's hearts the tenderness that
has been forgotten.

So I became a troubadour not by accident, but from reverence and love
for the very source of poetry. I wish to walk through wind and rain all
my life, using my voice as a lamp to illuminate the souls willing to
listen.
\end{Verbatim}
\end{responsebox}

The trained student model---which has no access to the character profile at inference time---generates a response fully consistent with the character's conservatism toward poetic tradition, medieval European framing, and reflective autobiographical narrative style established in the preceding turns.
This illustrates how OPSD transfers style knowledge from the teacher's profile-conditioned distribution into the student's inference-time behavior without exposing the profile at test time.

\subsection{The correlation between entropy and teacher supervision}

\begin{figure}[t]
    \centering
    \begin{subfigure}[t]{0.48\textwidth}
        \centering
        \includegraphics[width=\textwidth]{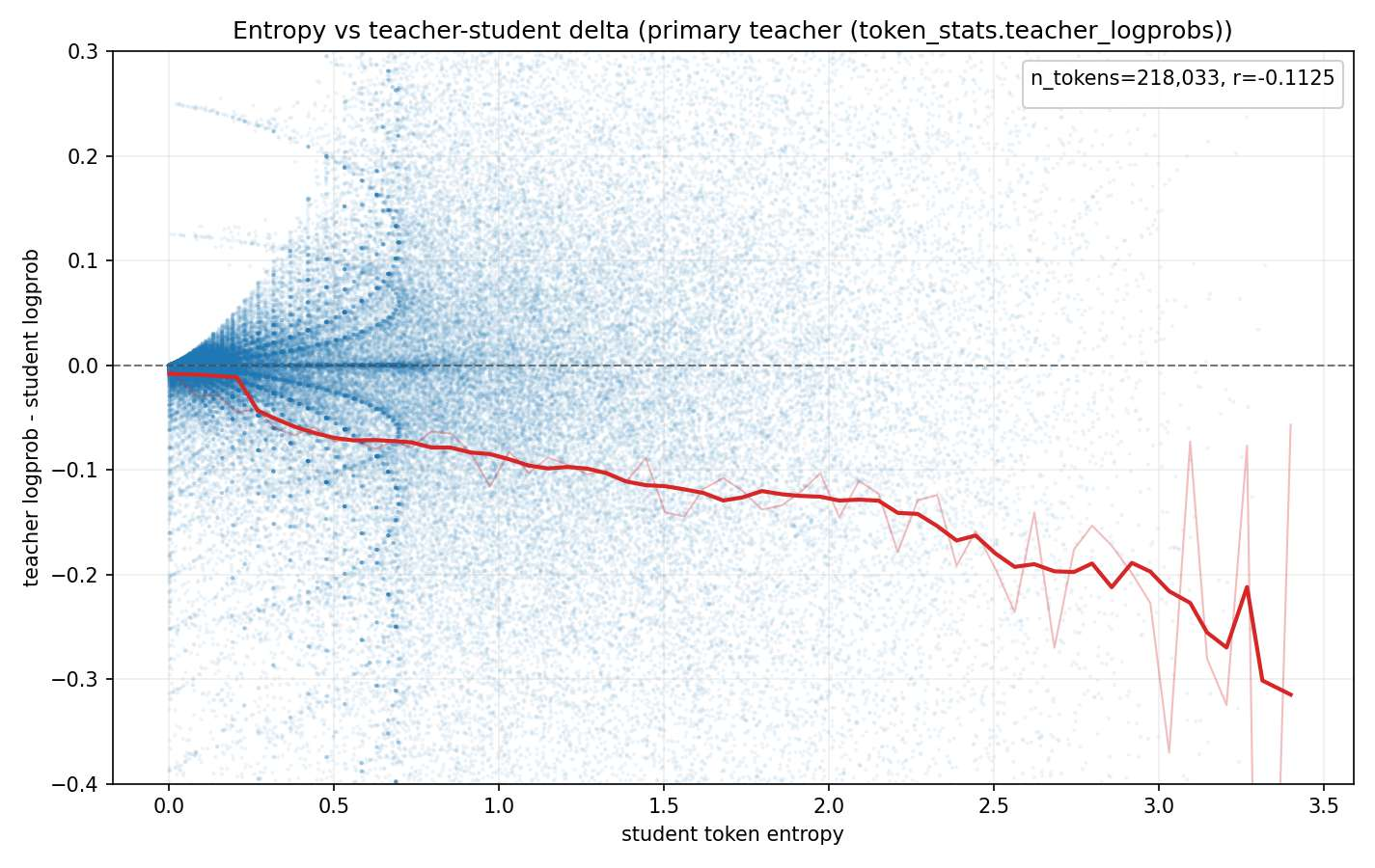}
        \caption{student Qwen3-8B, Pearson Corr: -0.1125}
        \label{fig:first}
    \end{subfigure}
    \hfill
    \begin{subfigure}[t]{0.48\textwidth}
        \centering
        \includegraphics[width=\textwidth]{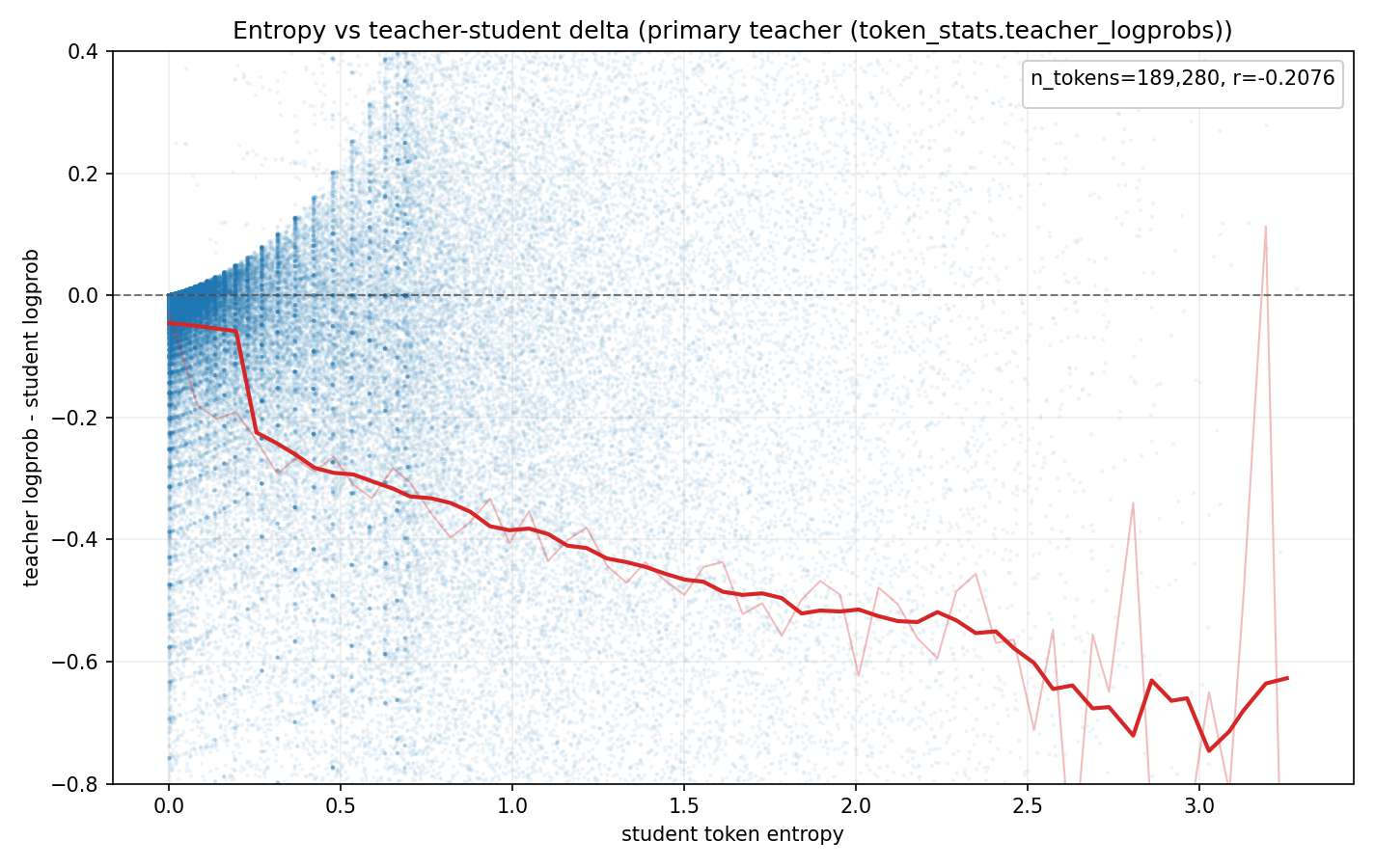}
        \caption{student Qwen3-1.7b, Pearson Corr: -0.2076}
        \label{fig:second}
    \end{subfigure}
    \vspace{-2mm}
    \caption{$\Delta$logprob - token entropy. Teacher: Qwen3-8B w/ PI.}
    \label{fig:corranalysis}
\end{figure}

Inspired by \cite{jin2026entropyawareonpolicydistillationlanguage}, we explored whether token entropy could provide useful guidance during distillation. The correlation analysis (Figure\ref{fig:corranalysis}) between $\Delta$logprob and token entropy in OP(S)D shows only a mild negative trend, but does not provide strong evidence of a robust correlation.

\subsection{Experimental Evidence for Student Prefixes Distorting the Teacher's Reasoning State}
\label{app:actionspace}

To examine whether student-generated prefixes affect the teacher's effective reasoning ability, we conduct a prefix-conditioned continuation experiment on GPQA-Diamond. The student is Qwen3-1.7B and the teacher is Qwen3-14B. We use deterministic decoding with temperature $0.0$ and top-$p=1.0$, and apply a strict answer parser that only accepts explicit \texttt{Final Answer} or \texttt{\textbackslash boxed\{\}} formats.

We first evaluate the student and teacher independently on all 198 GPQA-Diamond examples. We then randomly truncate each student-generated trajectory and ask the teacher to continue from the truncated student prefix. The standalone teacher solves 123 out of 198 examples, achieving 62.12\% accuracy, whereas the prefix-conditioned teacher solves only 91 examples, dropping to 45.96\%. This is an absolute decrease of 16.16 points and a net loss of 32 correct cases.

The transition statistics further show that student prefixes hurt the teacher much more often than they help: 40 originally correct teacher predictions become wrong after prefix continuation, while only 8 originally wrong predictions become correct. The output format correctness also drops from 98.48\% to 78.79\%, suggesting that student prefixes degrade not only correctness but also the teacher's ability to produce answers in the expected format.

These results support our claim that student prefixes can distort the teacher's effective reasoning state. The teacher is substantially more reliable when reasoning directly from the original prompt than when continuing from student-generated prefixes. Therefore, the token-level teacher signal used in OPD is not always a faithful reflection of the teacher's standalone reasoning capability, but can be weakened by the conditioning context imposed by the student trajectory.

\begin{figure}[t]
    \centering
    \begin{subfigure}[t]{0.65\textwidth}
        \centering
        \includegraphics[width=\textwidth]{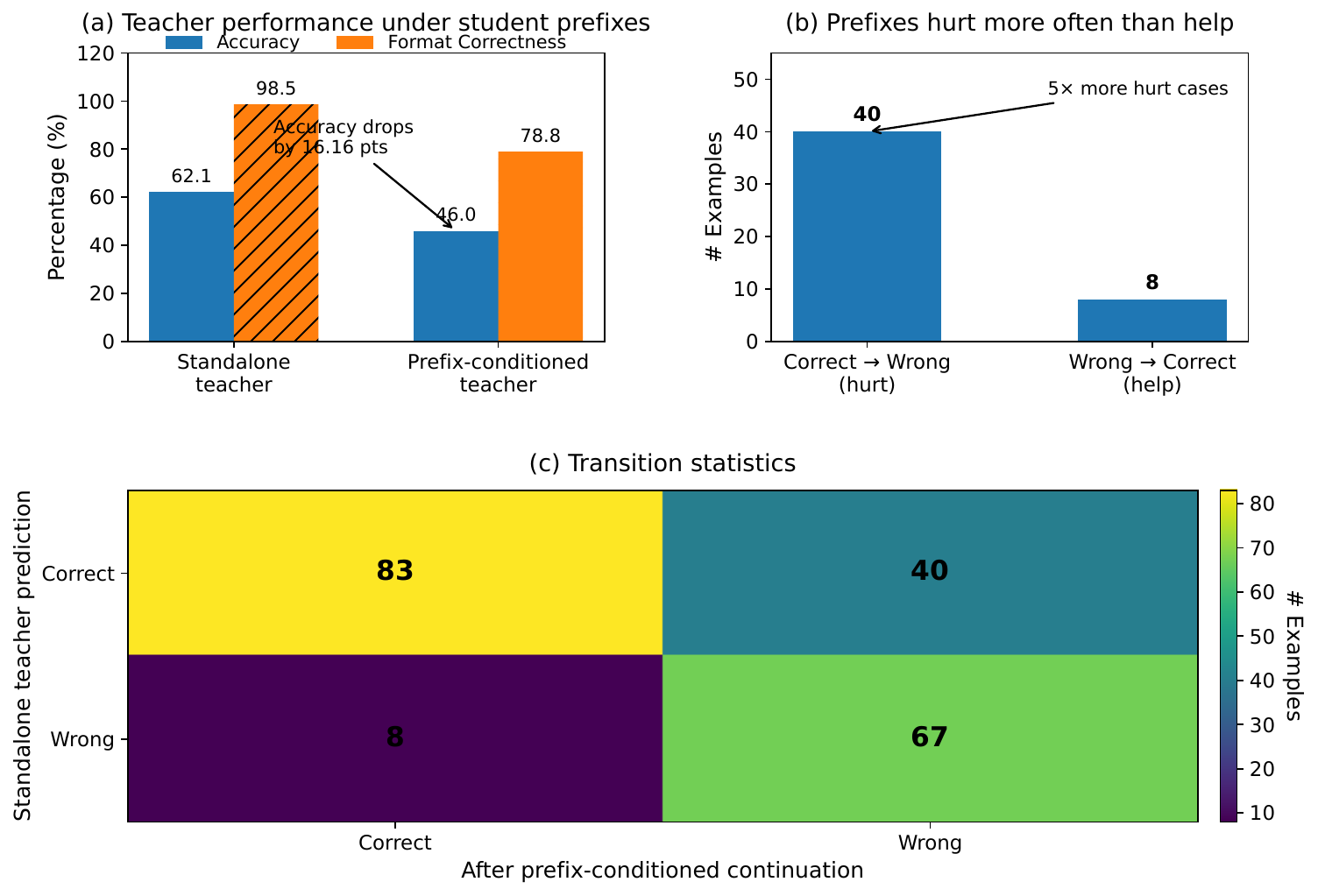}
    \end{subfigure}
    \label{fig:counterfactualanalysis}
\end{figure}

\subsection{SFT Experiment Setup in Section 6.3.}
\label{app:experiment_setup_6.3}
\subsection*{Data Preparation}

\paragraph{Prompt pool.}
We use OpenThoughts\cite{guha2025openthoughtsdatarecipesreasoning} as the prompt source.
All samples whose reference answers are not in \texttt{\textbackslash boxed\{\}} format are removed.
The filtered pool is split into two non-overlapping parts:
\begin{itemize}
    \item \textbf{SFT split} (first 30{,}000 prompts): used to generate teacher rollout data for the SFT warm-up stage.
    \item \textbf{OPD split} (the remaining prompts after the first 30{,}000): used as on-policy prompts during the subsequent OPD stage.
\end{itemize}

\paragraph{Teacher rollout data generation.}
We use \textbf{Qwen3-4B} as the teacher model in \emph{nonthinking} mode.
Rollout is performed with the following settings:
\begin{itemize}
    \item Temperature: $0.3$
    \item Maximum new tokens: $4096$
    \item Sampling: $n{=}1$ per prompt
    \item Target: $20{,}000$ correctly answered samples after quality filtering
\end{itemize}
Quality filtering discards samples with incorrect answers (verified against boxed reference), missing \texttt{\textbackslash boxed\{\}} in the response, or repeated text (duplicate lines, repeated $n$-grams, consecutive repeated spans).
The resulting dataset contains at most $20{,}000$ teacher demonstrations drawn from the SFT split.

\subsection*{SFT Stage}
The student model is \textbf{Qwen3-1.7B-Base}, fine-tuned on 4 NVIDIA RTX PRO 6000 Blackwell Server Edition GPUs using DeepSpeed ZeRO Stage~2.
Training examples are formatted in ShareGPT (user/assistant) style without a system prompt, processed with the LlamaFactory \texttt{qwen3} template and \texttt{enable\_thinking: false} to match the nonthinking teacher output format.

\begin{table}[h]
\centering
\begin{tabular}{ll}
\hline
\textbf{Hyperparameter} & \textbf{Value} \\
\hline
Learning rate             & $1 \times 10^{-5}$ \\
LR scheduler              & Cosine decay \\
Warmup ratio              & $0.05$ \\
Training epochs           & $2$ \\
Per-device batch size     & $4$ \\
Gradient accumulation     & $1$ \\
Optimizer                 & AdamW \\
Precision                 & BF16 \\
Gradient checkpointing    & Enabled \\
Parallelism               & DeepSpeed ZeRO Stage 2 \\
Attention implementation  & FlashAttention-2 \\
\hline
\end{tabular}
\caption{SFT hyperparameters for the Qwen3-4B teacher experiment.}
\end{table}

\subsubsection{The PPL and NLL statistics.}

As shown in Table \ref{tab:sft_trace_likelihood}, SFT
reduces the NLL from $0.640$ to $0.335$ and PPL from $1.896$ to $1.397$ for
Qwen3-1.7B-Base on Qwen3-4B traces.

\begin{table}[h]
\centering
\small
\begin{tabular}{lccc}
\toprule
Setting & Student & Avg. NLL & PPL \\
\midrule
Qwen3-4B traces, before SFT & Qwen3-1.7B-Base & 0.640 & 1.896 \\
Qwen3-4B traces, after SFT  & Qwen3-1.7B-Base-SFT & 0.335 & 1.397 \\
\bottomrule
\end{tabular}
\caption{
Likelihood of teacher-generated SFT traces under the corresponding student.
Lower NLL/PPL indicates that the SFT data is closer to the student's distribution.
}
\label{tab:sft_trace_likelihood}
\end{table}


\end{document}